\documentclass[12pt]{article}
\usepackage{enumerate}
\usepackage[T1]{fontenc}
\usepackage[usenames]{color}
\usepackage[colorlinks,
linkcolor=red,
anchorcolor=blue,
citecolor=blue
]{hyperref}
\usepackage{mathrsfs}
\usepackage[protrusion=true, expansion=true]{microtype}
\usepackage{float}
\usepackage{subfigure}
\usepackage{amsfonts,amsmath,amssymb,amsthm,mathtools,url,xspace,bm,multirow,relsize}
\usepackage{tikz,verbatim,xparse,scalerel,floatflt}
\usetikzlibrary{arrows,shapes}
\usetikzlibrary { decorations.pathmorphing, decorations.pathreplacing, decorations.shapes, }
\usetikzlibrary{decorations.pathreplacing}
\tikzset{mybrace/.style={decorate,decoration={brace,aspect=#1}}}
\usetikzlibrary{shapes.geometric,calc}
\usepackage{authblk}
\usepackage[bottom]{footmisc}
\usepackage{enumitem}
\usepackage{lscape}
\usepackage{caption}
\usepackage{soul}
\usepackage{graphicx}
\usepackage{url,array,dsfont,makecell,booktabs} 
\usepackage{algorithm,algorithmicx,algpseudocode}
\usepackage{diagbox, wrapfig}
\usepackage{titlesec}
\usepackage{math,mycite} 
\usepackage{caption}

\titlespacing*{\section}
{0pt}    % left margin
{0.8em}  % space before
{0.4em}  % space after
\titlespacing*{\subsection}
{0pt}    % left margin
{0.1em}  % space before
{0.3em}  % space after
\titlespacing*{\subsubsection}
{0pt}    % left margin
{0.1em}  % space before
{0.3em}  % space after

\ifx\counterwithout\undefined\usepackage{chngcntr}\fi
\counterwithout{equation}{section}

\newcommand{\anon}{1}

\makeatletter
\g@addto@macro\normalsize{%
	\setlength\abovedisplayskip{4pt}
	\setlength\belowdisplayskip{4pt}
	\setlength\abovedisplayshortskip{2pt}
	\setlength\belowdisplayshortskip{2pt}
}
\makeatother

\mathtoolsset{showonlyrefs=true}
\allowdisplaybreaks[1]

\begin{document}

\def\spacingset#1{\renewcommand{\baselinestretch}%
{#1}\small\normalsize} \spacingset{1}

\date{}

\if1\anon
{\title{\bf Online Inference of Constrained Optimization: Primal-Dual Optimality and Sequential Quadratic Programming}
\author{Yihang Gao\vspace{-0.45cm}\\
Department of Mathematics, National University of Singapore\vspace{0.15cm}\\
Michael K. Ng\\
Department of Mathematics, Hong Kong Baptist University\vspace{0.15cm}\\
Michael W. Mahoney\\
ICSI and Department of Statistics, University of California, Berkeley\vspace{0.15cm}\\
Sen Na\\
School of Industrial and Systems Engineering, Georgia Institute of Technology}
\maketitle} \fi

\if0\anon
{\bigskip\bigskip\bigskip
\begin{center}
{\LARGE\bf Title}
\end{center}
\medskip} \fi

%\bigskip
\begin{abstract}
We study online statistical inference for the solutions of stochastic optimization problems with equality and inequality constraints. Such problems are prevalent in statistics and machine learning, encompassing constrained $M$-estimation, physics-informed models, safe reinforcement learning, and algorithmic fairness. We develop a stochastic sequential quadratic programming (SSQP) method to solve these problems, where the~step direction is computed by sequentially performing a quadratic approximation of the objective and a linear approximation of the constraints. 
Despite having access to unbiased estimates of population gradients, a key challenge in constrained stochastic problems lies in dealing with the bias in the step direction. As such, we apply a momentum-style~gradient moving-average technique within SSQP to debias the step. We show that our method achieves global almost-sure convergence and exhibits local asymptotic normality with an \textit{optimal} primal-dual limiting covariance matrix in the sense of H\'ajek and Le Cam. In addition, we provide a plug-in covariance matrix estimator for practical inference. To our knowledge, the proposed SSQP method is the \textit{first fully online method} that attains primal-dual asymptotic minimax optimality without relying on projection operators onto the constraint set, which are generally intractable for nonlinear problems.~Through extensive experiments on benchmark nonlinear problems, as well as on constrained generalized linear models 
and portfolio allocation problems using both synthetic and real~data,~we~demonstrate superior performance of our method, showing that the method and its asymptotic behavior~not~only solve constrained stochastic problems~efficiently but also provide~valid~and~practical online inference in real-world applications.

\end{abstract}

\noindent%
{\it Keywords:} constrained model inference; primal-dual minimax optimality; stochastic SQP; gradient momentum
\vfill

\newpage
\spacingset{1.8} % DON'T change the spacing!

%\begin{refsection}
\section{Introduction}\label{sec:1}

We consider stochastic optimization problems with equality and box inequality constraints:$\quad$
\begin{equation}\label{original_problem0}
\min_{\bm{x} \in \mathbb{R}^{d}} \hspace{1em} f(\bm{x}) = \mathbb{E}_{\zeta \sim \P} \left[ F(\bm{x}; \zeta) \right] \quad \text{s.t.}  \hspace{1em} \bm{c}(\bm{x}) = \bm{0},\quad \bm{\ell} \leq \bm{x} \leq \bm{u}.
\end{equation}
Here, $\bm{\ell}, \bm{u}\in\mR^d$ denote the lower and upper bounds, respectively, with "$\leq$" representing~element-wise comparison; $F(\cdot; \zeta):\mR^d\rightarrow \mR$ denotes a realization of the stochastic objective $f$; and $\bm{c}:\mR^d\rightarrow\mR^m$ encodes deterministic equality constraints. 
Throughout the paper, we assume $f$, $\bm{c}$, $F(\cdot; \zeta)$ are twice continuously differentiable while potentially nonconvex and nonlinear.~Problem \eqref{original_problem0} readily accommodates nonlinear inequality constraints $\bc_{\mI}(\bx)\leq \0$ by introducing slack variables, i.e., reformulating them as $\bc_{\mI}(\bx) + \by = \b0$ and $\by \geq \b0$.

Problem \eqref{original_problem0} is ubiquitous with many statistical and machine learning applications, where constraints can encode prior domain knowledge, ensure models' identifiability, and reduce models' intrinsic dimensionality. We introduce concrete motivating examples in Appendix \ref{append:A}.
Given the ubiquity of Problem \eqref{original_problem0}, statisticians aim to estimate its (local) solution $\tx$ and perform statistical inference. Arguably, the most primitive estimator is constrained $M$-estimator, where we draw $n$ samples $\zeta_1,\ldots,\zeta_n\stackrel{\text{iid}}{\sim} \P$ and estimate the population loss $f$ by the empirical loss~$\hf_n$:$\quad\quad$
\begin{equation}\label{snequ:1}
\min_{\bm{x} \in \mathbb{R}^{d}} \hspace{1em} \hf_n(\bx)\coloneqq \frac{1}{n}\sum_{i=1}^{n} F(\bx;\zeta_i) \quad \text{s.t.} \hspace{1em} \bm{c}(\bm{x}) = \bm{0},\quad \bm{\ell} \leq \bm{x} \leq \bm{u}.
\end{equation}
In fact, it is known that the above constrained $M$-estimation attains \textit{asymptotic optimality} in H\'ajek and Le Cam's sense \citep{Hajek1972Local, LeCam1972Limits}. Roughly speaking, under certain regularity conditions, the (local) minimizer $\hat{\bm{x}}_{n}$ of Problem \eqref{snequ:1} exhibits asymptotic normality~with the smallest covariance matrix, given by
\begin{equation}\label{M-estimator}
\sqrt{n} \left( \hat{\bm{x}}_{n} - \tx \right) \stackrel{d}{\longrightarrow} \mathcal{N} \left( \bm{0},  \bm{L}^{\dag} \text{Cov}\left( \nabla F(\tx;\zeta) \right) \bm{L}^{\dag}\right),
\end{equation}
where $\bm{L} = \bP_{\text{Null}(\tJ)} \nabla_{\bx}^2\mL^\star\bm{P}_{\text{Null}(\tJ)}$ and $\dagger$ is the Moore-Penrose pseudoinverse. Here, we let
\begin{equation}\label{nsequ:2}
\mL(\bx, \blambda, \bmu) \coloneqq	f(\bx) + \blambda^\top \bc(\bx) +  \bmu_1^\top\left(\bell - \bx \right) + \bmu_2^\top\left(\bx - \bu \right)
\end{equation}
be the Lagrangian function of \eqref{original_problem0} with $\blambda\in\mR^m$ and $\bmu = (\bmu_1, \bmu_2)\in\mR^{2d}$ being the dual~variables associated with the equality and inequality constraints; $\nabla_{\bx}^2\mL^\star = \nabla_{\bx}^2\mL(\tx,\tlambda,\tmu)$ be the Lagrangian Hessian at the primal-dual solution $(\tx,\tlambda,\tmu)$ (with respect to $\bx$); and $\bP_{\text{Null}(\tJ)}$ denote the projection matrix onto the null space of $\bJ^\star = \bJ(\tx)$, where $\bJ(\bx)$ is the Jacobian of the active constraints at $\bx$ (see \eqref{def:activeJacobian}). We refer to \cite{Duchi2021Asymptotic, Davis2024Asymptotic} for the rigorous statement of the result \eqref{M-estimator}; and see Theorem \ref{thm:2} for primal-dual generalization.$\hskip1cm$

Although $\hat{\bm{x}}_{n}$ enjoys nice properties, the offline $M$-estimation often requires dealing with a full batch of samples, leading to significant computational and memory costs. Over the last few decades, online methods that can deal with online streaming data have been preferred.~One of the most fundamental online methods is Stochastic Gradient Descent (SGD) \citep{Robbins1951Stochastic, Kiefer1952Stochastic}. 
Recent work has significantly advanced local asymptotic analysis of SGD (and its variates) to enable online inference of $\tx$ with both iid and Markovian data. See \cite{Polyak1992Acceleration, Toulis2017Asymptotic, Fang2018Online, Toulis2021Proximal, Chen2020Statistical, Chen2024Online, Zhu2021Online, Lee2022Fast} and references therein.$\quad\quad$

The above literature all studied stochastic optimization in an unconstrained setting, while to accommodate constraints, \cite{Duchi2021Asymptotic} designed a Projected Riemannian Stochastic Gradient method, which to our knowledge is one of the \textit{first fully online} methods capable of achieving asymptotic optimality for constrained model parameters $\tx$.
The authors also identified a gap between problems with linear and nonlinear constraints, illustrating the reason why vanilla Projected SGD,
\begin{equation}\label{nsequ:3}
\bx_{k+1} = \text{Proj}_{\Omega}\rbr{\bx_k - \alpha_k\nabla F(\bx_k; \zeta_k)},\quad \quad \Omega = \{\bx\in\mR^d: \bc(\bx) = \0, \; \bell\leq \bx \leq \bu\},
\end{equation} 
can be sub-optimal for nonlinear constraints. Notably, the gap has recently been closed by \cite{Davis2024Asymptotic}, showing that the averaged Projected SGD, $\bar{\bx}_n = \frac{1}{n}\sum_{k=0}^{n-1}\bx_k$, actually attains the same asymptotic optimality for nonlinear constraints as displayed in \eqref{M-estimator}. A consistent limiting covariance estimator has also been proposed in the follow-up work \cite{Jiang2025Online}.

Despite these advances in online constrained inference, existing methods rely on a projection operator onto the feasible set $\text{Proj}_{\Omega}(\cdot)$, making them, as stated in \cite{Duchi2021Asymptotic}, \textit{more intellectually intriguing than practically applicable}. In particular, projections require solving nonlinear equations that is often intractable and needs global knowledge of the feasible~set~$\Omega$.~However, many problems (cf.~CUTEst in Section \ref{sec:5}) only have access to black-box oracles that return the local constraint value and Jacobian at a given input, making projections inapplicable. 
Even in the ideal case where we have global information of the feasible set, the projection at each step may only be approximately computed by solving an intensive optimization subproblem; yet the resulting approximation error is typically neglected in existing analyses, which rely heavily on exact computation of $\text{Proj}_{\Omega}(\cdot)$.
Furthermore, a key distinction between constrained and unconstrained problems lies in the presence of dual multipliers. The dual component is not involved in existing methods, yet it plays a crucial role in stationarity certification and active-set identification (as ensured by complementary slackness theorem; see \eqref{nsequ:KKT}). Thus,~\mbox{performing}~primal-dual joint inference is of fundamental interest. This motivation leads to the following question: \textit{What is the asymptotic minimax optimality for the primal–dual solution $(\tx, \tlambda, \tmu)$ of Problem \eqref{original_problem0}, and how to obtain the optimal joint online estimator?}

\subsection{Backbone: Sequential Quadratic Programming}\label{sec:1.1}

In this paper, we first establish primal-dual asymptotic minimax optimality of $(\tx, \tlambda, \tmu)$, and then design a \textit{Stochastic Sequential Quadratic Programming} (SSQP) method to attain the lower bound. The idea is to \textit{localize} the model along with constraints, and utilize dual~information of constrained local model when updating the estimate. In particular, at the $k$-th step, we perform a quadratic approximation of the loss and a linear approximation of the constraints~in~\eqref{original_problem0}:$\quad\quad$
\begin{equation}\label{nsequ:1}
\min_{\barDelta\bx_k\in\mR^d} \nabla F(\bx_k;\zeta_k)^\top\barDelta\bx_k + \frac{1}{2}\barDelta\bx_k^\top\barB_k\barDelta\bx_k,\;\; \text{s.t.}\;\; \bc_k + \nabla \bc_k\barDelta\bx_k = \0,\;\;  \bm{\ell} \leq \bx_k+ \barDelta\bx_k\leq \bm{u}.
\end{equation}
Here, we denote the constraint value and Jacobian by $\bc_k = \bc(\bx_k)$ and $\nabla \bc_k = \nabla \bc(\bx_k)$; and~$\barB_k$, rather than estimating $\nabla^2 f_k$, estimates the Lagrangian Hessian $\nabla^2_{\bx}\mL_k$ to leverage the curvature information of nonlinear constraints. After solving \eqref{nsequ:1} and obtaining the primal-dual~solution $(\barDelta\bx_k, \barblambda_k^{\text{sub}}, \barbmu_k^\text{sub})$, we update the iterate as
\begin{equation*}
(\bx_{k+1}, \blambda_{k+1}, \bmu_{k+1}) = (\bx_k, \blambda_k, \bmu_k) + \alpha_k (\barDelta\bx_k, \barblambda_k^{\text{sub}} - \blambda_k, \barbmu_k^\text{sub}-\bmu_k).
\end{equation*}
Although the above scheme streamlines the estimation procedure by solving a linear–quadratic program at each step, two fundamental challenges remain. First, even if the original nonlinear constraints $\bc(\bx) = \0$, $\bell \leq \bx \leq \bu$ admit a nonempty feasible set $\Omega$ (cf. \eqref{nsequ:3}), their linearization in \eqref{nsequ:1} may become infeasible.~Second, the inequality constraints truncate the noise in~a~way~such that, even if $\nabla F(\bx_k; \zeta_k)$ is an unbiased estimator of the true gradient $\nabla f_k$, the resulting step direction $\barDelta\bx_k$ can still be a biased estimate of its deterministic counterpart, thereby preventing convergence of the scheme. 
To address these two challenges, we respectively introduce a constraint relaxation technique and a gradient (and Hessian) moving average technique. Although momentum-based gradient method has been widely studied in optimization~contexts since the introduction in \cite{Polyak1964Some}, our method is the first attempt to incorporate both gradient and Hessian momentum into inequality-constrained problems and show promising asymptotic properties to facilitate inference. Furthermore, our method allows an adaptive random stepsize $\alpha_k$, which is often preferred in practice.

Specifically, with all above designs, we show that the \textit{last} SSQP iterate exhibits asymptotic normality, with an optimal primal-dual covariance matrix matching the theoretical lower bound. In particular, the marginal primal covariance of the SSQP estimator also matches that of the offline constrained $M$-estimator and online averaged Projected SGD estimator, as shown in \eqref{M-estimator}. Moreover, we propose a plug-in limiting covariance matrix estimator to facilitate practical inference. We demonstrate the superiority of our method through extensive experiments on benchmark nonlinear problems, constrained generalized linear models, and portfolio allocation problems with both synthetic data and real \texttt{Fama-French Portfolios} and \texttt{Chicago Air Pollution} data. Note that projection-based methods may not be applicable or competitive in the experiments (e.g., \cite{Duchi2021Asymptotic} considered a linear regression problem~with~$d = 2$).$\quad\quad$

\subsection{Related literature and contribution}

Inspired by the profound success of deterministic SQP in numerical optimization \citep{Nocedal2006Numerical}, recent research has extended SQP framework to stochastic settings \citep{Na2022adaptive, Na2023Inequality, Na2025Derivative, Curtis2023Worst, Curtis2024Sequential, Berahas2021Sequential,  Berahas2023Stochastic, Fang2024Fully, Fang2024Trust, Fang2025High}. However, existing work has primarily focused on equality-constrained \mbox{problems}.~Only~\cite{Na2023Inequality, Curtis2024Sequential} considered inequality constraints, yet both required strong constraint qualification conditions to circumvent infeasibility issue and a gradually increasing batch size for convergence, making them inapplicable in online settings. Moreover, all above numerical work studied only convergence of SSQP but fell short on uncertainty quantification. In contrast, we not only inherit the strengths of the above numerical methods by allowing adaptive stepsizes, but also significantly refine their designs to fit in \textit{fully online} settings. 
Furthermore, we bridge the two worlds of optimization and statistics, leveraging the SSQP methods to enable online constrained statistical inference, thereby addressing practical limitations of projection-based methods as stated in \cite{Duchi2021Asymptotic, Davis2024Asymptotic, Jiang2025Online}.$\quad\quad\quad$

This paper also relates to the growing literature on online inference via second-order methods. \cite{Bercu2020Efficient, Boyer2022asymptotic, Leluc2023Asymptotic, Cenac2025efficient} studied online Newton (or conditioned SGD) methods for (non)linear regression problems, establishing asymptotic normality for the resulting Newton estimators. \cite{Kuang2025Online} proposed a consistent covariance estimator for more general sketched Newton methods. \cite{Na2025Statistical} established asymptotic normality for a sketched SSQP method under equality constraints, while \cite{Du2025Online} leveraged a random scaling technique to \mbox{construct}~\mbox{pivotal}~statistics.
However, due to the absence of inequality constraints, none of these works address the challenges of infeasibility and biased step directions discussed in Section \ref{sec:1.1}. In fact, to the best of our knowledge, establishing asymptotic distribution of online methods with gradient (and Hessian) momentum remains open even in unconstrained SGD settings, while our work establishes such results for inequality-constrained problems. We clearly demonstrate how the~(adaptive) stepsize and momentum weight are interrelated: gradient averaging reduces direction~bias to facilitate convergence of the iterates, while past gradient information is also forgotten fast enough to attain optimal asymptotic normality.

\subsection{Organization and notation}

Section \ref{sec:2} introduces the preliminaries of Problem \eqref{original_problem0}, including the infeasibility issue, constraint qualification, and constraint relaxation. We also establish the primal–dual asymptotic minimax optimality of Problem \eqref{original_problem0}. Section \ref{sec:3} introduces the proposed SSQP method utilizing gradient moving average, and establishes its global almost-sure convergence guarantee.~Section \ref{sec:4} presents the asymptotic normality result and proposes a consistent plug-in covariance matrix estimator. Section \ref{sec:5} reports extensive experimental results on both synthetic and real data, followed by conclusions and future work discussions in Section \ref{sec:6}.

Throughout the paper, we use $\|\cdot\|$ to denote the $\ell_2$ norm for vectors and the spectral~norm for matrices.
We use boldface lowercase and capital letters ($\bm{a}$ and $\bm{A}$) to denote vectors and~matrices, respectively.~For a positive integer $m$, $[m] \coloneqq \{1, \cdots, m\}$. For an index set $\mA \subseteq [m]$, $\mA^{-} \coloneqq [m] \setminus \mA$ and $|\mA|$ denotes its cardinality. For a vector $\ba\in\mR^m$, $\ba_\mA\in\mR^{|\mA|}$ denotes~the~\mbox{subvector}~containing only entries with indices in $\mA$. Similarly, for a matrix $\bA\in\mR^{m\times d}$, $\bA_\mA\in\mR^{|\mA|\times d}$ denotes the submatrix containing only rows with indices in $\mA$.  
Without ambiguity, the $i$-th entry of a vector $\bm{a}$ is written as either $[\bm{a}]_i$ or $a_i$.
We use $\mathcal{O}(\cdot)$ to denote the standard big-$O$ \mbox{notation}:~$a_k = \mathcal{O}(b_k)$ if $|a_k/b_k| \leq C $ for some constant $C > 0$ and large enough $k$. We use $\odot$ to denote the Hadamard (element-wise) product.
For notational brevity, we write $\bm{c}_k := \bm{c}(\bm{x}_k)$ and $f_k := f(\bm{x}_k)$ (similarly for $\nabla \bc_k$, etc.) as abbreviations.

\section{Preliminary and Asymptotic Optimality}\label{sec:2}

Recall the Lagrangian function $\mL(\bx, \blambda, \bmu)$ of Problem \eqref{original_problem0} is defined in \eqref{nsequ:2}. Under proper constraint qualifications, a necessary condition for $\tx$ being a local solution of \eqref{original_problem0} is the KKT~conditions: there exist dual multipliers $\tlambda\in\mR^d$ and $\tmu = (\tmu_1, \tmu_2)\in\mR^{2d}$ such that
\begin{equation}\label{nsequ:KKT}
\begin{gathered}
\nabla_{\bx}\mL(\tx,\tlambda,\tmu) = \0,\quad\quad\quad \bc(\tx) = \0,\quad\quad\quad \bell\leq \tx\leq \bu, \\
\tmu\geq \0,\quad\quad (\tmu_1)^\top(\bell-\tx) = (\tmu_2)^\top(\tx-\bu) = 0.
\end{gathered}
\end{equation}
At any $\bx$, we let $\nabla \bc(\bx)\in\mR^{m\times d}$ be the Jacobian of $\bc(\bx)$.~We denote active sets for inequalities~as
\begin{equation}\label{nsequ:5}
\mA_\bell(\bx) = \{i\in[d]: [\bm{x}]_i = [\bm{\ell}]_i\}\quad\quad \text{ and }\quad\quad \mA_\bu(\bx) = \{i\in[d]: [\bm{x}]_i = [\bm{u}]_i\}.
\end{equation} 
The active constraints include equality and active inequality constraints, $(\bc(\bx); [\bell-\bx]_{\mA_\bell(\bx)}; [\bx-\bu]_{\mA_\bu(\bx)})$; and their full Jacobian is given~by
\begin{equation}\label{def:activeJacobian}
\bJ(\bx) \coloneqq \begin{pmatrix}
\nabla \bm{c}(\bm{x}) \\
-\bI_{\mA_\bell(\bx)}\\
\bI_{\mA_\bu(\bx)}
\end{pmatrix}\in\mR^{(m+|\mA_\bell(\bx)|+|\mA_\bu(\bx)|)\times d},
\end{equation}
where $\bI\in\mR^{d\times d}$ is the $d$-dimensional identity matrix.

\vspace{-0.1cm}
\subsection{Infeasibility and constraint relaxation}\label{sec:2.1}
\vspace{-0.1cm}

Compared to the projection operator $\text{Proj}_{\Omega}(\cdot)$ in \eqref{nsequ:3}, the constraint linearization in \eqref{nsequ:1} requires only local information and leads to a much simpler subproblem. However, an intrinsic difficulty of linearization is the issue of \textit{infeasibility}: even if the original constraint set $\Omega \neq \emptyset$, the linearized constraints may end up with an empty set.

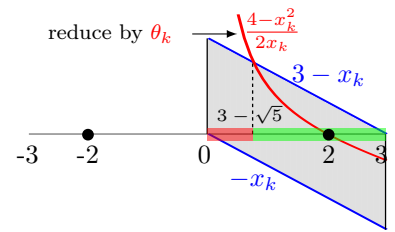
\begin{wrapfigure}[6]{R}{0.3\textwidth}
\vskip-0.8cm
\begin{tikzpicture}[align=center,transform shape]
\node[align=center] at (0,0){
\begin{tikzpicture}
\node at (0,0){
\resizebox{5cm}{3cm}{
\begin{tikzpicture}[transform shape]
\draw (-3,0)-- (3,0);
\foreach \i in {0,1,...,30}{
\node(m) at ($(0,3)!\i/30!(3,0)$){};
\node(n) at ($(0,0)!\i/30!(3,-3)$){};
\fill [black, opacity=0.005]
(3,0) -- (m.center) -- (n.center) -- (3,-3) -- cycle;}
\foreach \i in {0,1,...,30}{
\node(m) at ($(3,-3)!\i/30!(0,0)$){};
\node(n) at ($(3,0)!\i/30!(0,3)$){};
\fill [black, opacity=0.005] 
(0,0) -- (m.center) -- (n.center) -- (0,3) -- cycle;}								
\node at (1.52,0.38){
\begin{tikzpicture}[domain=0:3,transform shape]
\draw[color=blue,line width=0.5mm] plot (\x,-\x);
\draw[color=blue,line width=0.5mm]    plot (\x,3-\x);
\draw[color=red,domain=0.5:3,line width=0.5mm] plot ({\x},{(4-\x^2)/(2*\x)});
\end{tikzpicture}};		
\draw (0,0) -- (0,3);
\draw (3,0) -- (3,-3);									
\draw[dashed] (0.764,0) -- (0.764,2.236);											
\fill[red,opacity=.5] (0,-0.2) rectangle (0.764,0.2);
\fill[green,opacity=.5] (0.764,-0.2) rectangle (3,0.2);
\end{tikzpicture}}};
\node[font=\footnotesize] at (-2.4,-0.43) {-3};
\node[font=\footnotesize] at (2.3,-0.43) {3};
\node[font=\footnotesize] at (-0.05,-0.43) {0};
\node[font=\footnotesize] at (-1.6,-0.43) {-2};
\node[font=\footnotesize] at (1.6,-0.43) {2};
\node[scale=.4,fill,circle] at (-1.6,-0.17){};
\node[scale=.4,fill,circle] at (1.6,-0.17){};
\node[below,blue,font=\footnotesize] at (0.6,-0.5) {$-x_k$};
\node[below,blue,font=\footnotesize] at (1.6,0.9) {$3-x_k$};
\node[below,red,font=\footnotesize] at (0.85,1.65) {$\frac{4-x_k^2}{2x_k}$};
\node[font=\tiny,scale=1.1] at (0.55,0.1) {$3-\sqrt{5}$};
\node[text opacity=1,font=\scriptsize,rounded corners,line width=1] at (-1.3,1.17) (C1) {reduce by {\red $\theta_k$}};
\draw[-latex] ($(C1.east)+ (0.1,0)$) -- ($(C1.east) + (0.7,0)$);
\end{tikzpicture}};	
\end{tikzpicture}
\vskip-0.6cm
\caption{\textit{Insights of our constraint relaxation.}}\label{fig:1}
\end{wrapfigure}
Consider an illustrative example with $d=m=1$, where~we have $c(x) = x^2 - 4$ and constraints $0 \leq x \leq 3$. We know the constraint set $\Omega = \{x=\pm 2\}\cap [0,3] = \{2\}$ is nonempty. However, for any $x_k\in(0, 3]$, the linearized equality constraint~is~$x_k^2-4 + 2x_k \barDelta x_k = 0 \Leftrightarrow \barDelta x_k = (4-x_k^2)/(2x_k)$, depicted by the~red~curve in Figure \ref{fig:1}; while the linearized inequality constraint is $-x_k\leq \barDelta x_k\leq 3-x_k$, depicted by the shaded region.~From Figure \ref{fig:1}, we see that the linearized equality and inequality constraints intersect only if $x_k \geq 3 - \sqrt{5}$, suggesting an infeasibility issue~for Problem \eqref{nsequ:1} when $x_k \in (0, 3 - \sqrt{5})$. We need constraint relaxation in this regard. $\quad\;$

The constraint relaxation introduced in this paper stems from the observation that the~nonlinear curve in Figure \ref{fig:1} can be flattened by a proper scaling factor $\theta_k\in(0, 1]$, so that the~curve has more intersections with the shaded region. In particular, we consider the relaxed version~of~\eqref{nsequ:1}:
\begin{equation}\label{nsequ:4}
\theta_k\cdot \bc_k + \nabla \bc_k\barDelta\bx_k = \0,\quad\quad\quad \bell\leq \bx_k + \barDelta\bx_k \leq \bu.
\end{equation}
Naturally, this leads us to investigate the conditions under which a relaxation parameter $\theta_k$ exists or fails to exist. 
We find that this is closely tied to the extended generalized Mangasarian-Fromowitz constraint qualification (EGMFCQ, \cite{Xu2015Smoothing}, Definition 2.4), which is weaker than linear independence constraint qualification (LICQ) that needs to be imposed for asymptotic minimax optimality analysis; see \cite[Assumption B]{Duchi2021Asymptotic} and \cite[Example 2.1]{Davis2024Asymptotic} for references.

\begin{definition}[EGMFCQ v.s.~LICQ]\label{def_MFCQ}
The extended generalized Mangasarian-Fromowitz~constraint qualification is said to hold at $\bx \in \mathbb{R}^{d}$ satisfying $\bell\leq \bx\leq \bu$ if the two conditions~are~met:
\begin{enumerate}[label=(\alph*),topsep=-6pt]%(\roman*)(\arabic*)(\alph*)%\setcounter{enumi}{6}
\setlength\itemsep{-6pt}
\item The Jacobian $\nabla \bc(\bx) \in\mR^{m\times d}$ has full row rank.
\item There exists a vector $\bz \in \mR^{d}$ such that
\begin{equation}\label{MFCQ_z}
\bc(\bx) + \nabla \bm{c}(\bx)\bm{z} = \bm{0}, \quad\quad [\bm{z}]_i>0 \; \text{ for }\; i\in \mA_\bell(\bx),\quad\quad [\bm{z}]_i<0 \; \text{ for }\; i\in \mA_\bu(\bx).
\end{equation}
\end{enumerate}
The linear independence constraint qualification is said to hold at $\bx\in\mR^d$ if the Jacobian of~active constraints $\bJ(\bx)$ defined in \eqref{def:activeJacobian} has full row rank.
\end{definition}

Setting $\bp = -(\bc(\bx);\1_{\mA_\bell(\bx)};\1_{\mA_\bu(\bx)})$, we note that LICQ ensures the linear \mbox{system}~$\bJ(\bx)\bz = \bp$ has a solution, and such a solution satisfies \eqref{MFCQ_z}. Hence, LICQ implies EGMFCQ. We further establish the connection between the constraint relaxation in \eqref{nsequ:4} and EGMFCQ.

\begin{theorem}\label{thm:1}
For any point $\bell\leq \bx \leq \bu$, if EGMFCQ holds at $\bx$, there exists a threshold $\bar{\theta}\in(0, 1]$ such that for any $\theta\in[0, \bar{\theta}]$,
\begin{equation*}
\Omega(\bx; \theta) \coloneqq \{\bz\in\mR^d: \theta\cdot \bc(\bx) + \nabla \bc(\bx)\bz = \0,\; \bell\leq \bx + \bz \leq \bu\} \neq \emptyset.
\end{equation*}
Conversely, suppose a sequence of points $\bell\leq \bx_k\leq \bu$, $\forall k\geq 0$, admits a sequence~of~"sharp"~$\theta_k\in(0, 1]$ in the sense that $\Omega(\bx_k, \theta_k)\neq \emptyset$ but $\Omega(\bx_k, \min\{2\theta_k,1\})= \emptyset$.~If there exists a subsequence $\{k_l\}_l$ such that $\lim\limits_{l\rightarrow\infty}\theta_{k_l}=0$, then EGMFCQ fails to hold at any accumulation point of $\{\bx_{k_l}\}_l$.
\end{theorem}

Theorem \ref{thm:1} suggests that the condition of EGMFCQ ensures the existence of $\theta_k > 0$~that makes the linearized constraints \eqref{nsequ:4} feasible, i.e., $\Omega(\bx_k;\theta_k)\neq \emptyset$. 
Note that EGMFCQ or~even LICQ is indeed widely imposed in constrained optimization problems, which essentially characterizes the quality of linear approximation to nonlinear constraints \citep{Nocedal2006Numerical}.
Based on Theorem \ref{thm:1}, we can design the algorithm with constraint relaxation in the following way: at each step, we initialize $\theta_k=1$ and backtrack reducing $\theta_k$ until $\Omega(\bx_k;\theta_k)\neq \emptyset$.~Note~that checking $\Omega(\bx_k;\theta_k)\neq \emptyset$ can be efficiently done by solving a convex quadratic program:
\begin{equation}\label{equ:relax:convex}
\min_{\bz\in\mR^d}\;\; \|\theta_k\bc_k + \nabla \bc_k \bz\|^2 \quad \quad  \text{s.t.}\quad \quad \bell\leq \bx_k+\bz\leq\bu.
\end{equation}
We have $\Omega(\bx_k;\theta_k)\neq \emptyset$ if and only if the optimal value of the above program is 0.
Furthermore, if the selected sequence $\{\theta_k\}$ is not bounded away from zero, Theorem \ref{thm:1} suggests that a subsequence of $\{\bx_k\}$ will converge to a point that fails EGMFCQ, and certainly fails LICQ. In that case, we should re-initiate the algorithm to explore other local solutions, since any estimation procedures with asymptotic normality, either offline $M$-estimator or online projection-based~estimator, target only solutions that satisfy LICQ \citep{Duchi2021Asymptotic, Davis2024Asymptotic}.$\quad\;\;$

\subsection{Primal-dual asymptotic optimality}

In this section, we explore the primal-dual asymptotic minimax optimality of Problem \eqref{original_problem0}. In particular, we target a KKT triplet $(\tx,\tlambda,\tmu)$ satisfying \eqref{nsequ:KKT} and investigate what is the~best estimator one can expect in the sense of H\'ajek and Le Cam. To that end, we introduce some~assumptions and notations beforehand. 

\begin{assumption}[Strict Complementarity]\label{ass:1}
We assume $[\tmu_1]_{\mA_\bell(\tx)}>\0$ and $[\tmu_2]_{\mA_\bu(\tx)}>\0$.
\end{assumption}

\begin{assumption}[Second-Order Sufficient Condition]\label{ass:2}
The Lagrangian Hessian with respect to $\bx$, $\nabla_{\bx}^2\mL^\star = \nabla_{\bx}^2\mL(\tx, \tlambda, \tmu)$, is positive definite in the null space of active constraint Jacobian $\{\bz\in\mR^d: \bJ(\tx)\bz = \0\}$. In other words, there exists $\omega>0$ such that
\begin{equation*}
\bz^\top\nabla_{\bx}^2\mL^\star \bz\geq \omega\|\bz\|^2\quad\quad \text{ for all }\; \bz\in \{\bz\in\mR^d: \bJ(\tx)\bz = \0\}.
\end{equation*}
\end{assumption}

Assumptions \ref{ass:1} and \ref{ass:2}, together with LICQ at $\tx$, are standard in the sensitivity analysis of constrained nonlinear optimization \citep{Shapiro1990differential, Bonnans2000Perturbation}, and are~also imposed for minimax optimality analysis of the primal solution $\tx$ in the literature \citep{Duchi2021Asymptotic, Davis2024Asymptotic}. These assumptions essentially require that the KKT triplet $(\tx,\tlambda,\tmu)$ is a strict local solution of \eqref{original_problem0}, admitting a local convexity structure in the sense~that the Lagrangian is strongly convex in the tangent space of the constraints manifold.

For any $\bx\in\mR^d$ and $\bmu = (\bmu_1, \bmu_2)\in\mR^{2d}$, let us denote active dual components by $\bmu_{\mA(\bx)} = ([\bmu_1]_{\mA_\bell(\bx)}, [\bmu_2]_{\mA_\bu(\bx)})\in\mR^{|\mA_\bell(\bx)|+|\mA_\bu(\bx)|}$ and inactive dual components by~$\bmu_{\mA(\bx)^-}= ([\bmu_1]_{\mA_\bell(\bx)^-},$  
$[\bmu_2]_{\mA_\bu(\bx)^-})\in\mR^{2d-|\mA_\bell(\bx)|-|\mA_\bu(\bx)|}$.~For notational simplicity, we write \mbox{$\tmu_{\mA^\star} = \tmu_{\mA(\tx)}$}~and~note~that $\tmu_{(\mA^\star)^-} = \tmu_{\mA(\tx)^-} = \0$ by the KKT conditions in \eqref{nsequ:KKT}.~We also define the Lagrangian \mbox{Hessian}~matrix with respect to $(\bx,\blambda,\bmu_{\mA(\bx)})$, a covariance matrix, and a sandwich matrix as
\begin{equation}\label{def:cov}
\bm{H}^{\star} =  \begin{pmatrix}
\nabla_{\bx}^2\mL^\star	& (\bJ^\star)^\top\\
\bJ^\star & \0
\end{pmatrix},\quad \bSigma^\star = \begin{pmatrix}
\text{Cov}(\nabla F(\tx;\zeta)) & \0\\
\0 & \0
\end{pmatrix},\quad \bm{\Omega}^{\star} = \left(\bm{H}^{\star}\right)^{-1}\bm{\Sigma}^\star \left(\bm{H}^{\star}\right)^{-1}.
\end{equation}
The non-singularity of $\bm{H}^{\star}$ is ensured by LICQ (Definition \ref{def_MFCQ}) and SOSC (Assumption \ref{ass:2}).

The lower-bound analysis closely relates to the sensitivity analysis of Problem \eqref{original_problem0} with~respect to the distribution $\P$. We will explore how the primal-dual solution as a function of $\P$, $(\tx(\P), \tlambda(\P),\tmu(\P))$, varies when $\P$ is perturbed slightly. As such, we measure the perturbation size via the $\phi$-divergence, defined as
\begin{equation*}
D_{\phi}(\mathcal{P}^{\prime}, \mathcal{P}) = \int \phi\left(\frac{d \mathcal{P}^{\prime}}{d \mathcal{P}}\right) d \mathcal{P},
\end{equation*}
where $\phi: (0, \infty) \to \mathbb{R}$ is a $C^3$-smooth convex function satisfying $\phi(1) = 0$. We define an admissible neighborhood $\mathcal{B}(\mathcal{P}; \epsilon)$ of $\mathcal{P}$ to consist of all probability distributions  $\mathcal{P}^{\prime}$ with $D_{\phi}(\mathcal{P}^{\prime}, \mathcal{P}) \leq \epsilon$, such that there exists a primal-dual solution $(\tx(\P'),\tlambda(\P'),\tmu(\P'))$ to the perturbed Problem \eqref{original_problem0}. It is well known that the inactive constraint set $\mA(\tx(\P))^-$ is preserved under small perturbations (cf. \cite{Shapiro1990differential} Theorem 1). That is, $\mA_\bell(\tx(\P^\prime))^- = \mA_\bell(\tx(\P))^-$~and~$\mA_\bu(\tx(\P^\prime))^- \\ = \mA_\bu(\tx(\P))^-$ for $\epsilon$ small enough; and we trivially have $\tmu_{(\mA^\star)^-}(\P) = [\tmu(\P)]_{\mA(\tx(\P))^-} = \0$ in the neighborhood of $\P$. Below, we focus on the nontrivial part $\tw(\P) = (\tx(\P), \tlambda(\P),\tmu_{\mA^\star}(\P))$.

The next theorem asymptotically lower bounds the performance of any estimation procedure for finding the solution to \eqref{original_problem0}, by considering adversarially chosen small perturbations of the unperturbed distribution $\P$.

\begin{theorem}[Local primal-dual minimax optimality]\label{thm:2}

\hskip-4pt Suppose Assumptions \ref{ass:1}, \ref{ass:2},~and~LICQ hold at $(\tx,\tlambda,\tmu)$.
Let $\ell: \mathbb{R}^{d+m+|\mA_\bell^\star|+|\mA_\bu^\star|} \to [0,\infty)$ be any symmetric, quasiconvex, and~lower semicontinuous function, and let $\bw_k(\P'): \zeta_{1:k}\rightarrow \mathbb{R}^{d+m+|\mA_\bell^\star|+|\mA_\bu^\star|}$ be a sequence of estimators based on $k$ samples $\zeta_{1:k} \stackrel{iid}{\sim} \P'$. Then, the following inequality holds:
\begin{equation*}
\lim_{c \to +\infty} \liminf_{k \to +\infty} \sup_{\mathcal{P}^{\prime} \in \mathcal{B}(\mathcal{P}; c/k)} \mathbb{E}_{\mathcal{P}^{\prime}_{k}} \left[\ell\left(\sqrt{k}\left(\bw_k(\P') - \tw(\P')\right)\right) \right] \geq \mathbb{E}\left[\ell\left(Z\right) \right],
\end{equation*}
where $\mE_{\P'_{k}}[\cdot]$ is taken over $k$ iid samples $\zeta_{1:k}$ and $Z \sim \mathcal{N}\left(\bm{0}, \bm{\Omega}^{\star}\right)$.
\end{theorem}

Compared with \citep[Theorem 3.2]{Davis2024Asymptotic} and \citep[Theorem 1]{Duchi2021Asymptotic}, our result jointly captures both the primal solution and the active dual solution (since inactive dual solution remains zero). Notably, following a similar calculation as in \citep[Remark 5.9]{Na2025Statistical}, we observe that the marginal covariance of $\tOmega$ with respect to the primal solution matches those in the above works (i.e., \eqref{M-estimator}). Thus, Theorem \ref{thm:2} provides a complete~and sharp characterization of local asymptotic optimality for constrained estimation problems.$\quad\quad$ 

\vspace{-0.2cm}

\section{SSQP with Polyak's Momentum}\label{sec:3}

In this section, we introduce the proposed online SSQP estimation procedure for Problem~\eqref{original_problem0}, which leverages the constraint relaxation idea discussed in Section~\ref{sec:2.1} and employs gradient (and Hessian) moving average schemes based on Polyak's momentum to debias the step direction.$\quad$

\subsection{Estimation procedure}\label{sec:3.1}

The estimation procedure involves tuning parameters $\tau\in(0,1)$, $\psi\geq 0$, $p\geq 1$, and one stepsize and two averaging weight sequences $\alpha_k, \beta_k, \gamma_k>0$.~Given the current $k$-th primal-dual iterate $(\bx_k,\blambda_k, \bmu_k)$ with $\bmu_k=(\bmu_{1,k}, \bmu_{2,k})$, we initialize $\theta_k=1$ and perform
\begin{equation}\label{equ:select:theta}
\theta_k \leftarrow \tau\theta_k\quad\quad \text{ until } \quad \quad \text{convex problem } \eqref{equ:relax:convex} \text{ has an optimal value } 0.
\end{equation}
We reasonably expect the above scheme to terminate in finite time by Theorem \ref{thm:1}. In particular, if $\theta_k$ falls below an infinitesimal threshold, our theory implies that $\bx_k$ is approaching~an undesirable point at which even EGMFCQ (let alone LICQ) does not hold.

With the selected $\theta_k$ from \eqref{equ:select:theta}, we then draw a single sample $\zeta_k\sim \P$ and estimate the~objective gradient and Hessian as $\nabla F(\bx_k;\zeta_k)$ and $\nabla^2 F(\bx_k;\zeta_k)$, respectively. Instead of using~these noisy estimates directly, we perform gradient and Hessian moving averages:
\begin{equation}\label{equ:gradHess:average}
\barg_k = (1-\beta_k)\barg_{k-1} + \beta_k\nabla F(\bx_k;\zeta_k)\quad\quad \text{ and }\quad\quad \barQ_k = (1-\gamma_k)\barQ_{k-1} + \gamma_k\nabla^2 F(\bx_k;\zeta_k).
\end{equation}
The Lagrangian Hessian with respect to $\bx$ depends on the constraint curvature information, which we estimate as
\begin{equation}\label{equ:Hess}
\barB_k =  \barQ_k + \sum_{i=1}^{m}[\blambda_k]_i\nabla^2\bc_i(\bx_k) + \bDelta_k,
\end{equation}
where $\bDelta_k\in\mR^{d\times d}$ is a Hessian modification term. In fact, if only the global convergence\footnote{\spacingset{1}\footnotesize\hskip-0.25cm In our context of nonlinear problems, global convergence refers to convergence to a stationary point from \textit{any} initialization, as opposed to convergence to a global solution, which is generally unachievable~\mbox{without}~specific problem structures. That said, these two notions coincide in the case of convex problems.} of the estimation procedure is of interest, we do not require any second-order information and can simply set $\barB_k = \bI$. As such, the importance of the two averaging weight sequences, $\beta_k$ and $\gamma_k$, is different: the choice of $\beta_k$ is more delicate than that of $\gamma_k$, since first-order information plays a more crucial role in debiasing the step and facilitating asymptotic normality (cf.~Theorem~\ref{thm:5}).

With the above estimators, we construct an SSQP subproblem similar to \eqref{nsequ:1}, but using~averaged gradient (and Hessian) estimators along with the constraint relaxation:
\begin{equation}\label{equ:SQP:new}
\min_{\barDelta\bx_k\in\mR^d}\; \barg_k^\top\barDelta\bx_k + \frac{1}{2}\barDelta\bx_k^\top\barB_k\barDelta\bx_k,\;\; \text{s.t.}\;\; \theta_k\bc_k + \nabla \bc_k\barDelta\bx_k = \0,\;\;  \bm{\ell} \leq \bx_k+ \barDelta\bx_k\leq \bm{u}.
\end{equation}
Since solving a nonconvex inequality-constrained quadratic program can be NP-hard \citep{Burer2009Nonconvex}, we require the Hessian modification $\bDelta_k$ to convexify $\barQ_k + \sum_{i=1}^{m}[\blambda_k]_i\nabla^2\bc_i(\bx_k)$ 
such that $\barB_k$ is both upper and lower bounded (cf.~Assumption \ref{assump2}). Thus, we can solve \eqref{equ:SQP:new}. We slightly abuse notation by letting $\barDelta\bx_k$ denote the primal solution of \eqref{equ:SQP:new}, and $(\barblambda_k^{\text{sub}}, \barbmu_k^\text{sub})$ with $\barbmu_k^\text{sub} = (\barbmu_{1,k}^\text{sub}, \barbmu_{2,k}^\text{sub})$ denote any (one of) dual solutions. We then update the iterate~as~follows:$\quad$
\begin{equation}\label{equ:update}
(\bx_{k+1}, \blambda_{k+1}, \bmu_{k+1}) = (\bx_k, \blambda_k, \bmu_k) + \baralpha_k\cdot (\barDelta\bx_k, \barblambda_k^{\text{sub}} - \blambda_k, \barbmu_k^\text{sub}-\bmu_k),
\end{equation}
where $\baralpha_k>0$ is an adaptive, potentially random stepsize that satisfies a safeguard condition: $\alpha_k\leq \baralpha_k\leq \alpha_k + \psi\alpha_k^p$ for the tuning parameters $\psi>0$ and $p\geq 1$.

\begin{remark}[Uniqueness of subproblem dual solution]\label{rem:1}

We should emphasize that the dual~solution $(\barblambda_k^{\text{sub}}, \barbmu_k^\text{sub})$ of the strongly convex SSQP subproblem \eqref{equ:SQP:new} is not necessarily unique, as we do not impose strong constraint qualifications. In fact, in global convergence analysis, when evaluating the KKT residual at the iterate $\bx_k$, we do not rely directly on the dual iterates $(\blambda_k, \bmu_k)$ or $(\barblambda_k^{\text{sub}}, \barbmu_k^\text{sub})$. Note that the KKT condition is defined solely in terms of the primal variable (see \eqref{nsequ:KKT}); therefore, we only desire the existence of dual multipliers, possibly not unique, under which the KKT residual converges to zero. In other words, even though the dual variables~of~individual SSQP subproblems may not be unique, the global convergence of the KKT residual with~respect to the primal iterates can still be established.

On the other hand, our local analysis in Section \ref{sec:4} targets a local solution $\tx$ at which LICQ holds, as required by minimax optimality guarantee in Section \ref{sec:2}. In that case, we can examine~the local convergence behavior of the dual variables $(\blambda_k, \bmu_k)$, since the SSQP subproblem admits~a unique dual solution $(\barblambda_k^{\text{sub}}, \barbmu_k^\text{sub})$ under LICQ. 
In particular, as $\bx_k$ approaches $\bx^{\star}$ satisfying~LICQ, the inactive constraints at $\bx^{\star}$ for the original problem \eqref{original_problem0} remain inactive at $\bx_k+\barDelta\bx_k$ for the SSQP subproblem \eqref{equ:SQP:new}, implying \eqref{equ:SQP:new} has a unique dual solution $(\barblambda_k^{\text{sub}}, \barbmu_k^\text{sub})$.~Therefore,~our~local convergence analysis on the dual variable is well justified.

\end{remark}

\begin{remark}[Adaptive stepsize]\label{rem:2}

Compared to online projection-based methods \citep{Duchi2021Asymptotic, Davis2024Asymptotic}, we allow using a random adaptive stepsize $\baralpha_k$, as long as $\baralpha_k$~that may depend on the step direction is well controlled within the interval $[\alpha_k, \alpha_k+\psi\alpha_k^p]$. Here,~the interval length $\psi\alpha_k^p$ measures the degree of adaptivity. Setting $\psi=0$ corresponds to deterministic stepsize.~The adaptive designs have been proposed for SSQP schemes in numerical~optimization literature \citep{Berahas2021Sequential, Fang2024Fully, Curtis2024Sequential, Na2025Statistical}, where the stepsize is chosen via either line-search or trust-region strategies. All the proposed stepsize rules in those works satisfy the safeguard condition. Motivated by promising empirical performance of adaptive methods, we adopt a similar design of stepsize in our estimation~procedure. This additional flexibility, however, makes the statistical inference analysis more involved, as it requires controlling the effect of adaptivity on the limiting distribution.

\end{remark}

\begin{remark}[Bias in step direction]\label{rem:3}

\hskip-4pt Compared to SSQP designs for equality-constrained problems, a crucial challenge posed by inequality constraints is the bias in the step direction estimation.
\end{remark}

\vskip-0.1cm
\begin{wrapfigure}[6]{l}{0.39\textwidth}
\vskip-0.8cm\hskip-0.2cm
\begin{tikzpicture}
\node at (0,0) {\includegraphics[width=0.37\textwidth,height=0.165\textwidth,trim={2.9cm 3.6cm 1.5cm 3.8cm},clip]{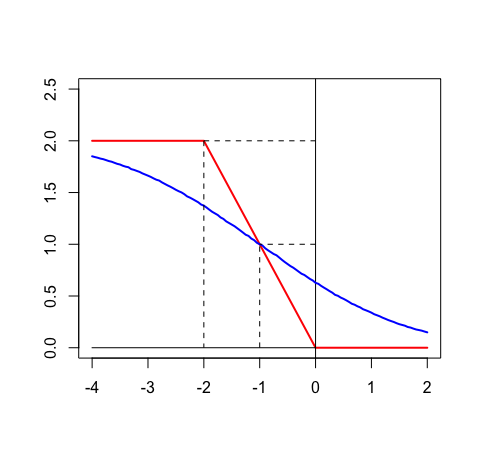}};
\node[font=\small] at (2.8,-1.55) {$\nabla f_k$};
\node[font=\small] at (1.05,-1.55) {$l=0$};
\node[font=\small] at (-1.05,-1.55) {$-u$};
\node[font=\small] at (-0.1,-1.55) {$-u/2$};
\node[font=\small] at (1.15,0.85) {$u$};
\node[font=\small] at (1.35,-.15) {$u/2$};
\node[red, font=\small] at (-2.65,1.2) {$\Delta x_k$};
\node[font=\small] at (2.3,-0.6) {${\blue \mE[ \barDelta x_k\mid x_k]}$};
\end{tikzpicture}
\vskip-0.4cm
\caption{\textit{Illustration of bias in step~direction estimation.}}\label{fig:2}
\end{wrapfigure}
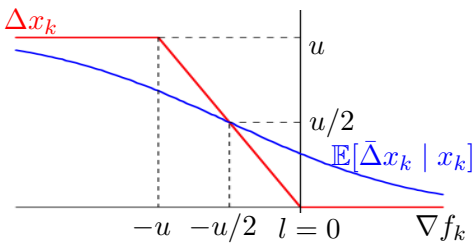
\noindent \textit{\hskip-2pt Consider the following example on $\mR$, where we let $\bar{B}_k=1$ and suppress equality constraints in \eqref{nsequ:1}: $\displaystyle\min_{\barDelta x_k \in \mR} 0.5\barDelta x_k^2 +\nabla F(x_k;\zeta_k) \barDelta x_k$ subject to $0 = l \leq \barDelta x_k \leq u$.~Without inequality constraints, this reduces to SGD with $\Delta x_k = -\nabla f_k = \mE[-\nabla F(x_k;\zeta_k) \mid x_k] = \mE[\barDelta x_k \mid x_k]$, as long as $\nabla F(x_k;\zeta_k)$ is unbiased.
However, with inequality constraints, we have $\Delta x_k = -\text{Project}(\nabla f_k, [-u, -l])$ shown as the red line in~Figure \ref{fig:2}, differing~from 
\begin{equation*}
\mE[\barDelta x_k \mid x_k] = -\mE[\text{Project}(\nabla F(x_k;\zeta_k), [-u, -l]) \mid x_k] \neq -\text{Project}(\mE[\nabla F(x_k;\zeta_k) \mid x_k], [-u, -l]),
\end{equation*}
shown as the blue line in Figure \ref{fig:2} with Gaussian noise $\nabla F(x_k;\zeta_k)\sim \N(\nabla f_k, \sigma^2)$. In other words, inequality constraints truncate the gradient noise in a nonlinear manner, such that an unbiased gradient estimate does not necessarily yield an unbiased step direction estimate. To~address~this issue, we leverage momentum-based averaging estimators in \eqref{equ:gradHess:average}. We can show that $\barDelta x_k$ with averaged estimators satisfies $\mE[\|\barDelta x_k - \Delta x_k\|^2]\rightarrow 0$ as $k\rightarrow \infty$, i.e., it is asymptotically unbiased (cf. Lemmas \ref{lemma20} and \ref{lemma19}). One significant contribution of this paper is to show that~gradient momentum preserves asymptotic optimality while debiasing step directions, as long as the weight sequence $\beta_k$ is chosen properly according to the stepsize $\alpha_k$.
}

\subsection{Global almost-sure convergence}\label{sec:3.2}

In this section, we establish global almost-sure convergence for the proposed SSQP estimation procedure in Section \ref{sec:3.1}. We show that the KKT residual of $\bx_k$ converges to zero from any~initialization. Our analysis generalizes existing SSQP literature on numerical optimization \citep{Na2023Inequality, Curtis2024Sequential} and relaxes restrictive constraint qualification conditions. $\quad\quad$

We define an $\ell_2$-penalized Lyapunov function to quantify the progress toward stationarity:
\begin{equation}\label{equ:varphi}
\varphi_\rho(\bx) = f(\bx) + \rho\|\bc(\bx)\|\quad\quad \text{ for } \; \rho>0.
\end{equation}
The penalty term biases the feasibility error of equality constraints, while inequality constraints are not penalized since SSQP iterates always stay within the box constraints (cf.~Lemma \ref{lem:2}).
In contrast to unconstrained problems, where methods aim to reduce the objective function $f$, reducing $f$ alone is not sufficient to justify the iteration progress toward stationarity in constrained problems, since a small objective value may be achieved by severely violating the constraints.
We will show that our method decreases the Lyapunov function \eqref{equ:varphi} for sufficiently~large~$\rho>0$, thereby vanishing the KKT residual, which is defined at the primal-dual triplet by$\hskip2cm$
\begin{equation}\label{equ:KKT}
\bR(\bx,\blambda,\bmu) = \left ( \begin{array}{c}
\nabla_{\bx}\mL(\bx,\blambda,\bmu) \\
\bc(\bx) \\
\bmu_1 \odot (\bell - \bx)\\
\bmu_2 \odot (\bx - \bu)
\end{array}\right).
\end{equation}
Here, $\odot$ denotes the Hadamard product and each component of $\bR(\bx, \blambda, \bmu)$ corresponds to one KKT condition in \eqref{nsequ:KKT}. Again, we do not measure the violation of the conditions $\bmu\geq 0$ and $\bell\leq \bx\leq \bu$ since they hold at every step. Thus, $\bR(\bx,\blambda,\bmu)=\0$ suggests we arrive~at~stationarity.\;\;\;

\subsubsection{Assumptions}

Now, we state the assumptions for global analysis. Assumption \ref{assump1} is tied to the constraint relaxation technique, which has been justified by Theorem \ref{thm:1} under EGMFCQ (weaker than~LICQ). Assumptions \ref{assump2} and \ref{assump3} are standard regularity conditions commonly imposed in the literature.

\begin{assumption}\label{assump1}
There exists a deterministic constant $\bar{\theta}\in(0, 1]$ such that the linearized~constraints \eqref{nsequ:4} are feasible for any relaxation parameter $\theta_k \in[0, \bar{\theta}]$. That is, $\Omega(\bx_k;\theta_k) \neq \emptyset$ for $\theta_k \in[0, \bar{\theta}]$.
\end{assumption}

Assumption \ref{assump1} requires the relaxation parameter $\theta_k$ to be bounded away from zero to ensure subproblem feasibility. By Theorem \ref{thm:1}, if $\theta_k$ is not bounded below, any accumulation point of the iterates $\bx_k$ fails to satisfy EGMFCQ and, certainly, LICQ. In that case, it becomes intractable to perform statistical inference for such an accumulation point, even using~\mbox{offline}~$M$-estimation or online projected methods, since asymptotic normality result in \eqref{M-estimator} of both methods, as well as primal-dual minimax optimality in Theorem \ref{thm:2}, crucially relies on LICQ.$\quad\quad$

\begin{assumption}\label{assump2}

We assume $\nabla f$, $\nabla \bc$ are Lipschitz continuous; that is, for all $\bm{\ell} \leq \bm{x}, \bm{y} \leq \bm{u}$, 
\begin{equation*}
\left\|\nabla f(\bm{x})-\nabla f(\bm{y}) \right\| \leq \kappa_{\nabla f} \left\|\bm{x} - \bm{y} \right\|, \quad\quad\quad \left\|\nabla \bm{c}(\bm{x})-\nabla \bm{c}(\bm{y}) \right\| \leq \kappa_{\nabla c} \left\|\bm{x} - \bm{y} \right\|
\end{equation*}
for some constants $\kappa_{\nabla f}, \kappa_{\nabla c}>0$.~Furthermore, the Hessian modification term $\bDelta_k\in\mR^{d\times d}$~in~\eqref{equ:Hess} ensures $\barB_k$ satisfy $\kappa_1 \bI \preceq \barB_k \preceq \kappa_2 \bI$ for some $0<\kappa_1 \leq \kappa_2$. In addition, we assume the dual multipliers of the SQP subproblem \eqref{equ:SQP:new} with the exact true gradient $\nabla f_{k}$, denoted by $(\blambda_k^{\text{sub}},  \bmu_k^{\text{sub}})$,~are bounded; that is, there exists $M_{\text{dual}}>0$ such that $\max\{\|\blambda_k^{\text{sub}}\|, \|\bmu_k^{\text{sub}}\|\} \leq M_{\text{dual}}$.
\end{assumption}

The Lipschitz condition is standard and naturally holds within a closed bounded feasible~set \citep{Berahas2023Stochastic, Curtis2024Sequential}. The condition on $\barB_k$ ensures the subproblem \eqref{equ:SQP:new} is strongly convex and admits a unique solution; this requirement can be readily relaxed to the SOSC condition, as stated in Assumption \ref{ass:2}, when a strong constraint qualification condition, e.g., LICQ, is imposed (see discussion after Assumption \ref{assump5} in~Section~\ref{sec:4}).~The~\mbox{boundedness}~of dual multipliers in (deterministic) SQP subproblems is a conventional assumption in constrained optimization. We show in Theorem \ref{lemma_mfcq} (Appendix \ref{Appendix:E}) that if the accumulation point of the iteration sequence $\bx_k$ is feasible and satisfies EGMFCQ, then the multipliers indeed remain bounded. This result provides a theoretical justification for the boundedness~assumption.$\quad$

\begin{assumption}\label{assump3}
We assume, for any $k\geq 0$, $\nabla F(\bx_k;\zeta_k)$ is an unbiased estimate of $\nabla f_k$ with bounded variance. In particular, $\mE_k[\nabla F(\bx_k;\zeta_k)] = \nabla f_k$ and $\mE_k[\|\nabla F(\bx_k;\zeta_k)- \nabla f_k\|^2]\leq \sigma_g^2$ for some $\sigma_g^2>0$. Here, $\mE_k[\cdot]$ denotes the conditional expectation given $\bx_k$.	
\end{assumption}

The unbiasedness and bounded variance of stochastic gradients are standard assumptions~in both unconstrained and constrained stochastic optimization \citep{Polyak1992Acceleration, Chen2020Statistical, Curtis2024Sequential, Na2025Statistical}.

\subsubsection{Almost-sure convergence}

The Lyapunov function $\varphi_{\rho}$ in \eqref{equ:varphi} serves as a surrogate of the original constrained problem,~providing a unified measure of objective value and constraint violation. Drawing intuition from~unconstrained optimization, a decrease in the Lyapunov function can be interpreted as progress~toward stationarity of constrained problem. Thus, we analyze the decrease of the Lyapunov function $\varphi_{\rho}(\bx)$ along the sequence of iterates. As such, we consider a local quadratic model~of~$\varphi_{\rho}(\bx)$:
\begin{equation}\label{nsequ:vap:loc}
\varphi_\rho^{\text{loc}}(\bx,\Delta\bx,\bm{B}) \coloneqq f(\bx) + \nabla f(\bx)^{\top} \Delta\bx + \frac{1}{2} \Delta\bx^{\top} \bm{B} \Delta\bx + \rho\left\|\bm{c}(\bx) + \nabla \bm{c}(\bx) \Delta\bx \right\|,
\end{equation}
and its difference measures the reduction of $\varphi_{\rho}$ at $\bx$ along the step $\Delta\bx$:
\begin{equation}\label{nsequ:vap:loc:diff}
\Delta \varphi_\rho^{\text{loc}}(\bx,\Delta\bx,\bm{B}) \coloneqq \varphi_\rho^{\text{loc}}(\bx,\0,\bm{B}) - \varphi_\rho^{\text{loc}}(\bx,\Delta\bx,\bm{B}).
\end{equation}

The next lemma establishes a reduction condition for the local model difference~$\Delta \varphi_\rho^{\text{loc}}(\bx,\Delta\bx,\bm{B})$, showing that 
the reduction at each step is proportional to both the step magnitude $\|\Delta\bx_k\|^2$~and the constraint violation $\|\bc_k\|$. Recall from Section \ref{sec:3.1} that $\barDelta\bx_k$ and $(\barblambda_{k}^{\text{sub}}, \barbmu_{k}^{\text{sub}})$ denote~the~primal and dual~solutions of the subproblem \eqref{equ:SQP:new}, while $\Delta\bx_k$ and $(\blambda_{k}^{\text{sub}}, \bmu_{k}^{\text{sub}})$ denote~their~counterparts when replacing gradient estimate $\barg_k$ by the true gradient $\nabla f_k$ in \eqref{equ:SQP:new}.

\begin{lemma}\label{lem:1}
Under Assumptions \ref{assump1} and \ref{assump2}, for any $\nu\in(0,1)$, as long as $\rho \geq \bar{\rho} \coloneqq \frac{M_{\text{dual}}}{(1-\nu) \tau \bar{\theta}}$,
\begin{equation*}
\Delta \varphi_\rho^{\text{loc}}(\bx_k,\Delta\bx_k,\barB_k) \geq \frac{1}{2} \Delta\bx_k^{\top} \barB_k \Delta\bx_k + \nu \rho \theta_k \left\| \bm{c}_{k}\right\| \quad \text{ for all }\; k\geq 0.
\end{equation*}
\end{lemma}

Since $\varphi_\rho^{\text{loc}}(\bm{x}, \Delta\bm{x}, \bm{B})$ closely approximates the Lyapunov function $\varphi_{\rho}(\bm{x})$, the above result implies a decrease in $\varphi_{\rho}$ itself. Consequently, the convergence of the iterates~toward a stationary point is closely linked to the vanishing of $\|\Delta\bx_k\|$ and $\|\bc_k\|$.

\begin{theorem}\label{thm:3}
Under Assumptions \ref{assump1}, \ref{assump2}, and \ref{assump3}, we specify the stepsize and \mbox{averaging}~weight sequences as $\alpha_k = \iota_1 (k+1)^{-b_1}$ and $\beta_k = \iota_2 (k+1)^{-b_2}$, for constants $\iota_1, \iota_2 >0$ and exponents~$b_1, b_2$ satisfying $b_1 \in (0.75,1]$ and $b_2 \in \left( 2-2b_1,2b_1-1\right)$. Then,
\begin{equation*}
\liminf_{k \to \infty} \cbr{\left\|\Delta\bx_k\right\| + \left\|\bm{c}_{k} \right\| } = 0\quad\quad \text{almost surely}.
\end{equation*}	
\end{theorem}

Finally, we establish the convergence of the KKT residual. Note that the KKT condition is defined in terms of the primal variable only (see \eqref{nsequ:KKT}); without the LICQ condition, the associated dual multipliers of Problem \eqref{original_problem0} are not unique. Thus, it suffices to establish the existence of (possibly nonunique) dual multipliers such that, together with $\bx_k$, the KKT residual converges~to zero.~As such, given $\bx_k$ of SSQP, we consider the multipliers defined by a least-squares problem:
\begin{equation}\label{est_opt_lagrangian}
\begin{split}
\min_{\blambda, \bmu}\quad & \|\nabla_{\bx}\mL(\bx_k,\blambda,\bmu)\|^2 + \| \bmu_{1} \odot (\bell- \bx_k)\|^2 + \| \bmu_{2} \odot (\bx_k - \bu) \|^2, \\
\text{s.t. }\;\; & \bm{\mu} \geq \bm{0}.
\end{split}
\end{equation}
We denote (one of) the optimal solution to~\eqref{est_opt_lagrangian} by $(\blambda_{k}^{\star}, \bmu_{k}^{\star})$. The next theorem shows that~the SSQP iterate $\bx_k$, coupled with $(\blambda_{k}^{\star}, \bmu_{k}^{\star})$, yields a vanishing KKT residual.

\begin{theorem}\label{thm:4}
Under the same conditions as in Theorem \ref{thm:3}, we have
\begin{equation*}
\lim_{k \to \infty} \bm{R}(\bx_k,\blambda_k^{\star},\bmu_k^{\star}) = \0 \quad\quad \text{almost surely}.
\end{equation*}
\end{theorem}

We note that in global analysis, two algorithmic components, the moving averaged second-order information of $\barB_k$ in \eqref{equ:Hess} and the dual update in \eqref{equ:update}, are not essential; one may simply set $\barB_k = \bI$ and suppress the dual update if only the consistency of the estimation is of interest. However, these components are crucial for local inference analysis, as they ensure the minimax optimality of the SSQP method.
On the other hand, our method involves two key sequences: $\alpha_k$ for the primal updates and $\beta_k$ for gradient averaging, which exhibit competing effects. In particular, a larger $\alpha_k$ accelerates the movement of the iterates but may impede the stabilization of the averaged gradients, while a smaller $\beta_k$ improves gradient accuracy but slows the adaptation of both the gradient sequence and the primal iterates. These coupled~\mbox{dynamics}~highlight the importance of carefully balancing the decay rates of $\alpha_k$ and $\beta_k$, as specified in Theorem~\ref{thm:3}, to ensure simultaneous convergence of the averaged gradients and the primal iterates.$\quad\;$

\section{Asymptotic Normality and Inference}\label{sec:4}

We now set the stage to present the local asymptotic analysis of our SSQP method in Section \ref{sec:3.1}.
We establish that the joint primal-dual variables, with proper stepsize control and~averaging weight sequences $\{\alpha_k, \beta_k, \gamma_k\}$, converge in distribution to a normal variable with limiting covariance $\bm{\Omega}^{\star}$ defined in \eqref{def:cov}.
The resulting limiting distribution attains the minimax~lower bound established in Theorem \ref{thm:2}, thereby demonstrating that the proposed method is asymptotically optimal in the sense of H\'ajek and Le Cam.~Furthermore, we provide a consistent plug-in estimator for the limiting covariance matrix, which enables practical online \mbox{statistical}~\mbox{inference},~including hypothesis testing and confidence interval construction.

\subsection{Asymptotic normality of the last iterate}

We first introduce a local assumption on the limiting point $\tx$ to which the sequence $\{\bx_k\}$ converges. This assumption is standard in both the lower-bound analysis (cf. Theorem \ref{thm:2})~and the upper-bound analyses of offline $M$-estimation and online projection-based methods \citep{Duchi2021Asymptotic, Davis2024Asymptotic}.

\begin{assumption}\label{assump5}
The iteration sequence $\bx_k$ converges almost surely to a strict local solution $\tx$, which satisfies: (1) LICQ in Definition \ref{def_MFCQ} (implying the uniqueness of the dual solution $(\blambda^{\star}, \bmu^{\star})$); (2) strict complementarity in Assumption \ref{ass:1}; and (3) second-order sufficient condition (SOSC) in Assumption \ref{ass:2}.
\end{assumption}

With Assumption \ref{assump5}, we now elaborate on the role of the Hessian matrix $\barB_k$. Note that the global analysis does not rely on any second-order information; hence, one may simply set~$\barB_k = \bI$ in the SSQP subproblem \eqref{equ:SQP:new} and all results in Section \ref{sec:3.2} still hold.
In contrast, for the local analysis, where $\bx_k$ is sufficiently close to a strict local solution $\tx$ that satisfies LICQ and SOSC, we expect $\barB_k$ to serve as a good approximation to the Lagrangian Hessian $\nabla^2_{\bx}\mL_k$, which motivates the Hessian averaging update in \eqref{equ:gradHess:average}. In fact, the conditions of LICQ and SOSC~at~$\tx$~extend to a local neighborhood. Thus, as long as $\bx_k$ is close to $\tx$ (which is the case) and $\barB_k$ is~close to $\nabla_{\bx}^2\mL_k$, these conditions apply to the SSQP subproblem \eqref{equ:SQP:new} and further imply the \textit{local} uniqueness of its primal-dual solution. 
Since no constraint qualification is imposed at~$\bx_k$~globally, the Hessian information in the subproblem is not helpful for determining its solutions in the global analysis. 
Note that Hessian approximation in the local analysis is necessary even for deterministic (quasi-)second-order methods, so called the Dennis–Mor\'e condition; see \cite{Dennis1974characterization}, \citep[Chapter 18]{Nocedal2006Numerical}, and \citep[Section~5]{Liu2000Robust}~for~details.

To precisely state the above argument, let us consider any $\epsilon>0$ such that
\begin{equation}\label{equ:eps}
0<\epsilon \leq 0.5\min\cbr{[\bu-\bell]_i, [\tx-\bell]_j, [\bu-\tx]_j: i\in \mA^\star, j\in (\mA^\star)^-},
\end{equation}
where we recall from Section \ref{sec:2} that $\mA^\star = \mA_\bell^\star\cup\mA_\bu^\star$ denotes the index set of active inequality constraints and $(\mA^\star)^{-}=[d]\setminus\mA^\star$ denotes its inactive complement. We define two sets at~$\bx_k$, $\quad\;$
\begin{equation}\label{equ:Akeps}
\mA_{\bell,k}(\epsilon) = \{i: \left[\bx_{k} - \bell\right]_{i} \leq \epsilon\} \quad\quad\text{ and }\quad\quad \mA_{\bu,k}(\epsilon) = \{i: \left[\bu - \bx_{k}\right]_{i} \leq \epsilon\},
\end{equation}
as our guess of true active sets $\mA_\bell^\star$ and $\mA_\bu^\star$, and let $\mA_{k}(\epsilon) \coloneqq \mA_{\bell,k}(\epsilon) \cup \mA_{\bu,k}(\epsilon)$. We will show next that $\mA_{\bell,k}(\epsilon)=\mA_\bell^\star$, $\mA_{\bu,k}(\epsilon)=\mA_\bu^\star$, and both also coincide with the active set of SSQP subproblem \eqref{equ:SQP:new} when $k$ is large enough, indicating that our method successfully identifies the true active set. 
Now, since SOSC holds at $\tx$ (cf. Assumption \ref{ass:2}), and given $\mA_{\bell,k}(\epsilon)$ and~$\mA_{\bu,k}(\epsilon)$ in \eqref{equ:Akeps}, we require the regularizer $\bDelta_k$ to convexify the estimate of $\nabla_{\bx}^2\mL_k$, i.e., $\barQ_k + \sum_{i=1}^{m}[\blambda_k]_i\nabla^2\bc_i(\bx_k)$ in \eqref{equ:Hess}, to ensure $\bz^\top\barB_k\bz \geq \omega \|\bz\|^2$, $\forall \bz\in \{\bz: \bJ_k(\epsilon)\bz = \0\}$, where $\bJ_k(\epsilon)=(\nabla\bc_{k}; -\bI_{\mA_{\bell,k}(\epsilon)}; \bI_{\mA_{\bu,k}(\epsilon)})$ is active constraints Jacobian.
The following lemma shows that, under this construction of $\barB_k$, the SSQP subproblem admits a \textit{unique} solution in a neighborhood of $\bm{0}$ when $\bx_{k}$ is close to $\bx^{\star}$. In particular, if we fail to solve the subproblem \eqref{equ:SQP:new} with the above $\barB_k$, we should turn~to~safely~regularize $\barB_{k}$ to be positive definite to ensure the subproblem well-defined.

\begin{lemma}\label{lemma_almost_convg}
Under Assumptions \ref{assump3} and \ref{assump5}, we let $\epsilon$ satisfy \eqref{equ:eps} and $\beta_k = \iota_2(k+1)^{-b_2}$ satisfy $\iota_2>0$, $b_2\in(0.5,1]$. For any run of the method, there exists a (potential random)~threshold $K^\star<\infty$ such that for all $k\geq K^\star$,
\begin{enumerate}[label=(\alph*),topsep=0pt]
\setlength\itemsep{0.0em}
\item The relaxation parameter satisfies $\theta_k=1$.
\item The estimated active sets satisfy  $\mA_{\bell,k}(\epsilon) = \mA_{\bell}^{\star}$ and $\mA_{\bu,k}(\epsilon) = \mA_{\bu}^{\star}$. 
\item The subproblems \eqref{equ:SQP:new} with both the averaged gradient $\barg_k$ and true gradient $\nabla f_k$~\mbox{admit}~unique local solutions in a neighborhood of $\0$; furthermore, $\barDelta\bx_{k} \to \bm{0}$ and $\Delta\bx_{k} \to \bm{0}$ as $k\rightarrow\infty$.
\item The active sets identified by the subproblems satisfy $\mA_\bell(\bx_k + \barDelta\bx_k) = \mA_\bell(\bx_k + \Delta\bx_k) = \mA_\bell^{\star}$ and $\mA_\bu(\bx_k + \barDelta\bx_k) = \mA_\bu(\bx_k + \Delta\bx_k) = \mA_\bu^{\star}$.
\end{enumerate}

\end{lemma}

We additionally emphasize that, by Lemma \ref{lemma_almost_convg}(a) and under LICQ at $\tx$, the relaxation~parameter $\theta_k$ eventually equals one so that the linearized constraints become locally exact. We~next explore almost-sure convergence of the dual iterates $(\blambda_k,\bmu_k)$ and the Hessian approximation~$\barB_k$ of the SSQP method. For the latter purpose, we require the stochastic Hessian estimates to~be unbiased and have bounded variance.

\begin{assumption}\label{assump7}
We assume, for any $k\geq 0$, $\nabla^2 F(\bx_k;\zeta_k)$ is an unbiased estimate of $\nabla^2 f_k$ with bounded variance. In particular, $\mE_k[\nabla^2 F(\bx_k;\zeta_k)] = \nabla^2 f_k$ and $\mE_k[\|\nabla^2 F(\bx_k;\zeta_k)- \nabla^2 f_k\|^2]\leq \sigma_H^2$ for some $\sigma_H^2>0$. Here, $\mE_k[\cdot]$ denotes the conditional expectation given $\bx_k$.	
\end{assumption}

\begin{lemma}\label{lemma_almost_eq}
Under Assumptions \ref{assump3} and \ref{assump5}, let $\alpha_k=\iota_1(k+1)^{-b_1}$ and $\beta_k = \iota_2(k+1)^{-b_2}$~satisfy $\iota_1, \iota_2>0$, $b_1\in(0,1]$, $b_2\in(0.5,1]$.~Then, we have $(\bm{x}_k, \bm{\lambda}_{k},  \bm{\mu}_{k}) \to (\bm{x}^{\star}, \bm{\lambda}^{\star}, \bm{\mu}^{\star})$ as $k\rightarrow\infty$~almost surely. Furthermore, suppose Assumption \ref{assump7} holds and let the averaging weight sequence of the Hessian in \eqref{equ:gradHess:average} satisfy $\gamma_{k}=\iota_{3} (k+1)^{-b_3}$ with $\iota_3 > 0$, $b_3 \in (0.5,1]$. Then, we also have~$\Bar{\bB}_{k} \to \nabla_{\bx}^2\mathcal{L}^\star$ as $k\rightarrow\infty$ almost surely.
\end{lemma}

Before stating normality result, we strengthen the bounded variance condition on stochastic gradients in Assumption \ref{assump3} to a bounded $(2+\delta)$ moment. For notational simplicity, we reuse the symbol $\sigma_g^2$ from Assumption \ref{assump3} to denote the upper bound of $(2+\delta)$ moment. 
This moment condition is mild and standard in existing literature on both unconstrained and constrained stochastic approximation methods \citep{Polyak1992Acceleration, Toulis2021Proximal, Chen2020Statistical, Chen2024Online, Zhu2021Online, Lee2022Fast, Duchi2021Asymptotic, Na2025Statistical}. Some methods may require even higher-order moments; for example, projected SGD in \cite{Davis2024Asymptotic} imposes a bounded fourth moment to establish normality.

\begin{assumption}\label{assump9}
We assume, for any $k\geq 0$, $\nabla F(\bx_{k};\zeta_{k})$ is unbiased and has bounded $(2+\delta)$ moment for some $\delta>0$. In particular, $\mathbb{E}_{k}[\|\nabla F(\bx_{k};\zeta_{k})-\nabla f_k\|^{2+\delta}]\leq \sigma_g^2$ for some \mbox{$\sigma_g^2>0$}.~Furthermore, we assume $\mE_k[\nabla F(\bx_k;\zeta_k)\nabla F(\bx_k;\zeta_k)^\top]\rightarrow \mE[\nabla F(\tx;\zeta)\nabla F(\tx;\zeta)^\top]$ as $\bx_k\rightarrow\tx$.
\end{assumption}

We are now ready to present the asymptotic normality result for the SSQP iterates.~Following the notation in the lower-bound analysis of Theorem \ref{thm:2}, we denote $\bw_{k} = (\bx_{k}, \blambda_{k},\left[\bmu_{k}\right]_{\mathcal{A}^{\star}})$ and $\bw^{\star} = (\bx^{\star},\blambda^{\star},\bmu^{\star}_{\mathcal{A}^{\star}})$ as the nontrivial part of the primal-dual solution, noting that $\bmu^{\star}_{(\mathcal{A}^{\star})^-} = \0$.

\begin{theorem}\label{thm:5}
Consider the SSQP method in Section \ref{sec:3.1} and suppose Assumptions \ref{assump5}, \ref{assump7},~\ref{assump9} hold. We specify the stepsize and averaging weight sequences $\alpha_k = \iota_1 (k+1)^{-b_1}$, $\beta_k = \iota_2 (k+1)^{-b_2}$, $\gamma_k = \iota_3 (k+1)^{-b_3}$, and the stepsize adaptivity gap parameter $p$ in \eqref{equ:update} to satisfy
\begin{equation*}
\max\cbr{0.5, \frac{2-2\delta}{2+\delta}}< b_1 \leq 1, \quad 0.5< b_2 < b_1,\quad 0.5< b_3 \leq 1,\quad p> 1.5-\frac{b_2}{2b_1},
\end{equation*}
and $\iota_1>2/3$ if $b_1=1$ and $\iota_3>0.25b_1$ if $b_3=1$. Then, we have (recall $\tOmega$ in~\eqref{def:cov})
\begin{equation*}
1/\sqrt{\baralpha_k}\cdot(\bw_{k} - \bw^{\star}) \stackrel{d}{\longrightarrow} \mathcal{N} \left( \bm{0}, \eta \cdot \bm{\Omega}^{\star} \right)\quad\quad \text{ with }\quad 
\eta =\begin{cases*}
0.5 & \text{if } $b_1 < 1$,\\
\iota_1/(2\iota_1-1) & \text{if } $b_1=1$.
\end{cases*}
\end{equation*}
For the inactive dual components, $[\bmu_{k}]_{(\mathcal{A}^{\star})^-}$ vanishes almost surely with a rate $\|[\bmu_{k}]_{(\mathcal{A}^{\star})^-}\| = o(k^{-b})$ for any $b>0$ if $b_1<1$ and $O(k^{-\iota_1})$ if $b_1=1$.
\end{theorem}

We consider a specific setup where $\alpha_k = (k+1)^{-1}$. The following corollary shows that~SSQP with this stepsize attains optimal $\sqrt{k}$-consistency with the limiting covariance $\bm{\Omega}^{\star}$ matching~the minimax lower bound in the sense of H\'ajek and Le Cam, as established in Theorem \ref{thm:2}.$\quad\quad\quad$

\begin{corollary}\label{cor:1}
Under Assumptions \ref{assump5}, \ref{assump7}, \ref{assump9}, we let $\iota_1=b_1=1$, $b_2, b_3 \in (0.5, 1)$, $\iota_2, \iota_3>0$, $p\geq 1.5$. Then, $\sqrt{k} (\bw_{k} - \bw^{\star}) \stackrel{d}{\longrightarrow} \mathcal{N} \left( \bm{0}, \bm{\Omega}^{\star} \right)$.
\end{corollary}

Theorem \ref{thm:5} establishes asymptotic normality for primal and active dual variables of SSQP under general setups of parameters. To our knowledge, this result illustrates that SSQP with gradient momentum is the \textit{first} method to attain joint primal-dual asymptotic optimality, extending prior works \cite{Duchi2021Asymptotic, Davis2024Asymptotic} that explore only primal~asymptotic optimality and rely on expensive projection operators. The main idea of SSQP is to~perform sequential linear-quadratic approximations to original nonlinearly constrained problem.$\quad\quad$

The primary technical challenge in our analysis arises from handling the gradient momentum. In particular, existing literature on asymptotic normality of first- and second-order stochastic approximation methods relies crucially on the conditional independence of stochastic gradients \citep{Polyak1992Acceleration, Toulis2021Proximal, Chen2020Statistical, Chen2024Online, Zhu2021Online, Lee2022Fast, Duchi2021Asymptotic, Leluc2023Asymptotic, Davis2024Asymptotic, Na2025Statistical}. In contrast, the averaged gradients in our setting are inherently dependent across iterations. 
As a result, our analysis must carefully examine the interplay between the stepsize sequence $\alpha_k$ and the averaging weight sequence $\beta_k$.
We show that achieving asymptotic normality requires the weights $\beta_k$ to decay more slowly than the stepsizes $\alpha_k$, i.e., $b_2 < b_1$. Although gradient momentum helps reduce stochastic noise and allows the method to mimic the behavior of its deterministic counterparts, the slowly decaying weights ensure that the method still effectively uses a gradient estimate $\barg_k$ that is close to the stochastic gradient $\nabla F(\bx_k;\zeta_k)$, thereby preserving asymptotically optimal behavior of the iterates. 
We also note that the choice of $b_1$ is always feasible for any $\delta > 0$, with larger values of $\delta$ permitting larger feasible ranges. When $\delta = 1$ (i.e., $\nabla F(\bx_k;\zeta_k)$ has a bounded third moment), we can take $b_1 \in (0.5, 1]$. In addition, the~adaptivity gap condition is satisfied as long as $p \geq 1.5$, and we remark that numerical literature often sets $p = 2$ \citep{Berahas2021Sequential}. Finally, the conditions on $\iota_1$, $\iota_3$ address the corner~cases when $b_1$ or $b_3 = 1$, and setting $\iota_1 = \iota_3 = 1$ is always sufficient.

\subsection{Practical inference: plug-in covariance estimation}

To perform online statistical inference based on Theorem \ref{thm:5}, we must also estimate the limiting covariance matrix $\tOmega$. Existing literature considers either plug-in estimators \citep{Chen2020Statistical, Na2025Statistical} or batch-means estimators \citep{Zhu2021Online, Kuang2025Online}. In this paper, we provide a simple plug-in estimator since the dominant computational cost of SSQP lies in solving the subproblem instead of covariance estimation. We also note that the~plug-in estimator does not require any additional gradient evaluations. Define
\begin{equation*}
\bH_k = \begin{pmatrix}
\bar{\bB}_{k} & \bJ_{k}^{\top} \\
\bJ_{k} & \0
\end{pmatrix}, \quad 
\bSigma_k = \begin{pmatrix}
\text{Cov}(\{\nabla F(\bx_i;\zeta_i)\}_{i=0}^k) & \0\\
\0 & \0
\end{pmatrix},\quad 
\bm{\Omega}_{k} = \bm{H}_{k}^{-1} \bm{\Sigma}_{k} \bm{H}_{k}^{-1},
\end{equation*}
where $\barB_k$ is from \eqref{equ:Hess}; $\bJ_k = (\nabla\bc_{k}; -\bI_{\mA_\bell(\bx_k + \barDelta\bx_k)}; \bI_{\mA_\bu(\bx_k + \barDelta\bx_k)})$ is the Jacobian of the active constraints with the active set identified by the SSQP subproblem (cf. Lemma \ref{lemma_almost_convg}); and the~sample covariance is defined as
\begin{multline*}
\text{Cov}(\{\nabla F(\bx_i;\zeta_i)\}_{i=0}^k) = \frac{1}{k+1}\sum_{i=0}^{k} \nabla F(\bx_{i};\zeta_{i}) \nabla F(\bx_{i};\zeta_{i})^{\top}  \\ - \rbr{\frac{1}{k+1}\sum_{i=0}^{k} \nabla F(\bx_{i};\zeta_{i})}\rbr{ \frac{1}{k+1}\sum_{i=0}^{k} \nabla F(\bx_{i};\zeta_{i})}^{\top}.
\end{multline*}
Note that $\bm{\Omega}_{k}$ can be updated recursively during the iterations, as it relies on the same stochastic gradient evaluations as the averaged gradient $\bar{\bg}_{k}$, namely $\{\nabla F(\bx_i;\zeta_i)\}_{i=0}^k$. Consequently, the plug-in estimator $\bm{\Omega}_{k}$ incurs negligible additional computational cost.

The following theorem establishes the almost-sure convergence of $\bOmega_k$. We strengthen the bounded $(2+\delta)$ moment condition of gradient estimates to bounded $4$-th moment, which~is~also standard for other plug-in covariance estimators \citep{Chen2020Statistical, Na2025Statistical}.$\quad$

\begin{theorem}\label{thm:6}
Under the conditions of Theorem \ref{thm:5} and further assuming $\delta\geq 2$ in Assumption \ref{assump9}, we have $\bm{\Omega}_{k} \to \bm{\Omega}^{\star}$ as $k\rightarrow\infty$ almost surely.
\end{theorem}

With the above theorem, we are now able to construct confidence intervals or regions for $\tw = (\tx, \tlambda, \tmu_{\mA^\star})$. For example, fixing a desired coverage probability $1 - q$ with $q \in (0,1)$,~the $100(1-q)\%$ confidence region of $\tw$ is given by
\begin{equation}\label{equ:coverage}
P(\tw \in \mC_{k, q}) \rightarrow 1-q\quad \text{ as }\quad k\rightarrow\infty,
\end{equation}
where $\mC_{k,q} = \{\bw: (\bw-\bw_k)^\top\bOmega_k^\dagger(\bw-\bw_k)/\baralpha_k\leq \chi^2_{d,1-q}\}$. Here, $\chi^2_{d,1-q}$ is the $(1-q)$-quantile of $\chi^2_d$ distribution and $\bOmega_k^\dagger$ is the pseudo-inverse of $\bOmega_k$. The result \eqref{equ:coverage} follows from the distribution of quadratic forms and the fact that $\operatorname{rank}(\tOmega) = d$ \citep[Corollary 1.3.6a]{Christensen2020Plane}.

\section{Experimental Studies}\label{sec:5}

In this section, we perform comprehensive experiments to demonstrate the effectiveness of the SSQP estimation procedure in Section \ref{sec:3.1}, validate asymptotic normality results, and illustrate its applicability to practical statistical inference tasks, such as performing hypothesis testing and constructing confidence intervals for the constrained model parameters. We evaluate the empirical performance of our method on a diverse set of tasks, including benchmark nonlinear constrained optimization problems from the CUTEst library \citep{Gould2014CUTEst, Fowkes2022PyCUTEst}, constrained regression problems involving linear, logistic, and Poisson models, as well as portfolio allocation problems. We study both synthetic data and real \texttt{Fama-French~Portfolios} and \texttt{Chicago Air Pollution} data.

We use the same set of hyper-parameters for all experiments. Specifically, we set $\tau = 0.5$, $\psi = 1$, and $p = 2$.
The stepsize control sequence is set to $\alpha_k = (k+1)^{-0.751}$, while~the~weight sequences for the gradient and Hessian are set to $\beta_k = (k+1)^{-0.501}$ and $\gamma_k = (k+1)^{-1}$,~respectively, which satisfy the conditions in Theorems~\ref{thm:4} and~\ref{thm:5}.
Our implementation code~is~provided in the public repository: \url{https://github.com/yihang-gao/SSQP}.

\subsection{CUTEst benchmark problems}\label{sec:5.1}

The CUTEst library collects numerous constrained problems that are widely used for benchmarking optimization methods. We apply our method to a subset of CUTEst problems and introduce stochastic perturbations to gradients and Hessians in order to simulate noisy environments. In particular, at each step, we let $\nabla F(\bx_k;\zeta_k)= \nabla f_k + \bE_{k,\nabla f}$ and $\nabla^2 F(\bx_k;\zeta_k) = \nabla^2 f_k + \bE_{k,\nabla^2 f}$, where the deterministic quantities $\nabla f_k$ and $\nabla^2 f_k$ are provided by the CUTEst package, and $\bE_{k,\nabla f}$ and $\bE_{k,\nabla^2 f}$ denote the noise vector and matrix, respectively. 
We study two types of~noise: light-tailed Gaussian noise, where each entry $\bE_{k,\nabla f}, \bE_{k,\nabla^2 f}$ follows $\N(0, \sigma^2)$,~and heavy-tailed $t$-distribution noise, where each entry $\bE_{k,\nabla f}, \bE_{k,\nabla^2 f}$ follows $t_{df}$. We vary $\sigma^2\in\{1, 10^{-1}, 10^{-2}, \\ 10^{-4}\}$ and $df\in\{3,4,5,9\}$. Note that the noise of $t_9$ closely approximates that of $\N(0,1)$.$\quad\quad\quad$

We compare our SSQP method with two state-of-the-art baselines: ActiveSet-SSQP \citep{Na2023Inequality} and Biased-SSQP \citep{Curtis2024Sequential}. The ActiveSet-SSQP method adaptively~increases the batch size to reduce the noise in the step direction, and is hence not an online method and is more sample-intensive.~In contrast, the Biased-SSQP method is an online method but lacks bias reduction through gradient momentum.

\noindent$\bullet$ \textbf{Global convergence.} For each problem instance, we run~$K = 10^5$ iterations; and for each method, we evaluate it using the KKT residual $\|\bR_K\|$ defined in~\eqref{equ:KKT} and the feasibility~error $\|\bc_K\|$. The performance comparison is shown in Figure~\ref{fig:3}. From the figure, we observe that Biased-SSQP exhibits noticeable deviation from the optimal solution, primarily due to the~bias in its step direction induced by the inequality constraints. ActiveSet-SSQP, on the other hand, performs robustly even under high noise levels but heavily depends on increasing batch size to ensure a sufficiently accurate step direction.
In contrast, our proposed method requires only a single sample per iteration to estimate both the gradient and Hessian, while effectively reducing stochastic noise and debiasing the step through gradient averaging. We see that when the noise level is moderate (Gaussian noise with $\sigma^2\in\{10^{-1}, 10^{-2}, 10^{-4}\}$),~our~method~achieves~performance comparable to, or even better than, ActiveSet-SSQP.

\vskip-0.1cm
{\spacingset{1.1}
\begin{figure}[h!]
\setlength{\subfigcapskip}{-0.15cm}
\centering
\subfigure[KKT residuals]{\includegraphics[width=0.4\textwidth]{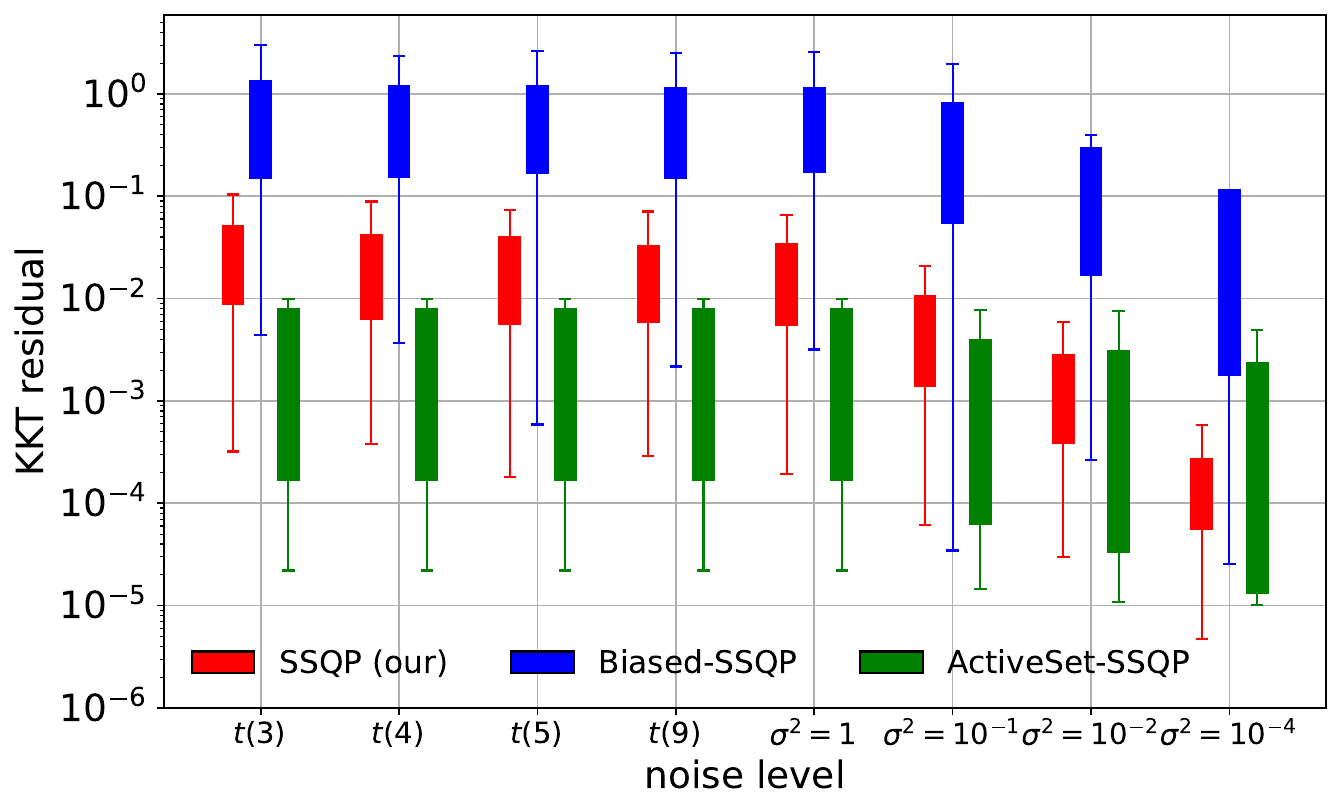}}\quad\quad
\subfigure[Feasibility errors]{\includegraphics[width=0.4\textwidth]{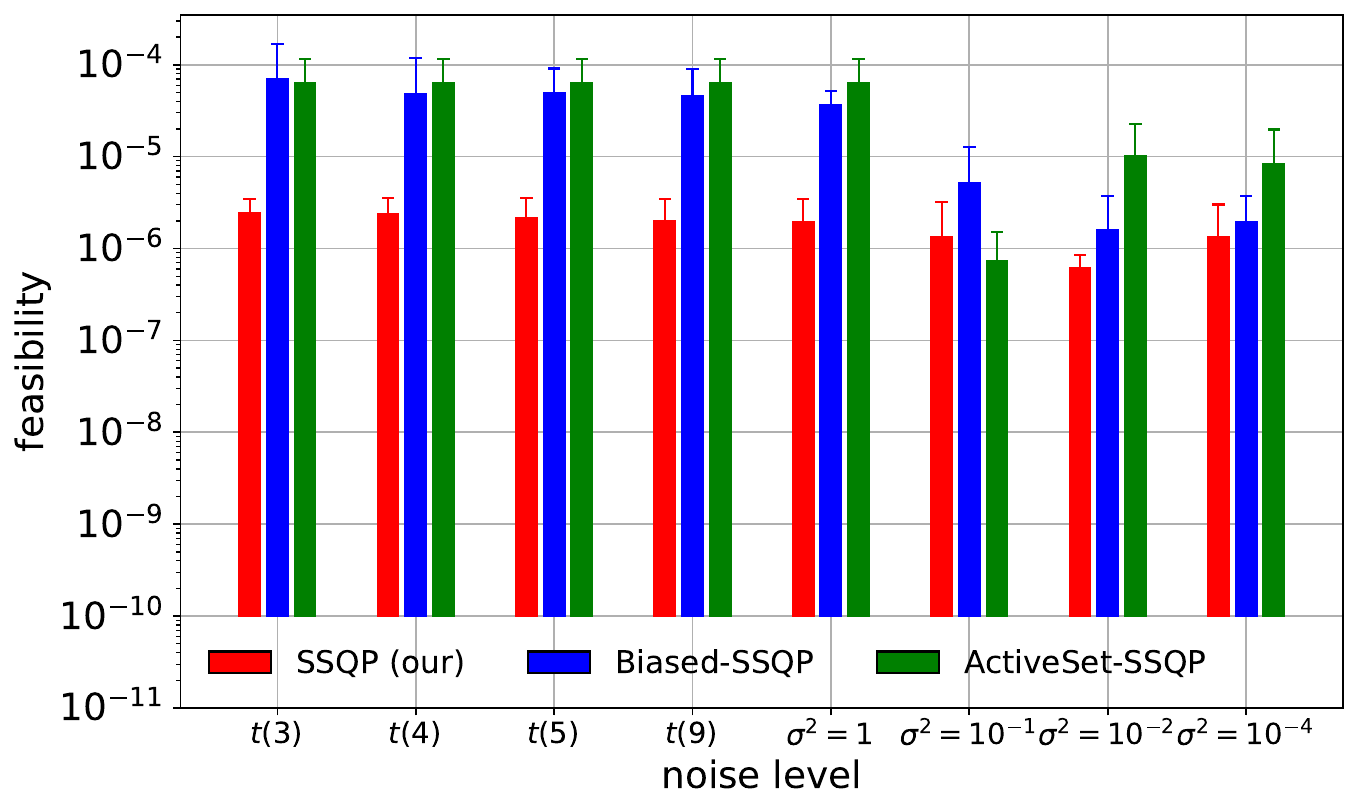}}
\vskip-0.4cm
\caption{\textit{Boxplots of KKT residuals and feasibility errors for CUTEst problems. For each noise setting, three boxplots are shown, corresponding to the SSQP (ours), Biased-SSQP, and ActiveSet-SSQP methods.}}
\label{fig:3}\vskip-0.5cm
\end{figure}
}

\noindent$\bullet$ \textbf{Local normality.} 
We next examine local asymptotic normality of the SSQP iterates. 
For each problem, we estimate the averaged model parameter $\1^{\top}\bm{x}^{\star}/d$ and construct its nominal~$95\%$ confidence interval. The performance of the constructed intervals is evaluated in terms of their empirical coverage rate (CovRate) and average length (AvgLen), each averaged over 200 independent runs. Results of 5 HS-type CUTEst problems are reported in Table~\ref{tab:1}. We find that the confidence intervals produced by SSQP achieve empirical coverage probabilities that closely match the nominal $95\%$ level, thereby providing strong empirical support for the theoretical asymptotic normality guarantee established in Theorem~\ref{thm:5}. 
Furthermore, the average length of the confidence intervals increases as the noise level grows, a behavior also consistent with our theoretical expectations, since the asymptotic covariance matrix $\bm{\Omega}^{\star}$ is proportional to $\mathrm{Cov}\!\left(\nabla F(\bm{x}^{\star}; \zeta)\right)$ as defined in \eqref{def:cov}.

{\spacingset{1.1}
\begin{table}[h]
\scriptsize
\centering
\caption{\textit{The coverage rate (CovRate) and length of confidence intervals (AvgLen) for 5 HS-type CUTEst constrained problems. The standard deviation of the interval length is also reported.}}
\label{tab:1}
\begin{tabular}{ccccccc}
\hline\\[-4pt]
\multirow{2}{*}{\textbf{Problem}} & \textbf{Noise Level} & \multicolumn{2}{c}{\textbf{Gaussian}} & \textbf{Noise Level}  & \multicolumn{2}{c}{\textbf{Student t}} \\[2pt]
&($\sigma^2$) & \textbf{CovRate}(\%) & \textbf{AvgLen} & $(df)$ & \textbf{CovRate}(\%) & \textbf{AvgLen}\\[2pt]
\hline\\[-4pt]
\multirow{5}{*}{\textbf{HS41}} & 1 & 97.0 & 2.50E-2 (7.73E-4) & 3 & 86.0 & 3.77E-2 (1.90E-3) \\[2pt]
& $10^{-1}$ & 97.5 & 7.59E-3 (7.03E-5) & 4 & 93.0 & 3.06E-2 (1.28E-3) \\[2pt]
& $10^{-2}$ & 97.0 & 2.40E-3 (8.69E-6) & 5 & 94.0 & 2.79E-2 (9.25E-4)\\[2pt]
& $10^{-4}$ & 97.5 & 2.40E-4 (5.95E-7)& 9 & 97.0 & 2.45E-2 (7.10E-4)\\[2pt]
\hline\\[-4pt]
\multirow{5}{*}{\textbf{HS65}} & 1 & 94.5 & 1.87E-3 (6.82E-6) & 3 & 96.5 & 3.18E-3 (1.66E-4)\\[2pt]
& $10^{-1}$ & 94.5 & 5.92E-4 (1.59E-6) & 4 & 95.0 & 2.59E-3 (1.72E-5)\\[2pt]
& $10^{-2}$ & 95.0 & 1.87E-4 (4.96E-7) & 5 & 95.0 & 2.37E-3 (1.11E-5)\\[2pt]
& $10^{-4}$ & 94.5 & 1.87E-5 (4.97E-8) & 9 & 94.5 & 2.08E-3 (8.36E-6)\\[2pt]
\hline\\[-4pt]
\multirow{5}{*}{\textbf{HS68}} & 1 & 97.0 & 2.31E-1 (4.85E-2) & 3 & 95.5 & 3.00E-1 (1.32E-1)\\[2pt]
& $10^{-1}$ & 98.0 & 5.09E-2 (2.33E-3) & 4 & 94.5 & 2.08E-1 (6.09E-2) \\[2pt]
& $10^{-2}$ & 98.5 & 1.58E-2 (2.23E-4) & 5 & 95.0 & 1.81E-1 (5.07E-2)\\[2pt]
& $10^{-4}$ & 95.5 & 1.58E-3 (4.56E-6) & 9 & 94.5 & 1.48E-1 (3.52E-2) \\[2pt]
\hline\\[-4pt]
\multirow{5}{*}{\textbf{HS71}} & 1 & 97.0 & 1.95E-3 (1.44E-5) & 3 & 94.0 & 3.34E-3 (1.23E-4)\\[2pt]
& $10^{-1}$ & 96.5 &  6.17E-4 (1.93E-6) & 4 & 96.0 &  2.74E-3 (6.79E-3)\\[2pt]
& $10^{-2}$ & 96.5 & 1.95E-4 (5.20E-7) & 5 & 96.5 & 2.49E-3 (2.51E-5)\\[2pt]
& $10^{-4}$ & 98.5 & 1.95E-5 (5.08E-8) & 9 & 95.0 & 2.19E-3 (2.12E-5) \\[2pt]
\hline\\[-4pt]
\multirow{5}{*}{\textbf{HS81}} & 1 & 94.5 & 3.49E-2 (3.17E-3)  & 3 & 91.0 &  5.04E-2 (7.56E-3)\\[2pt]
& $10^{-1}$ & 97.0 & 1.13E-2 (4.77E-5) & 4 & 94.0 & 4.21E-2 (3.51E-3)\\[2pt]
& $10^{-2}$ & 98.0 &  3.58E-3 (9.63E-6) & 5 &  94.5 & 3.88E-2 (2.42E-3)\\[2pt]
& $10^{-4}$ & 98.0 & 3.59E-4 (9.22E-7) & 9 &  95.0 & 3.43E-2 (2.10E-3) \\[2pt]
\hline
\end{tabular}\vskip-0.4cm
\end{table}
}

\noindent$\bullet$ \textbf{Effectiveness of gradient averaging.} 
Empirically, our online method consistently outperforms the other online Biased-SSQP method. This improvement is primarily attributed to the momentum-style gradient moving-average scheme: as iterations proceed, the averaged~gradient progressively approximates the true population gradient, enabling the method to emulate~the~behavior of deterministic SQP while maintaining full stochasticity. In contrast, Biased-SSQP updates its iterates using noisy gradients, leading to inherent biases in step estimates that~accumulate over time and result in deviations from the optimal solution.
As shown in Figure \ref{fig:4}, we visualize the difference between the averaged gradients and the true gradients across~iterations. The results 
align with our theory that averaged gradients (solid lines) progressively~approach the true gradients;  while non-averaged gradients (dashed lines) exhibit constant variance.$\quad$

\vskip-0.1cm
{\spacingset{1.1}
\begin{figure}[h!]
\centering
\setlength{\subfigcapskip}{-0.15cm}
\subfigure[HS32]{\includegraphics[width=0.4\textwidth]{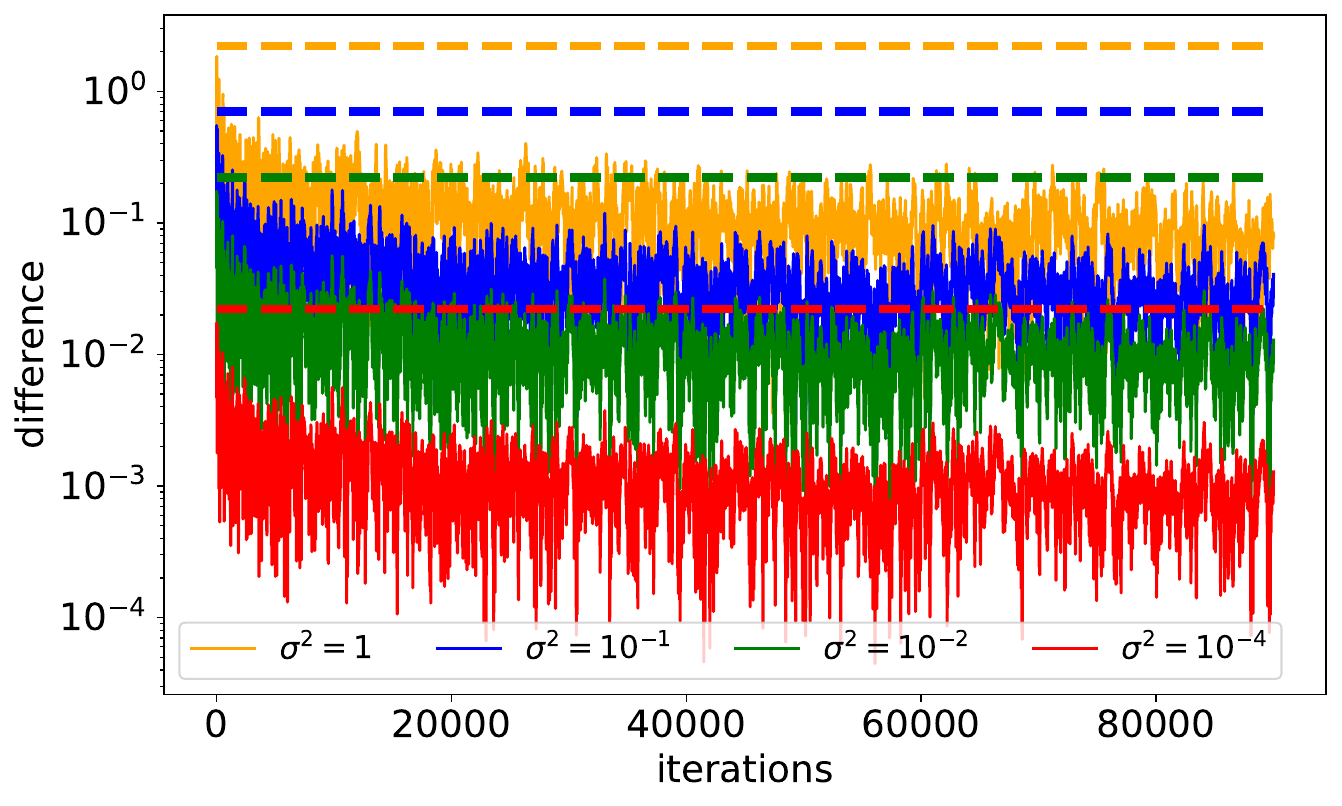}\label{fig:avg_grad1}}\quad\quad
\subfigure[FCCU]{\includegraphics[width=0.4\textwidth]{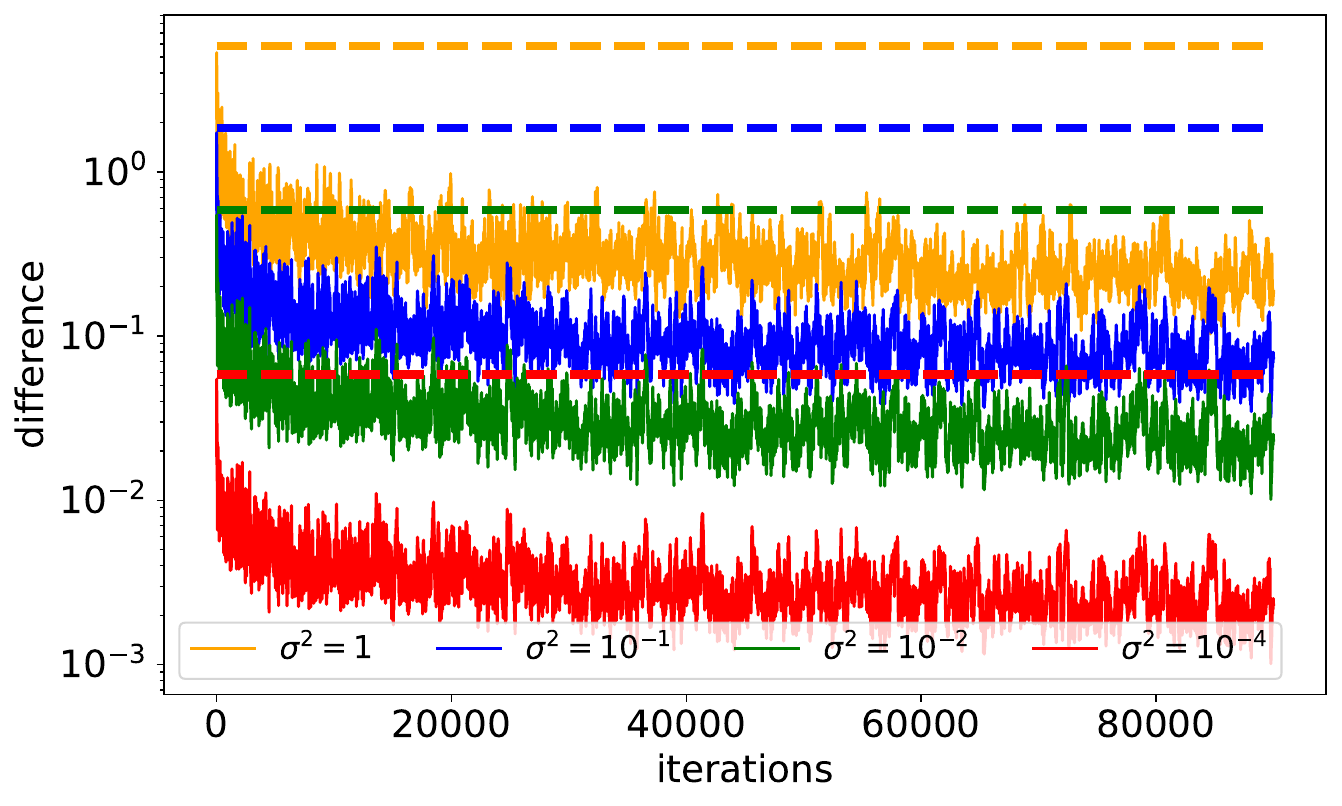}\label{fig:avg_grad2}}
\vskip-0.4cm
\caption{\textit{Differences between the averaged gradients and the true gradients on HS32 and FCCU problems. Solid lines: trajectories of gradient difference between the averaged gradients and the exact gradients during iterations, i.e., $\left\| \Bar{\bm{g}}_k -\nabla f_{k}\right\|$. Dashed lines: expected error without averaging, i.e., $\mathbb{E} \left[\left\| \nabla F(\bx_k;\zeta_k) -\nabla f_{k}\right\|\mid \bx_k \right]$.}}	\vskip-0.5cm
\label{fig:4}
\end{figure} 
}

\subsection{Constrained regression problems}\label{sec:5.2}

We further implement our method on constrained regression problems, including both linear and logistic regression models. In this study, each sample is denoted by the covariate-response pair $\zeta_k = (\zeta_{k}^{\ba}, \zeta_{k}^{b})$, and the response is generated based on different regression models. In particular, for linear regression, we let $\zeta_{k}^{b} = \zeta_{k}^{\ba \top} \tx + \varepsilon_k$ with $\varepsilon_{k}\stackrel{iid}{\sim}\N(0,1)$ and set $F(\bm{x}; \zeta_k) = \tfrac{1}{2}(\zeta_{k}^{b} - \zeta_{k}^{\ba \top} \bm{x})^2$. For logistic regression, we let $P (\zeta_{k}^{b} \mid \zeta_{k}^{\ba}) = 1/\{1 + \exp (-\zeta_{k}^{b} \cdot \zeta_{k}^{\ba \top}\tx)\}$ with $\zeta_{k}^{b} \in \{-1, 1\}$ and set $F(\bm{x}; \zeta_k) = \log (1 + \exp (-\zeta_{k}^{b} \cdot \zeta_{k}^{\ba \top}\bm{x}))$. 
For both models, we set the true model parameters~as~$\tx = \left(\frac{3}{2d},\ldots,\frac{3}{2d},\frac{1}{2d},\ldots,\frac{1}{2d}\right)$, where the first half of the components~are~$\frac{3}{2d}$~and the second half are $\frac{1}{2d}$. 
We then impose the probability simplex constraints $\Omega \coloneqq \{\bm{x} \in \mathbb{R}^d : \bm{1}^{\top}\bm{x} = 1,\, \bm{x} \geq \bm{0}\}$ in the estimation procedure. We generate covariate vector $\zeta_k^{\ba}$ from $\zeta_k^{\ba} \sim \N (\bm{\mu}^{\ba}, \bm{\Sigma}^{\ba})$,~where~$\bmu^\ba = (1,\ldots,1,-1,\ldots,-1)$. Following unconstrained regression settings \citep{Chen2020Statistical, Zhu2021Online}, we consider three structures for $\bSigma^\ba$: (i) Identity: $\bm{\Sigma}^\ba = \bm{I}$; (ii) Toeplitz: $[\bSigma^\ba]_{ij} = r^{|i-j|}$ with $r\in\{0.4,0.5,0.6\}$; and (iii) Equi-correlation: $[\bSigma^\ba]_{ij} = r$ for $i\neq j$ and $[\bSigma^\ba]_{ii}=1$ with $r\in\{0.1,0.2,0.3\}$. For each setting of $\bSigma^\ba$, we vary the problem dimension $d\in\{5,10,20,30\}$ and construct $95\%$ confidence interval for $\bmu^{\ba \top}\tx$, i.e., the parameters difference of two groups.$\quad\quad$

We measure the performance by repeating each experiment 200 times with varying random seeds and reporting the empirical coverage rate (CovRate) and average length (AvgLen) of the constructed confidence intervals. The results are summarized in Tables \ref{tab:2} and \ref{tab:3}.
From the tables, we observe that the empirical coverage probabilities are close to the nominal $95\%$ level across all settings, thereby providing strong empirical support for our local asymptotic~\mbox{normality}~guarantees and demonstrating the practical effectiveness of our method. Furthermore, the average length of the confidence intervals remains on the order of $10^{-2}$, consistent with the findings reported by \cite{Chen2020Statistical, Zhu2021Online, Na2025Statistical}. The low standard~deviation of the interval lengths relative to their means further indicates the robustness of the~proposed inference procedure across different covariance structures.

{\spacingset{1.1}
\begin{table}[h]
\centering
\caption{\textit{The coverage rate (CovRate) and length of confidence intervals (AvgLen) for constrained linear regression problems. The standard deviation of the interval length is also reported.}}
\label{tab:2}
\resizebox{\textwidth}{!}{
\begin{tabular}{ccccccc}
\hline\\[-6pt]
\textbf{Cov Matrix} & \textbf{Dim $d$} & \textbf{CovRate}(\%) & \textbf{AvgLen} & \textbf{Dim $d$}  & \textbf{CovRate}(\%) & \textbf{AvgLen} \\[2pt]
\hline\\[-6pt]
\multirow{2}{*}{\textbf{Identity}} & 5 & 93.5 & 3.73E-2 (1.74E-4) &  20 & 92.5 & 4.00E-2 (1.33E-4) \\[2pt]
& 10 & 96.5 & 3.91E-2 (1.47E-4) & 30 & 92.5 & 4.03E-2 (1.53E-4)\\[2pt]
\hline\\[-6pt]
\multirow{2}{*}{\textbf{Toeplitz} $(r=0.4)$} & 5 & 94.0 & 3.71E-2 (1.68E-4) & 20 & 96.0 & 3.93E-2 (1.38E-4) \\[2pt]
& 10 & 94.5 & 3.82E-2 (1.62E-4) & 30 & 93.0 & 3.98E-2 (1.52E-4) \\[2pt]
\hline\\[-6pt]
\multirow{2}{*}{\textbf{Toeplitz} $(r=0.5)$} & 5 & 94.0 & 3.74E-2 (1.67E-4) & 20 & 96.0 & 3.91E-2 (1.38E-4) \\[2pt]
& 10 & 95.5 & 3.82E-2 (1.60E-4)& 30 & 93.0 & 3.95E-2 (1.61E-4) \\[2pt]
\hline\\[-6pt]
\multirow{2}{*}{\textbf{Toeplitz} $(r=0.6)$} & 5 & 94.5 & 3.78E-2 (1.70E-4) &  20 & 96.5 & 3.90E-2 (1.36E-4) \\[2pt]
& 10 & 94.5 & 3.83E-2 (1.68E-4) & 30 & 93.5 & 3.94E-2 (1.60E-4) \\[2pt]
\hline\\[-6pt]
\multirow{2}{*}{\textbf{EquiCorr} $(r=0.1)$} & 5 & 93.5 & 3.76E-2 (1.58E-4) &  20 & 94.0 & 4.01E-2 (1.35E-4)\\[2pt]
&  10 & 93.0 & 3.92E-2 (1.40E-4) & 30 & 92.5 & 4.05E-2 (1.56E-4) \\[2pt]
\hline\\[-6pt]
\multirow{2}{*}{\textbf{EquiCorr} $(r=0.2)$} & 5 & 92.5 & 3.79E-2 (1.59E-4) &  20 & 93.5 & 4.02E-2 (1.26E-4) \\[2pt]
& 10 & 95.0 & 3.94E-2 (1.50E-4)  & 30 & 96.0 & 4.05E-2 (1.44E-4)\\[2pt]
\hline\\[-6pt]
\multirow{2}{*}{\textbf{EquiCorr} $(r=0.3)$} & 5 & 92.5 & 3.83E-2 (1.65E-4) &  20 & 93.0 & 4.03E-2 (1.31E-4) \\[2pt]
& 10 & 95.0 & 3.96E-2 (1.46E-4) & 30 & 93.5 & 4.05E-2 (1.49E-4) \\[2pt]
\hline\\[-6pt]
\end{tabular}}\vskip-0.6cm
\end{table}
}

{\spacingset{1.1}
\begin{table}[h]
\centering
\caption{\textit{The coverage rate (CovRate) and length of confidence intervals (AvgLen) for constrained logistic regression problems. The standard deviation of the interval length is also reported.}}\label{tab:3}
\resizebox{\textwidth}{!}{
\begin{tabular}{ccccccc}
\hline\\[-6pt]
\textbf{Cov Matrix} & \textbf{Dim} $d$ & \textbf{CovRate}(\%) & \textbf{AvgLen} & \textbf{Dim} $d$  & \textbf{CovRate}(\%) & \textbf{AvgLen} \\[2pt]
\hline\\[-6pt]
\multirow{2}{*}{\textbf{Identity}} & 5 & 96.5 & 4.46E-2 (7.97E-5) & 20 & 94.5 & 5.87E-2 (7.13E-5) \\[2pt]
& 10 & 94.5 & 5.87E-2 (7.13E-5) & 30 & 93.0 & 7.34E-2 (7.90E-5) \\[2pt]
\hline\\[-6pt]
\multirow{2}{*}{\textbf{Toeplitz} $(r=0.4)$} & 5 & 94.5 & 4.46E-2 (9.06E-5) & 20 & 92.5 & 6.86E-2 (1.01E-4) \\[2pt]
& 10 & 95.5 & 5.83E-2 (8.59E-5) & 30 & 93.5 & 7.30E-2 (1.13E-4) \\[2pt]
\hline\\[-6pt]
\multirow{2}{*}{\textbf{Toeplitz} $(r=0.5)$} & 5 & 95.0 & 4.46E-2 (8.91E-5) & 20 & 94.0 & 6.84E-2 (1.08E-4) \\[2pt]
& 10 & 94.5 & 5.83E-2 (8.77E-5) & 30 & 93.0 & 7.28E-2 (1.24E-4) \\[2pt]
\hline\\[-6pt]
\multirow{2}{*}{\textbf{Toeplitz} $(r=0.6)$} & 5 & 94.5 & 4.47E-2 (9.63E-5) & 20 & 92.5 & 6.82E-2 (1.19E-4) \\[2pt]
& 10 & 94.0 & 5.83E-2 (8.77E-5) & 30 & 94.5 & 7.26E-2 (1.32E-4)\\[2pt]
\hline\\[-6pt]
\multirow{2}{*}{\textbf{EquiCorr} $(r=0.1)$} & 5 & 95.0 & 4.47E-2 (9.22E-5) & 20 & 93.0 & 6.69E-2 (9.40E-5) \\[2pt]
& 10 & 94.0 & 5.89E-2 (7.81E-5) & 30 & 93.5 & 7.40E-2 (9.27E-5) \\[2pt]
\hline\\[-6pt]
\multirow{2}{*}{\textbf{EquiCorr} $(r=0.2)$} & 5 & 96.0 & 4.47E-2 (8.86E-4) & 20 & 95.0 & 7.00E-2 (1.05E-4) \\[2pt]
& 10 & 95.0 & 5.92E-2 (7.32E-5) & 30 & 92.5 & 7.46E-2 (1.02E-4) \\[2pt]
\hline\\[-6pt]
\multirow{2}{*}{\textbf{EquiCorr} $(r=0.3)$} & 5 & 95.0 & 4.48E-2 (8.59E-5) & 20 & 93.5 & 7.05E-2 (1.09E-4)  \\[2pt]
& 10 & 96.0 & 5.95E-2 (7.94E-4) & 30 & 94.5 & 7.52E-2 (1.09E-4) \\[2pt]
\hline\\[-6pt]
\end{tabular}}\vskip-0.2cm
\end{table}
}

\subsection{Portfolio allocation: Fama-French dataset}\label{sec:5.3}

We implement our method on portfolio allocation problems, using $30$ portfolios selected from the Fama–French Portfolios dataset, subject to the gross-exposure constraint \citep{Fan2007Variable, Fan2012Vast, Du2022High}: $\Omega \coloneqq \{\bx \in \mathbb{R}^d : \bm{1}^{\top}\bm{x} = 1,\, \|\bm{x}\|_{1} \leq c \}$, 
where we set $c = 3$ and $\bm{x}$ denotes the portfolio weight vector; a negative weight signifies shorting an asset. 
Let $\bmu^\ba, \bm{\Sigma}^\ba$ denote~the mean vector and covariance matrix of the asset returns $\zeta^\ba\in\mR^{30}$. We consider four widely studied portfolio models: (i) Global Minimum Variance \textbf{(GMV)}: $\displaystyle\min_{\bx \in \Omega} \bm{x}^{\top}\bm{\Sigma}^\ba\bm{x}$; (ii) Mean-Variance \textbf{(MV)}: $\displaystyle\min_{\bx \in \Omega} -\bx^{\top}\bmu^\ba + \bx^{\top}\bSigma^\ba\bx$; (iii) Exponential Utility \textbf{(EXP)}: $\displaystyle \min_{\bm{x} \in \Omega} \mE[\exp (-\eta_1\bm{x}^{\top}\zeta^{\ba})]$, where $\eta_1>0$ is the risk-aversion parameter and set to $\eta_1 = 0.1$; and (iv) Logarithmic Utility \textbf{(LOG)}: $\displaystyle \min_{\bx \in \Omega} -\mE[\log (\bm{x}^{\top}\zeta^{\ba} + \eta_2)]$, where $\eta_2 > 0$ is a regularization parameter ensuring the validity~of~the logarithm and set to $\eta_2 = 15$.

For each month, we use historical daily data from the preceding year as training samples and apply both the SSQP method (dealing with one sample at a time) and the $M$-estimation method (using full samples to estimate $\bmu^\ba, \bm{\Sigma}^\ba$) to solve the resulting constrained stochastic problem and obtain a portfolio weight vector. This weight vector is then held fixed throughout the following month for evaluation. Specifically, the out-of-sample performance of the estimated portfolio weights is evaluated using four standard metrics computed over 30 months during the period 2021–2023: accumulative return, maximum drawdown, Sharpe ratio, and Sortino~ratio.
The accumulative return quantifies the overall gain or loss of a portfolio strategy, while the other three metrics balance return with risk aspect: the maximum drawdown measures the~largest observed decline from a peak to a trough; the Sharpe ratio relates the portfolio's mean return to its total risk measured by the standard deviation of returns; and the Sortino ratio refines this measure by considering only downside volatility, i.e., the standard deviation of negative~returns. The results are summarized in Table \ref{tab:4}.

{\spacingset{1.1}
\begin{table}[h]
\centering
\caption{\textit{The comparison of SSQP and $M$-estimation on Fama-French Portfolios Dataset from 2021-2023. For each metric, the best-performing model is shown in bold.}}
\label{tab:4}
\resizebox{0.9\textwidth}{!}{
\begin{tabular}{lcccc}
\hline\\[-6pt]
\textbf{Model} & \textbf{Return} (\%) & \textbf{Max Drawdown} & \textbf{Sharpe Ratio} & \textbf{Sortino Ratio} \\[2pt]
\hline\\[-6pt]
\textbf{Equal Weight} & 15.10 & \textbf{0.22} & 0.73 & 1.15 \\[2pt]
\hline\\[-6pt]
\textbf{GMV} (ours) & 34.94 & 0.27 & 2.81 & 4.28 \\[2pt]
\textbf{GMV} ($M$-est) & 33.43 & 0.27 & 2.71 & 4.14 \\[2pt]
\hline\\[-6pt]
\textbf{MV} (ours) & 42.21 & 0.28 & \textbf{3.36} & \textbf{5.09} \\[2pt]
\textbf{MV} ($M$-est) & 40.31 & 0.28 & \textbf{3.29} & \textbf{5.02} \\[2pt]
\hline\\[-6pt]
\textbf{EXP} (ours) & 52.50 & 0.32 & 2.60 & 3.98 \\[2pt]
\textbf{EXP} ($M$-est) & 51.85 & 0.31 & 2.55 & 3.86 \\[2pt]
\hline\\[-6pt]
\textbf{LOG} (ours) & \textbf{54.86} & 0.33 & 2.45 & 3.59 \\[2pt]
\textbf{LOG} ($M$-est) & \textbf{55.08} & 0.32 & 2.46 & 3.57\\[2pt]
\hline\\[-6pt]
\end{tabular}}\vskip-0.4cm
\end{table}
}

From Table \ref{tab:4}, we see that SSQP achieves comparable performance to offline $M$-estimation across all metrics. In particular, the performance differences across all four models and four metrics are within 0.15, with the largest observed in the global minimum variance model under the Sortino ratio metric, where SSQP outperforms $M$-estimation by 0.14.~Among \mbox{different}~\mbox{portfolio} models, we observe that the portfolio strategy based on the logarithmic utility model achieves the highest accumulative return, which is consistent with the empirical findings of \cite{Du2022High}. The mean-variance model yields higher Sharpe and Sortino ratios, while the simple equal-weight strategy yields the smallest maximum drawdown.

Figure \ref{fig:5} further visualizes the estimated weights of two representative stocks under two~portfolio models, exponential utility and logarithmic utility. 
From the figure, we observe a strong~correlation between the temporal adjustments of the weights and the dynamics of the corresponding stock returns. In particular, as the return decreases, the weight assigned to the stock also decreases. Moreover, an abrupt drop in the stock return is promptly followed by a widening blue band, indicating an increase in the estimated standard deviation of the associated weight. This behavior aligns with economic intuition and demonstrates the interpretability and \mbox{validity}~of~the estimated weights under constrained model estimation.

\vskip-0.1cm
{\spacingset{1.1}
\begin{figure}[ht]
\setlength{\subfigcapskip}{-0.15cm}
\centering
\subfigure[Stock 1, EXP]{\includegraphics[width=0.242\textwidth]{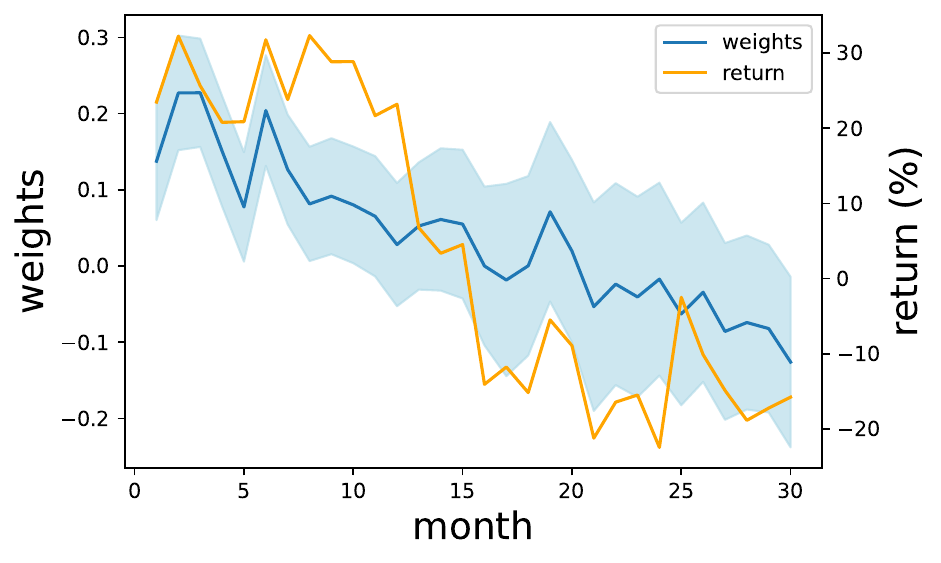}\label{exp1}}
\subfigure[Stock 2, EXP]{\includegraphics[width=0.242\textwidth]{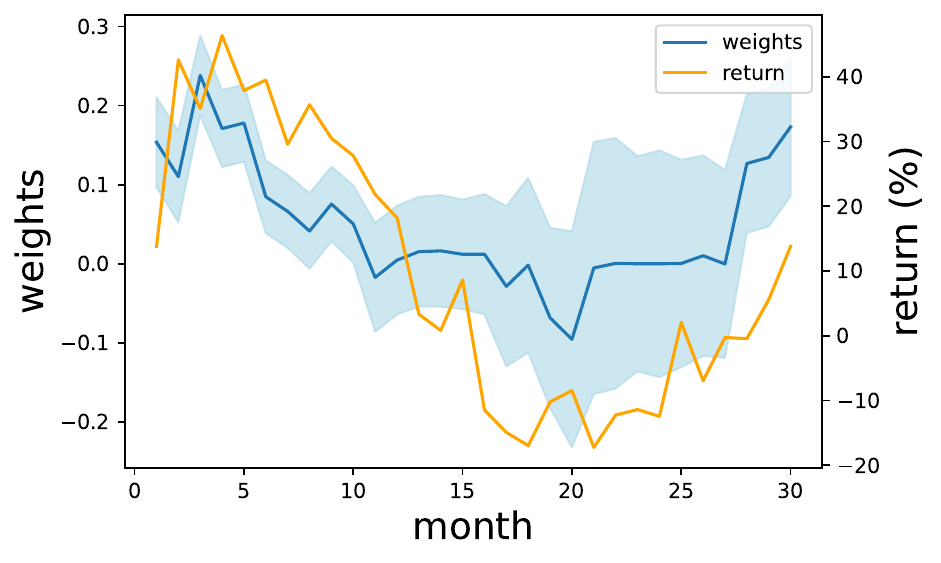}\label{exp2}}
\subfigure[Stock 1, LOG] {\includegraphics[width=0.242\textwidth]{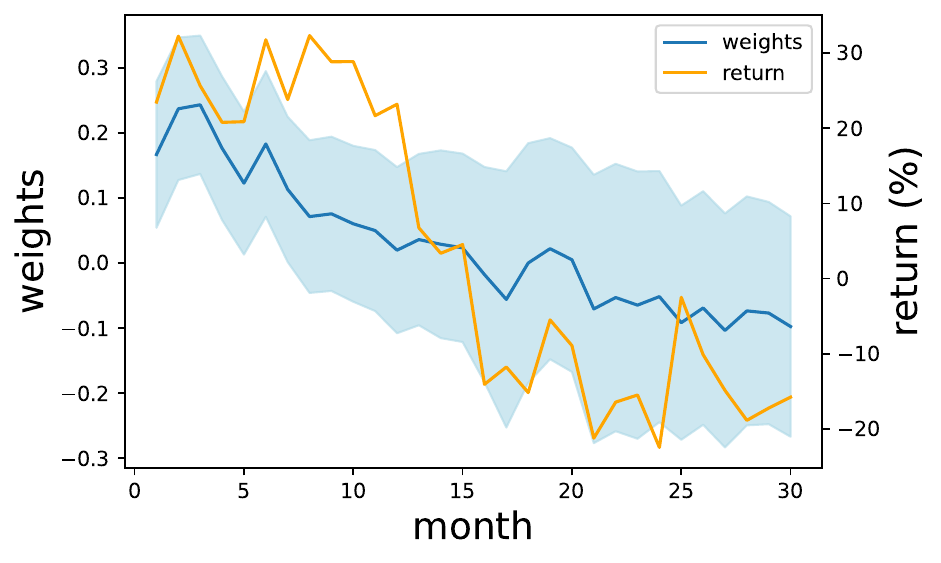}\label{log1}}
\subfigure[Stock 2, LOG]{		\includegraphics[width=0.242\textwidth]{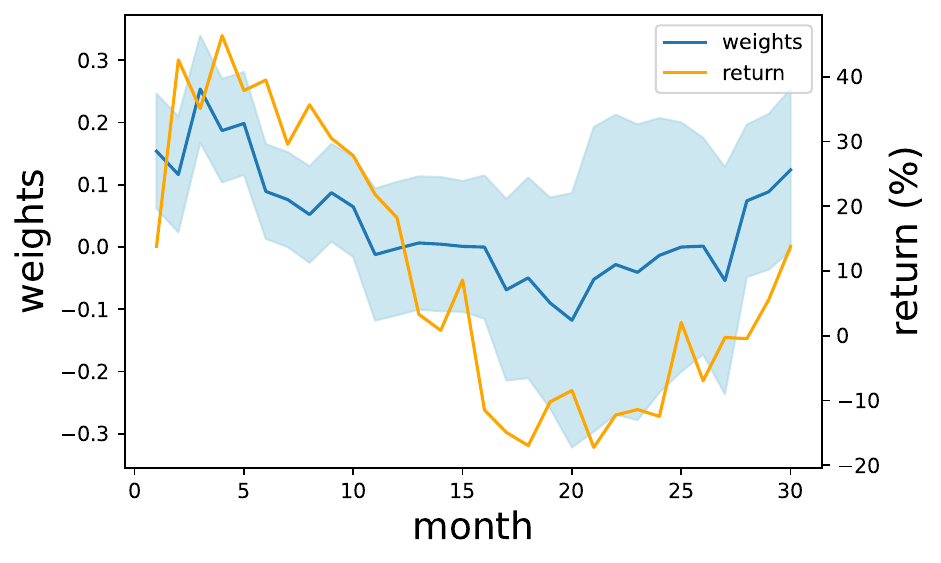}\label{log2}}
\vskip-0.3cm
\caption{		
\textit{
Trajectories of portfolio weights and corresponding stock returns. We show two stocks under two portfolio models. The blue lines depict the predicted weights for the stock, while the shaded blue bands indicate the estimated standard deviations of these weights, computed based on the derived asymptotic normality results. The yellow lines depict the returns of the same~stock.$\quad\;\;$}}	
\label{fig:5}\vskip-0.5cm
\end{figure}
}

\subsection{Poisson regression: Chicago air pollution data}\label{sec:5.4}

In this section, we compare unconstrained and constrained methods for generalized linear regression problems. We use daily air pollution and death rate data for Chicago \citep{Wood2017Generalized} to analyze the relationship between death rates and air pollution levels.
Following \cite{Toulis2017Asymptotic}, we fit a Poisson regression model that regresses the death counts $(\zeta^b)$ on six variables: Intercept, Time, PM$_{10}$, PM$_{2.5}$, SO$_2$, and O$_3$ $(\zeta^\ba)$.~In particular, we assume $\zeta^{b} | \zeta^{\ba} \sim \text{Poisson}(\lambda(\zeta^{\ba}))$ with $\log \lambda(\zeta^{\ba}) = \zeta^{\ba\top}\tx$. 
In this study, we may want to incorporate the prior domain knowledge that higher pollutant concentrations are expected to increase mortality, and thus impose nonnegativity constraints on the coefficients of the pollution covariates. Accordingly, we compare the following two methods:
\vskip-1.3cm
\begin{equation}\label{equ:model:1}
\begin{minipage}{0.45\textwidth}
\begin{equation}
\min_{\bx} \; \mE_\zeta[
\zeta^{b}\,\zeta^{\ba \top}\bm{x}
- \exp (\zeta^{\ba \top}\bm{x})],
\end{equation}
\end{minipage}
\hfill
\begin{minipage}{0.45\textwidth}
\begin{equation}
\begin{split}
\min_{\bx} \;\; & \mE_\zeta[
\zeta^{b}\,\zeta^{\ba \top}\bm{x}
- \exp (\zeta^{\ba \top}\bm{x})],\\
\text{ s.t. } \; & \bx_{3:6}\geq \0.	
\end{split}
\end{equation}
\end{minipage}
\end{equation}

The above unconstrained Poisson model is fitted using the \texttt{statsmodels} package in Python \citep{Seabold2010Statsmodels}, while the constrained model is fitted using the proposed SSQP method. We summarize the estimated coefficients along with their 95\% confidence intervals~and $p$-values in Table~\ref{tab:5} (left).
From this table, we observe that the unconstrained method estimates the coefficient of O$_3$ to be negative and significant, which contradicts prior domain knowledge and can be difficult to interpret. In contrast, the constrained method correctly identifies the~coefficient of O$_3$ as active. Among other variables, the two methods produce comparable coefficient estimates. For example, the coefficients estimated by the unconstrained method lie within the 95\% confidence intervals constructed by the constrained method, and vice versa. Furthermore, both methods identify the Intercept, Time, and SO$_2$ as significant, while PM$_{10}$ as insignificant. The only difference occurs with PM$_{2.5}$, which is insignificant under the unconstrained~method but significant under the constrained method.

{\spacingset{1.1}
\begin{table}[h!]
\centering
\caption{\textit{Summary of two Poisson regression models \eqref{equ:model:1} (left) and \eqref{equ:model:2} (right) applied to the Chicago air pollution data.}
}\label{tab:5}
\begin{minipage}{0.495\textwidth}
\centering
\resizebox{\textwidth}{!}{
\begin{tabular}{ccccc}
\hline\\[-6pt]
\textbf{Var}  & \textbf{Method} & \textbf{Coeff} ($10^{-2}$) & \textbf{95\% CI} ($10^{-2}$) & \textbf{$p$-Value} \\[2pt]
\hline\\[-6pt]
\multirow{2}{*}{\textbf{Intercept}} & {Uncons} & 4.6968 & [4.690, 4.704] & <0.001 \\[2pt]
& {Cons} & 4.6974 & [4.692, 4.703] & <0.001 \\[2pt]
\hline\\[-6pt]
\multirow{2}{*}{\textbf{Time}} & {Uncons} & 0.95 & [0.17, 1.74] & 0.008 \\[2pt]
& {Cons} & 1.13 & [0.64, 1.63] & <0.001 \\[2pt]
\hline\\[-6pt]
\multirow{2}{*}{\textbf{PM$_{10}$}} & {Uncons} & 0.42 & [-0.57, 1.42] &  0.396 \\[2pt]
& {Cons} & 0.13 & [-0.56,0.79] &  0.371 \\[2pt]
\hline\\[-6pt]
\multirow{2}{*}{\textbf{PM$_{2.5}$}} &  {Uncons} & 0.72 &  [-0.08, 1.52] & {\blue 0.103} \\[2pt]
& {Cons} & 0.65 & [0.02, 1.28] & {\red 0.023} \\[2pt]
\hline\\[-6pt]
\multirow{2}{*}{\textbf{SO$_2$}} & {Uncons} & 1.38 & [0.58, 2.20] & 0.001 \\[2pt]
& {Cons} & 2.08 & [1.43, 2.73] & <0.001 \\[2pt]
\hline\\[-6pt]
\multirow{2}{*}{\textbf{O$_3$}} & {Uncons} & {\blue -2.97} & [-3.70, -2.24] & {\blue <0.001} \\[2pt]
& {Cons}  & {\red 0.00} & {\red active}  & \\[2pt]
\hline\\[-6pt]
\end{tabular}}
\end{minipage}
\hfill
\begin{minipage}{0.495\textwidth}
\centering
\vskip-0.7cm
\resizebox{\textwidth}{!}{
\begin{tabular}{ccccc}
\hline\\[-6pt]
\textbf{Var}  & \textbf{Method} & \textbf{Coeff} ($10^{-2}$) & \textbf{95\% CI} ($10^{-2}$) & \textbf{$p$-Value} \\[2pt]
\hline\\[-6pt]
\multirow{2}{*}{\textbf{Intercept}} & {Uncons} & 4.6972 & [4.690, 4.704] & <0.001 \\[2pt]
& {Cons} & 4.6973 & [4.692, 4.703] & <0.001\\[2pt]
\hline\\[-6pt]
\multirow{2}{*}{\textbf{Time}} & {Uncons} & 1.21 & [0.53, 1.89] & 0.001 \\[2pt]
& {Cons} & 1.13 & [0.64, 1.63] & <0.001 \\[2pt]
\hline\\[-6pt]
\multirow{2}{*}{\textbf{PM$_{10}$}} & {Uncons} &  -0.86 &  [-1.82, 0.10] & { 0.062} \\[2pt]
& {Cons} & 0.11 & [-0.52, 0.74] & { 0.362} \\[2pt]
\hline\\[-6pt]
\multirow{2}{*}{\textbf{PM$_{2.5}$}} & {Uncons} & 1.37 & [0.51, 2.23] & {\red 0.001} \\[2pt]
& {Cons} & 0.65 & [0.01, 1.28] & {\red 0.022} \\[2pt]
\hline\\[-6pt]
\multirow{2}{*}{\textbf{SO$_2$}} & {Uncons} & 2.06 & [1.28, 2.84] & <0.001 \\[2pt]
& {Cons} & 2.08 & [1.42, 2.73] & <0.001 \\[2pt]
\hline\\[-6pt]
\end{tabular}}
\end{minipage}
\end{table}\vskip-0.5cm
}

When we encounter a negative coefficient of O$_3$ after fitting the model, we may next perform model modification by simply removing this variable for sake of interpretability. As such, we next consider a sub-model comparison where we remove O$_3$ variable:
\vskip -1cm
\begin{equation}\label{equ:model:2}
\begin{minipage}{0.45\textwidth}
\begin{equation}
\min_{\bx} \; \mE_\zeta[
\zeta^{b}\,\zeta^{\ba \top}\bm{x} - \exp (\zeta^{\ba \top}\bm{x})],
\end{equation}
\end{minipage}
\hfill
\begin{minipage}{0.45\textwidth}
\begin{equation}
\begin{split}
\min_{\bx} \;\; & \mE_\zeta[\zeta^{b}\,\zeta^{\ba \top}\bm{x} - \exp (\zeta^{\ba \top}\bm{x})],\\
\text{ s.t. } \; & \bx_{3:5}\geq \0.	
\end{split}
\end{equation}
\end{minipage}
\end{equation}
The results are summarized in Table~\ref{tab:5} (right). From this table, we observe that the two methods output largely similar results. In particular, both methods identify the Intercept, Time, and SO$_2$ as significant, and PM$_{10}$ as insignificant, consistent with the conclusions drawn from~the~full model. Notably, both methods also identify PM$_{2.5}$ as significant, which aligns with the constrained method fitted on the full model. 
To summarize, the unconstrained method may yield~inconsistency between fitting the full and sub models, while our constrained method preserves~consistency when active variables on the boundary are removed. This illustrates the benefits of~applying our constrained methods when incorporating prior domain knowledge into model fitting.\;

\section{Conclusion and Future Work}\label{sec:6}

In this paper, we studied online statistical inference for the solutions of stochastic optimization problems with equality and inequality constraints. We developed a stochastic sequential quadratic programming method that incorporates a moving-average gradient scheme to correct the bias in the stochastic step direction induced by inequality constraints. We established~global almost-sure convergence and proved that the proposed method achieves local asymptotic normality with a minimax-optimal primal–dual limiting covariance matrix in the sense of H\'ajek and Le Cam. 
Furthermore, we proposed a plug-in covariance matrix estimator for practical~inference. Extensive experiments on benchmark nonlinear problems from the~CUTEst~test~set,~as well as on constrained generalized linear models and portfolio allocation tasks using both synthetic and real data, demonstrated the superior empirical performance of our method and confirmed its effectiveness of online constrained inference in practice.

As for future directions, it would be significant to provide a non-asymptotic analysis that quantifies how quickly the stochastic iterates approach the limiting distribution.
For example, \cite{Anastasiou2019Normal} derived a non-asymptotic convergence rate of averaged SGD to a normal distribution by applying a non-asymptotic martingale central limit theorem. Establishing comparable bounds for SSQP methods on constrained problems would further \mbox{highlight}~the effectiveness of this type of methods. Furthermore, recent work has shown promising advances~in applying SGD methods to high-dimensional settings \citep{Li2025Statistical}.~Developing a high-dimensional theory for constrained model estimation would require incorporating regularization techniques such as sparsity-inducing penalties, manifold constraints, or low-rank constraints to enable~valid analysis. Integrating these techniques with the proposed SSQP method can open new pathways toward scalable, structure-aware online inference for high-dimensional constrained problems.$\;\;$

\printbibliography[title={References}]

@Article{Polyak1992Acceleration,
  author    = {Polyak, B. T. and Juditsky, A. B.},
  journal   = {SIAM Journal on Control and Optimization},
  title     = {Acceleration of Stochastic Approximation by Averaging},
  year      = {1992},
  issn      = {1095-7138},
  month     = jul,
  number    = {4},
  pages     = {838--855},
  volume    = {30},
  doi       = {10.1137/0330046},
  publisher = {Society for Industrial & Applied Mathematics (SIAM)},
}

@Article{Chen2020Statistical,
  author    = {Chen, Xi and Lee, Jason D. and Tong, Xin T. and Zhang, Yichen},
  journal   = {The Annals of Statistics},
  title     = {Statistical inference for model parameters in stochastic gradient descent},
  year      = {2020},
  issn      = {0090-5364},
  month     = feb,
  number    = {1},
  volume    = {48},
  doi       = {10.1214/18-aos1801},
  publisher = {Institute of Mathematical Statistics},
}

@Article{Leluc2023Asymptotic,
  author  = {Leluc, R{\'e}mi and Portier, Fran{\c{c}}ois},
  journal = {Transactions on Machine Learning Research},
  title   = {Asymptotic Analysis of Conditioned Stochastic Gradient Descent},
  year    = {2023},
}

@InProceedings{Anastasiou2019Normal,
  author       = {Anastasiou, Andreas and Balasubramanian, Krishnakumar and Erdogdu, Murat A},
  booktitle    = {Conference on Learning Theory (COLT)},
  title        = {Normal approximation for stochastic gradient descent via non-asymptotic rates of martingale {CLT}},
  year         = {2019},
  organization = {PMLR},
  pages        = {115--137},
  url          = {http://proceedings.mlr.press/v99/anastasiou19a.html},
}

@Article{Na2023Inequality,
  author    = {Na, Sen and Anitescu, Mihai and Kolar, Mladen},
  journal   = {Mathematical Programming},
  title     = {Inequality constrained stochastic nonlinear optimization via active-set sequential quadratic programming},
  year      = {2023},
  issn      = {1436-4646},
  month     = mar,
  number    = {1–2},
  pages     = {279--353},
  volume    = {202},
  doi       = {10.1007/s10107-023-01935-7},
  publisher = {Springer Science and Business Media LLC},
}

@Article{Fang2024Fully,
  author    = {Fang, Yuchen and Na, Sen and Mahoney, Michael W. and Kolar, Mladen},
  journal   = {SIAM Journal on Optimization},
  title     = {Fully Stochastic Trust-Region Sequential Quadratic Programming for Equality-Constrained Optimization Problems},
  year      = {2024},
  issn      = {1095-7189},
  month     = jun,
  number    = {2},
  pages     = {2007--2037},
  volume    = {34},
  doi       = {10.1137/22m1537862},
  publisher = {Society for Industrial & Applied Mathematics (SIAM)},
}

@Article{Na2022adaptive,
  author    = {Na, Sen and Anitescu, Mihai and Kolar, Mladen},
  journal   = {Mathematical Programming},
  title     = {An adaptive stochastic sequential quadratic programming with differentiable exact augmented lagrangians},
  year      = {2022},
  issn      = {1436-4646},
  month     = jun,
  number    = {1–2},
  pages     = {721--791},
  volume    = {199},
  doi       = {10.1007/s10107-022-01846-z},
  publisher = {Springer Science and Business Media LLC},
}

@Article{Curtis2023Worst,
  author    = {Curtis, Frank E. and O’Neill, Michael J. and Robinson, Daniel P.},
  journal   = {Mathematical Programming},
  title     = {Worst-case complexity of an SQP method for nonlinear equality constrained stochastic optimization},
  year      = {2023},
  issn      = {1436-4646},
  month     = jun,
  doi       = {10.1007/s10107-023-01981-1},
  publisher = {Springer Science and Business Media LLC},
}

@Article{Curtis2024Sequential,
  author    = {Curtis, Frank E. and Robinson, Daniel P. and Zhou, Baoyu},
  journal   = {SIAM Journal on Optimization},
  title     = {Sequential Quadratic Optimization for Stochastic Optimization with Deterministic Nonlinear Inequality and Equality Constraints},
  year      = {2024},
  issn      = {1095-7189},
  month     = nov,
  number    = {4},
  pages     = {3592--3622},
  volume    = {34},
  doi       = {10.1137/23m1556149},
  publisher = {Society for Industrial & Applied Mathematics (SIAM)},
}

@Article{Berahas2023Stochastic,
  author    = {Berahas, Albert S. and Curtis, Frank E. and O’Neill, Michael J. and Robinson, Daniel P.},
  journal   = {Mathematics of Operations Research},
  title     = {A Stochastic Sequential Quadratic Optimization Algorithm for Nonlinear-Equality-Constrained Optimization with Rank-Deficient Jacobians},
  year      = {2023},
  issn      = {1526-5471},
  month     = oct,
  doi       = {10.1287/moor.2021.0154},
  publisher = {Institute for Operations Research and the Management Sciences (INFORMS)},
}

@Article{Duchi2021Asymptotic,
  author    = {Duchi, John C. and Ruan, Feng},
  journal   = {The Annals of Statistics},
  title     = {Asymptotic optimality in stochastic optimization},
  year      = {2021},
  issn      = {0090-5364},
  month     = feb,
  number    = {1},
  volume    = {49},
  doi       = {10.1214/19-aos1831},
  publisher = {Institute of Mathematical Statistics},
}

@Article{Berahas2021Sequential,
  author    = {Berahas, Albert S. and Curtis, Frank E. and Robinson, Daniel and Zhou, Baoyu},
  journal   = {SIAM Journal on Optimization},
  title     = {Sequential Quadratic Optimization for Nonlinear Equality Constrained Stochastic Optimization},
  year      = {2021},
  issn      = {1095-7189},
  month     = jan,
  number    = {2},
  pages     = {1352--1379},
  volume    = {31},
  doi       = {10.1137/20m1354556},
  publisher = {Society for Industrial & Applied Mathematics (SIAM)},
}

@Article{Xu2015Smoothing,
  author    = {Xu, Mengwei and Ye, Jane J. and Zhang, Liwei},
  journal   = {SIAM Journal on Optimization},
  title     = {Smoothing SQP Methods for Solving Degenerate Nonsmooth Constrained Optimization Problems with Applications to Bilevel Programs},
  year      = {2015},
  issn      = {1095-7189},
  month     = jan,
  number    = {3},
  pages     = {1388--1410},
  volume    = {25},
  doi       = {10.1137/140971580},
  publisher = {Society for Industrial & Applied Mathematics (SIAM)},
}

@Article{Boyer2022asymptotic,
  author    = {Boyer, Claire and Godichon-Baggioni, Antoine},
  journal   = {Computational Optimization and Applications},
  title     = {On the asymptotic rate of convergence of Stochastic Newton algorithms and their Weighted Averaged versions},
  year      = {2022},
  issn      = {1573-2894},
  month     = dec,
  number    = {3},
  pages     = {921--972},
  volume    = {84},
  doi       = {10.1007/s10589-022-00442-3},
  publisher = {Springer Science and Business Media LLC},
}

@Article{Robinson1976Stability,
  author    = {Robinson, Stephen M.},
  journal   = {SIAM Journal on Numerical Analysis},
  title     = {Stability Theory for Systems of Inequalities, Part II: Differentiable Nonlinear Systems},
  year      = {1976},
  issn      = {1095-7170},
  month     = sep,
  number    = {4},
  pages     = {497--513},
  volume    = {13},
  doi       = {10.1137/0713043},
  publisher = {Society for Industrial & Applied Mathematics (SIAM)},
}

@InCollection{Robbins1971convergence,
  author    = {H. Robbins and D. Siegmund},
  booktitle = {Optimizing Methods in Statistics},
  publisher = {Elsevier},
  title     = {A convergence theorem for non negative almost supermartingales and some applications},
  year      = {1971},
  pages     = {233--257},
  doi       = {10.1016/b978-0-12-604550-5.50015-8},
}

@InProceedings{Seabold2010Statsmodels,
  author    = {Seabold, Skipper and Perktold, Josef},
  booktitle = {Proceedings of the 9th Python in Science Conference},
  title     = {Statsmodels: Econometric and statistical modeling with python},
  year      = {2010},
  number    = {61},
  volume    = {57},
}

@Article{Shapiro2000asymptotics,
  author    = {Shapiro, Alexander},
  journal   = {The Annals of Statistics},
  title     = {On the asymptotics of constrained local $M$-estimators},
  year      = {2000},
  issn      = {0090-5364},
  month     = may,
  number    = {3},
  volume    = {28},
  doi       = {10.1214/aos/1015952006},
  publisher = {Institute of Mathematical Statistics},
}

@Article{Sen1979Asymptotic,
  author    = {Sen, Pranab Kumar},
  journal   = {The Annals of Statistics},
  title     = {Asymptotic Properties of Maximum Likelihood Estimators Based on Conditional Specification},
  year      = {1979},
  issn      = {0090-5364},
  month     = sep,
  number    = {5},
  volume    = {7},
  doi       = {10.1214/aos/1176344785},
  publisher = {Institute of Mathematical Statistics},
}

@Article{Dupacova1988Asymptotic,
  author    = {Dupacova, Jitka and Wets, Roger},
  journal   = {The Annals of Statistics},
  title     = {Asymptotic Behavior of Statistical Estimators and of Optimal Solutions of Stochastic Optimization Problems},
  year      = {1988},
  issn      = {0090-5364},
  month     = dec,
  number    = {4},
  volume    = {16},
  doi       = {10.1214/aos/1176351052},
  publisher = {Institute of Mathematical Statistics},
}

@Article{Chen2018Neural,
  author  = {Chen, Ricky TQ and Rubanova, Yulia and Bettencourt, Jesse and Duvenaud, David K},
  journal = {Advances in Neural Information Processing Systems (NeurIPS)},
  title   = {Neural ordinary differential equations},
  year    = {2018},
  volume  = {31},
}

@Article{Raissi2019Physics,
  author    = {Raissi, M. and Perdikaris, P. and Karniadakis, G.E.},
  journal   = {Journal of Computational Physics},
  title     = {Physics-informed neural networks: A deep learning framework for solving forward and inverse problems involving nonlinear partial differential equations},
  year      = {2019},
  issn      = {0021-9991},
  month     = feb,
  pages     = {686--707},
  volume    = {378},
  doi       = {10.1016/j.jcp.2018.10.045},
  publisher = {Elsevier BV},
}

@Article{Wang2021Learning,
  author    = {Wang, Sifan and Wang, Hanwen and Perdikaris, Paris},
  journal   = {Science Advances},
  title     = {Learning the solution operator of parametric partial differential equations with physics-informed DeepONets},
  year      = {2021},
  issn      = {2375-2548},
  month     = oct,
  number    = {40},
  volume    = {7},
  doi       = {10.1126/sciadv.abi8605},
  publisher = {American Association for the Advancement of Science (AAAS)},
}

@Article{Gould2014CUTEst,
  author    = {Gould, Nicholas I. M. and Orban, Dominique and Toint, Philippe L.},
  journal   = {Computational Optimization and Applications},
  title     = {CUTEst: a Constrained and Unconstrained Testing Environment with safe threads for mathematical optimization},
  year      = {2014},
  issn      = {1573-2894},
  month     = aug,
  number    = {3},
  pages     = {545--557},
  volume    = {60},
  doi       = {10.1007/s10589-014-9687-3},
  publisher = {Springer Science and Business Media LLC},
}

@Article{Fowkes2022PyCUTEst,
  author    = {Fowkes, Jaroslav and Roberts, Lindon and B{\H{u}}rmen, {\'A}rp{\'a}d},
  journal   = {Journal of Open Source Software},
  title     = {PyCUTEst: an open source Python package of optimization test problems},
  year      = {2022},
  issn      = {2475-9066},
  month     = oct,
  number    = {78},
  pages     = {4377},
  volume    = {7},
  doi       = {10.21105/joss.04377},
  publisher = {The Open Journal},
}

@Article{Daniel1973Stability,
  author    = {James W. Daniel},
  journal   = {Mathematical Programming},
  title     = {Stability of the solution of definite quadratic programs},
  year      = {1973},
  month     = {dec},
  number    = {1},
  pages     = {41--53},
  volume    = {5},
  doi       = {10.1007/bf01580110},
  publisher = {Springer Science and Business Media {LLC}},
}

@Article{Du2022High,
  author    = {Jin-Hong Du and Yifeng Guo and Xueqin Wang},
  journal   = {Journal of the American Statistical Association},
  title     = {High-Dimensional Portfolio Selection with Cardinality Constraints},
  year      = {2022},
  month     = {nov},
  number    = {542},
  pages     = {779--791},
  volume    = {118},
  doi       = {10.1080/01621459.2022.2133718},
  publisher = {Informa {UK} Limited},
}

@InProceedings{Fan2007Variable,
  author       = {Fan, Jianqing},
  booktitle    = {Proceedings of the 4th International Congress of Chinese Mathematicians},
  title        = {Variable screening in high-dimensional feature space},
  year         = {2007},
  organization = {Citeseer},
  pages        = {735--747},
  volume       = {2},
}

@Article{Fan2012Vast,
  author    = {Jianqing Fan and Jingjin Zhang and Ke Yu},
  journal   = {Journal of the American Statistical Association},
  title     = {Vast Portfolio Selection With Gross-Exposure Constraints},
  year      = {2012},
  month     = {jun},
  number    = {498},
  pages     = {592--606},
  volume    = {107},
  doi       = {10.1080/01621459.2012.682825},
  publisher = {Informa {UK} Limited},
}

@Article{Na2019High,
  author  = {Na, Sen and Yang, Zhuoran and Wang, Zhaoran and Kolar, Mladen},
  journal = {Journal of Machine Learning Research},
  title   = {High-dimensional Varying Index Coefficient Models via Stein's Identity.},
  year    = {2019},
  pages   = {152--1},
  volume  = {20},
}

@Article{Na2021High,
  author    = {Sen Na and Mladen Kolar},
  journal   = {Bernoulli},
  title     = {High-dimensional index volatility models via Stein's identity},
  year      = {2021},
  month     = {may},
  number    = {2},
  volume    = {27},
  doi       = {10.3150/20-bej1238},
  publisher = {Bernoulli Society for Mathematical Statistics and Probability},
}

@Article{Shapiro1985Asymptotic,
  author    = {Shapiro, Alexander},
  journal   = {Biometrika},
  title     = {Asymptotic distribution of test statistics in the analysis of moment structures under inequality constraints},
  year      = {1985},
  number    = {1},
  pages     = {133--144},
  volume    = {72},
  doi       = {10.1093/biomet/72.1.133},
  publisher = {Oxford University Press ({OUP})},
}

@Article{Zafar2019Fairness,
  author    = {Zafar, Muhammad Bilal and Valera, Isabel and Gomez-Rodriguez, Manuel and Gummadi, Krishna P},
  journal   = {The Journal of Machine Learning Research},
  title     = {Fairness constraints: A flexible approach for fair classification},
  year      = {2019},
  number    = {1},
  pages     = {2737--2778},
  volume    = {20},
  publisher = {JMLR. org},
}

@Article{Cuomo2022Scientific,
  author    = {Cuomo, Salvatore and Di Cola, Vincenzo Schiano and Giampaolo, Fabio and Rozza, Gianluigi and Raissi, Maziar and Piccialli, Francesco},
  journal   = {Journal of Scientific Computing},
  title     = {Scientific Machine Learning Through Physics–Informed Neural Networks: Where we are and What’s Next},
  year      = {2022},
  issn      = {1573-7691},
  month     = jul,
  number    = {3},
  volume    = {92},
  doi       = {10.1007/s10915-022-01939-z},
  publisher = {Springer Science and Business Media LLC},
}

@Article{Krishnapriyan2021Characterizing,
  author  = {Krishnapriyan, Aditi and Gholami, Amir and Zhe, Shandian and Kirby, Robert and Mahoney, Michael W},
  journal = {Advances in Neural Information Processing Systems (NeurIPS)},
  title   = {Characterizing possible failure modes in physics-informed neural networks},
  year    = {2021},
  volume  = {34},
}

@InProceedings{Negiar2023Learning,
  author    = {Geoffrey N{\'e}giar and Michael W. Mahoney and Aditi Krishnapriyan},
  booktitle = {International Conference on Learning Representations},
  title     = {Learning differentiable solvers for systems with hard constraints},
  year      = {2023},
}

@Book{Vaart1998Asymptotic,
  author    = {Vaart, A. W. van der},
  publisher = {Cambridge University Press},
  title     = {Asymptotic Statistics},
  year      = {1998},
  month     = oct,
  doi       = {10.1017/cbo9780511802256},
}

@InProceedings{Hajek1972Local,
  author    = {H{\'a}jek, Jaroslav},
  booktitle = {Proceedings of the sixth Berkeley symposium on mathematical statistics and probability},
  title     = {Local asymptotic minimax and admissibility in estimation},
  year      = {1972},
  pages     = {175--194},
  volume    = {1},
}

@InProceedings{LeCam1972Limits,
  author       = {Le Cam, Lucien},
  booktitle    = {Proceedings of the Sixth Berkeley Symposium on Mathematical Statistics and Probability},
  title        = {Limits of experiments},
  year         = {1972},
  organization = {University of California Press Berkeley-Los Angeles},
  pages        = {245--261},
  volume       = {1},
}

@Article{Davis2024Asymptotic,
  author    = {Davis, Damek and Drusvyatskiy, Dmitriy and Jiang, Liwei},
  journal   = {The Annals of Statistics},
  title     = {Asymptotic normality and optimality in nonsmooth stochastic approximation},
  year      = {2024},
  issn      = {0090-5364},
  month     = aug,
  number    = {4},
  volume    = {52},
  doi       = {10.1214/24-aos2401},
  publisher = {Institute of Mathematical Statistics},
}

@Article{Robbins1951Stochastic,
  author    = {Herbert Robbins and Sutton Monro},
  journal   = {The Annals of Mathematical Statistics},
  title     = {A Stochastic Approximation Method},
  year      = {1951},
  month     = {sep},
  number    = {3},
  pages     = {400--407},
  volume    = {22},
  doi       = {10.1214/aoms/1177729586},
  publisher = {Institute of Mathematical Statistics},
}

@Article{Kiefer1952Stochastic,
  author    = {J. Kiefer and J. Wolfowitz},
  journal   = {The Annals of Mathematical Statistics},
  title     = {Stochastic Estimation of the Maximum of a Regression Function},
  year      = {1952},
  month     = {sep},
  number    = {3},
  pages     = {462--466},
  volume    = {23},
  doi       = {10.1214/aoms/1177729392},
  publisher = {Institute of Mathematical Statistics},
}

@Article{Toulis2017Asymptotic,
  author    = {Toulis, Panos and Airoldi, Edoardo M.},
  journal   = {The Annals of Statistics},
  title     = {Asymptotic and finite-sample properties of estimators based on stochastic gradients},
  year      = {2017},
  issn      = {0090-5364},
  month     = aug,
  number    = {4},
  volume    = {45},
  doi       = {10.1214/16-aos1506},
  publisher = {Institute of Mathematical Statistics},
}

@Article{Zhu2021Online,
  author    = {Zhu, Wanrong and Chen, Xi and Wu, Wei Biao},
  journal   = {Journal of the American Statistical Association},
  title     = {Online Covariance Matrix Estimation in Stochastic Gradient Descent},
  year      = {2021},
  issn      = {1537-274X},
  month     = jul,
  number    = {541},
  pages     = {393--404},
  volume    = {118},
  doi       = {10.1080/01621459.2021.1933498},
  publisher = {Informa UK Limited},
}

@Article{Lee2022Fast,
  author    = {Lee, Sokbae and Liao, Yuan and Seo, Myung Hwan and Shin, Youngki},
  journal   = {Proceedings of the AAAI Conference on Artificial Intelligence},
  title     = {Fast and Robust Online Inference with Stochastic Gradient Descent via Random Scaling},
  year      = {2022},
  issn      = {2159-5399},
  month     = jun,
  number    = {7},
  pages     = {7381--7389},
  volume    = {36},
  doi       = {10.1609/aaai.v36i7.20701},
  publisher = {Association for the Advancement of Artificial Intelligence (AAAI)},
}

@Article{Toulis2021Proximal,
  author    = {Toulis, Panos and Horel, Thibaut and Airoldi, Edoardo M.},
  journal   = {Journal of the Royal Statistical Society Series B: Statistical Methodology},
  title     = {The Proximal Robbins–Monro Method},
  year      = {2021},
  issn      = {1467-9868},
  month     = dec,
  number    = {1},
  pages     = {188--212},
  volume    = {83},
  doi       = {10.1111/rssb.12405},
  publisher = {Oxford University Press (OUP)},
}

@Book{Nocedal2006Numerical,
  author    = {Nocedal, Jorge and Wright, Stephen J},
  publisher = {Springer New York},
  title     = {Numerical optimization},
  year      = {2006},
  doi       = {10.1007/978-0-387-40065-5},
  journal   = {Springer Series in Operations Research and Financial Engineering},
}

@Article{Nagaraj1991Estimation,
  author    = {Nagaraj, Neerchal K. and Fuller, Wayne A.},
  journal   = {The Annals of Statistics},
  title     = {Estimation of the Parameters of Linear Time Series Models Subject to Nonlinear Restrictions},
  year      = {1991},
  issn      = {0090-5364},
  month     = sep,
  number    = {3},
  volume    = {19},
  doi       = {10.1214/aos/1176348242},
  publisher = {Institute of Mathematical Statistics},
}

@Article{Fang2024Trust,
  author  = {Fang, Yuchen and Na, Sen and Mahoney, Michael W and Kolar, Mladen},
  journal = {arXiv preprint arXiv:2409.15734},
  title   = {Trust-Region Sequential Quadratic Programming for Stochastic Optimization with Random Models},
  year    = {2024},
}

@Article{Bercu2020Efficient,
  author    = {Bercu, Bernard and Godichon, Antoine and Portier, Bruno},
  journal   = {SIAM Journal on Control and Optimization},
  title     = {An Efficient Stochastic Newton Algorithm for Parameter Estimation in Logistic Regressions},
  year      = {2020},
  issn      = {1095-7138},
  month     = jan,
  number    = {1},
  pages     = {348--367},
  volume    = {58},
  doi       = {10.1137/19m1261717},
  publisher = {Society for Industrial & Applied Mathematics (SIAM)},
}

@Article{Chen2024Online,
  author    = {Chen, Xi and Lai, Zehua and Li, He and Zhang, Yichen},
  journal   = {Journal of the American Statistical Association},
  title     = {Online Statistical Inference for Stochastic Optimization via Kiefer-Wolfowitz Methods},
  year      = {2024},
  issn      = {1537-274X},
  month     = jan,
  number    = {548},
  pages     = {2972--2982},
  volume    = {119},
  doi       = {10.1080/01621459.2023.2296703},
  publisher = {Informa UK Limited},
}

@Article{Fang2018Online,
  author  = {Fang, Yixin and Xu, Jinfeng and Yang, Lei},
  journal = {Journal of Machine Learning Research},
  title   = {Online bootstrap confidence intervals for the stochastic gradient descent estimator},
  year    = {2018},
  number  = {78},
  pages   = {1--21},
  volume  = {19},
}

@Article{Jiang2025Online,
  author  = {Jiang, Liwei and Roy, Abhishek and Balasubramanian, Krishna and Davis, Damek and Drusvyatskiy, Dmitriy and Na, Sen},
  journal = {The 38th Annual Conference on Learning Theory (COLT)},
  title   = {Online covariance estimation in nonsmooth stochastic approximation},
  year    = {2025},
}

@Article{Kuang2025Online,
  author  = {Kuang, Wei and Anitescu, Mihai and Na, Sen},
  journal = {arXiv preprint arXiv:2502.07114},
  title   = {Online covariance matrix estimation in sketched newton methods},
  year    = {2025},
}

@Article{Na2025Statistical,
  author  = {Na, Sen and Mahoney, Michael},
  journal = {Journal of Machine Learning Research},
  title   = {Statistical inference of constrained stochastic optimization via sketched sequential quadratic programming},
  year    = {2025},
  number  = {33},
  pages   = {1--75},
  volume  = {26},
}

@Article{Polyak1964Some,
  author    = {Polyak, B.T.},
  journal   = {USSR Computational Mathematics and Mathematical Physics},
  title     = {Some methods of speeding up the convergence of iteration methods},
  year      = {1964},
  issn      = {0041-5553},
  month     = jan,
  number    = {5},
  pages     = {1--17},
  volume    = {4},
  doi       = {10.1016/0041-5553(64)90137-5},
  publisher = {Elsevier BV},
}

@Article{Fang2025High,
  author  = {Fang, Yuchen and Lavaei, Javad and Na, Sen},
  journal = {arXiv preprint arXiv:2503.19091},
  title   = {High Probability Complexity Bounds of Trust-Region Stochastic Sequential Quadratic Programming with Heavy-Tailed Noise},
  year    = {2025},
}

@Article{Cenac2025efficient,
  author    = {Cénac, Peggy and Godichon-Baggioni, Antoine and Portier, Bruno},
  journal   = {Bernoulli},
  title     = {An efficient averaged stochastic Gauss-Newton algorithm for estimating parameters of nonlinear regressions models},
  year      = {2025},
  issn      = {1350-7265},
  month     = feb,
  number    = {1},
  volume    = {31},
  doi       = {10.3150/23-bej1637},
  publisher = {Bernoulli Society for Mathematical Statistics and Probability},
}

@Article{Du2025Online,
  author  = {Du, Xinchen and Zhu, Wanrong and Wu, Wei Biao and Na, Sen},
  journal = {arXiv preprint arXiv:2505.18327},
  title   = {Online Statistical Inference of Constrained Stochastic Optimization via Random Scaling},
  year    = {2025},
}

@Book{Bonnans2000Perturbation,
  author    = {Bonnans, J. Frédéric and Shapiro, Alexander},
  publisher = {Springer New York},
  title     = {Perturbation Analysis of Optimization Problems},
  year      = {2000},
  doi       = {10.1007/978-1-4612-1394-9},
}

@Article{Shapiro1990differential,
  author    = {Shapiro, Alexander},
  journal   = {Mathematical Programming},
  title     = {On differential stability in stochastic programming},
  year      = {1990},
  issn      = {1436-4646},
  month     = may,
  number    = {1–3},
  pages     = {107--116},
  volume    = {47},
  doi       = {10.1007/bf01580855},
  publisher = {Springer Science and Business Media LLC},
}

@Book{Dontchev2009Implicit,
  author    = {Dontchev, Asen L. and Rockafellar, R. Tyrrell},
  publisher = {Springer New York},
  title     = {Implicit Functions and Solution Mappings: A View from Variational Analysis},
  year      = {2009},
  isbn      = {9780387878218},
  doi       = {10.1007/978-0-387-87821-8},
  issn      = {1439-7382},
  journal   = {Springer Monographs in Mathematics},
}

@Article{Burer2009Nonconvex,
  author    = {Burer, Samuel and Letchford, Adam N.},
  journal   = {SIAM Journal on Optimization},
  title     = {On Nonconvex Quadratic Programming with Box Constraints},
  year      = {2009},
  issn      = {1095-7189},
  month     = jan,
  number    = {2},
  pages     = {1073--1089},
  volume    = {20},
  doi       = {10.1137/080729529},
  publisher = {Society for Industrial & Applied Mathematics (SIAM)},
}

@Article{Li2025Statistical,
  author  = {Li, Jiaqi and Lou, Zhipeng and Schmidt-Hieber, Johannes and Wu, Wei Biao},
  journal = {Conference on Neural Information Processing Systems (NeurIPS)},
  title   = {Statistical Guarantees for High-Dimensional Stochastic Gradient Descent},
  year    = {2025},
}

@Book{Wood2017Generalized,
  author    = {Wood, Simon N.},
  publisher = {Chapman and Hall/CRC},
  title     = {Generalized Additive Models: An Introduction with R},
  year      = {2017},
  month     = may,
  doi       = {10.1201/9781315370279},
}

@Article{Liu2000Robust,
  author    = {Liu, Xin-wei and Yuan, Ya-xiang},
  journal   = {SIAM Journal on Scientific Computing},
  title     = {A Robust Algorithm for Optimization with General Equality and Inequality Constraints},
  year      = {2000},
  issn      = {1095-7197},
  month     = jan,
  number    = {2},
  pages     = {517--534},
  volume    = {22},
  doi       = {10.1137/s1064827598334861},
  publisher = {Society for Industrial & Applied Mathematics (SIAM)},
}

@Article{Dennis1974characterization,
  author    = {Dennis, J. E. and Moré, Jorge J.},
  journal   = {Mathematics of Computation},
  title     = {A characterization of superlinear convergence and its application to quasi-Newton methods},
  year      = {1974},
  issn      = {1088-6842},
  number    = {126},
  pages     = {549--560},
  volume    = {28},
  doi       = {10.1090/s0025-5718-1974-0343581-1},
  publisher = {American Mathematical Society (AMS)},
}

@Book{Christensen2020Plane,
  author    = {Christensen, Ronald},
  publisher = {Springer International Publishing},
  title     = {Plane Answers to Complex Questions: The Theory of Linear Models},
  year      = {2020},
  doi       = {10.1007/978-3-030-32097-3},
  issn      = {2197-4136},
  journal   = {Springer Texts in Statistics},
}

@Article{Na2025Derivative,
  author  = {Na, Sen},
  journal = {arXiv preprint arXiv:2510.22458},
  title   = {Derivative-Free Sequential Quadratic Programming for Equality-Constrained Stochastic Optimization},
  year    = {2025},
}

@Book{Hall2014Martingale,
  author    = {Hall, Peter and Heyde, Christopher C},
  publisher = {Academic press},
  title     = {Martingale limit theory and its application},
  year      = {2014},
}

@Book{Duflo1997Random,
  author    = {Duflo, Marie},
  editor    = {Stephen S. Wilson},
  publisher = {Springer},
  title     = {Random iterative models},
  year      = {1997},
  address   = {Berlin},
  isbn      = {3540571000},
  note      = {Aus dem Franz. übers.},
  number    = {34},
  series    = {Applications of mathematics},
  ppn_gvk   = {1612000525},
}
%\end{refsection}

%\bibliographystyle{my-plainnat}
%\bibliography{ref}

\newpage
\begin{center}
{\large\bf Appendix: Online Inference of Constrained Optimization: Primal-Dual Optimality and Sequential Quadratic Programming}
\end{center}

\appendix
%\pagebreak
\numberwithin{equation}{section}
\numberwithin{theorem}{section}

%\begin{refsection}
\begingroup
\addtolength{\jot}{4pt}

\section{Motivating Examples}\label{append:A}

The question of when constraints are needed in statistical machine learning problems may be divided into two scenarios. The first scenario refers to problems that are otherwise ill-posed and impossible to solve. For example, in principal component analysis, we estimate the leading eigenvector of the population covariance matrix $\bSigma$. However, maximizing $\bx^\top\bSigma\bx$ is not well-defined without the constraint $\|\bx\|^2=1$. The second scenario refers to problems where constraints incorporate prior model information and enable data to be explained by simpler hypotheses. We introduce several examples that can be cast into the form of Problem \eqref{original_problem0}. $\quad\quad$

\subsection{Constrained regression}\label{sec_constrained_regression}

Given a set of feature-response data $\{(\zeta_1^{\ba}, \zeta_1^b), \ldots, (\zeta_n^{\ba}, \zeta_n^b)\}$, we consider fitting a smooth monotone function $z: \zeta^{\ba}\rightarrow\zeta^b$. Monotonic regression has many applications, such as estimating cumulative distribution functions, and monotonic relationships between two variables are prevalent in various studies. For example, demand is typically a non-increasing function of price, and the rate of chemical processes is an increasing function of temperature. See our study on environmental data in Section \ref{sec:5.4}. A simple approach to fit a monotone function is to select~some basis functions $\{z_1(\cdot),\ldots, z_d(\cdot)\}$ and define
\begin{equation*}
z(\zeta^{\ba}; \bx) = \sum_{i=1}^{d}x_i z_i(\zeta^{\ba}),
\end{equation*}
where $\bx = (x_1,\ldots, x_d)$ are regression coefficients. Then, we minimize the mean-squared error and impose the monotonicity, e.g., non-increasing, via constraints on the coefficients $\bx$:
\begin{equation*}
\min_{\bx}\; \frac{1}{n}\sum_{i=1}^{n} F(\bx; \zeta_i) \coloneqq \frac{1}{n}\sum_{i=1}^{n} (\zeta^b_i - z(\zeta^{\ba}_i;\bx))^2 \quad \text{s.t.}\quad z'(\omega_j;\bx) \coloneqq \sum_{i=1}^{d}x_i z_i'(\omega_j) \leq 0,\;\; j \in[m],
\end{equation*}
where $\omega_1,\ldots,\omega_m$ are a fine grid of points over the range of $\zeta^{\ba}$.

Constraints are also imposed in other statistical problems. In portfolio optimization (cf. Section \ref{sec:5.3}), $\bx$ refers to the asset weights, and it is common to constrain the estimation within the probability simplex $\{\bx\in\mR^d: \1^\top \bx = 1, \bx\geq \0\}$ \citep{Du2022High}.~Alternatively, one may control the risk exposures to a known threshold by imposing box constraints $\|\bx\|_\infty\leq c$, or control allocations among sectors by imposing affine constraints \mbox{$A\bx = \bb$}, where a negative weight signifies shorting the asset \citep{Fan2007Variable, Fan2012Vast}. In semiparametric index models, we impose $\{\bx\in\mR^d:\|\bx\|^2=1, x_1>0\}$ 
to ensure models' identifiability \citep{Na2019High, Na2021High}. In factor analysis, constraints can prevent Heywood cases that cause negative estimates for the variance \citep{Shapiro1985Asymptotic}. In algorithmic fairness, constraints are applied to prevent classifiers from yielding disparate outcomes based on sensitive features such as gender and ethnicity \citep{Zafar2019Fairness}. 
For more examples of constrained regression models, we~refer to \cite{Sen1979Asymptotic, Dupacova1988Asymptotic, Nagaraj1991Estimation, Shapiro2000asymptotics}.

\subsection{Physics-informed machine learning}

A recent trending topic in machine learning is its applications on scientific problems, where~models must adhere to domain knowledge (such as physical laws) that is often described by partial differential equation (PDE) constraints \citep{Cuomo2022Scientific}. 
Consider a smooth neural network model $z(\cdot;\bx): \zeta^{\ba} \rightarrow \zeta^b$, where $\zeta^{\ba} = (s, t)$ is the spatial-temporal pair, $\zeta^b$ is the measurement of some quantity, and $\bx$ denotes the network parameters (e.g., weight matrices). If $z(\cdot;\bx)$ models the transport of the quantity, then it should satisfy the transport equation ($\gamma$ is given):
\begin{equation}\label{snequ:2}
\partial_{t} z(\zeta^{\ba};\bx) + \gamma\cdot\partial_{s}z(\zeta^{\ba};\bx)  = \0.
\end{equation}
Applying PDE constraints to the model $z(\cdot;\bx)$ leads to many important network \mbox{architectures}, including Neural ODEs \citep{Chen2018Neural}, DeepONets \citep{Wang2021Learning}, and physics-informed neural networks \citep{Raissi2019Physics}. In these problems, the desired model is solved from the following constrained optimization:
\begin{equation*}
\min_{\bx} \; \mE_{\zeta\sim \P}[F(\bx; \zeta)] \quad \; \text{s.t.}\quad\; \mathcal{C}_{\text{PDE}}(z(\omega_j;\bx)) = \b0, \;\; j\in[m],
\end{equation*}
where $F(\bx;\zeta)$ is the data fitting loss (e.g., mean-squared error in Section \ref{sec_constrained_regression}), $(\omega_1,\ldots,\omega_m)$ are sampled points in the space of $\zeta^{\ba}$ including both interior and boundary points, and $\mC_{\text{PDE}}(\cdot)$ is a PDE operator with respect to $\zeta^{\ba}$ that encodes domain knowledge (e.g., transport equation as in \eqref{snequ:2}). 
Numerous research has shown that failure to enforce these constraints can lead to~serious interpretability and instability issues in neural network models \citep{Krishnapriyan2021Characterizing,Negiar2023Learning}.

\section{Preparation Definitions and Lemmas}

\begin{lemma}[Theorem 3 in \cite{Robinson1976Stability}]\label{aux:lem:1}
If $\bell\leq \barbx\leq \bu$ satisfies EGMFCQ, then there exists a neighborhood $\mB(\bar\bx;\bar{r}) \coloneqq \{\bell\leq \bx\leq \bu: \|\bx- \barbx\| \leq \bar{r}\}$ for some sufficiently small radius $\bar{r}>0$, such that all points in the neighborhood also satisfy EGMFCQ. 
\end{lemma}

\begin{definition}[Regular mapping sequence, \cite{Vaart1998Asymptotic}]\label{def_regular_mapping}
Let $\mathcal{W} \subseteq \mathbb{R}^{r}$ be a neighborhood of the origin.~A sequence of mappings $\Gamma_{k}: \mathcal{W} \to \mathbb{R}^{d}$ is said to be regular with derivative $\bm{D} \in \mathbb{R}^{d \times r}$~if
\begin{equation*}
\lim_{k \to \infty} \sqrt{k} \left(\Gamma_{k}(\bw) - \Gamma_{k}(\0)\right) = \bm{D} \bw, \quad \text{for all }~\bw \in \mathcal{W}.
\end{equation*}
\end{definition}

\begin{lemma}[H\'ajek and Le Cam's local minimax theorem, \cite{Vaart1998Asymptotic}]\label{theorem_hajek_lecam}
Let $\{\mT_{k},\mathcal{F}_{k},\mathcal{P}_{k,\bv}\}_{\bv \in \mathcal{W}}$ be a locally asymptotically normal family with precision $\bm{Q} \succeq \0$.~Let~$\Gamma_k:\mathcal{W}\rightarrow\mR^d$ be a regular mapping sequence with derivative $\bm{D}$, and let $\ell: \mathbb{R}^{d} \to [0,\infty)$ be any symmetric, quasiconvex, and lower semicontinuous function. Then, for any sequence of estimators $T_{k}: \mT_k\rightarrow \mR^d$, the following inequality holds:
\begin{equation*}
\sup_{\mathcal{W}_0 \subseteq \mathcal{W}, |\mathcal{W}_0|<\infty} \liminf_{k \to \infty} \max_{\bv \in \mathcal{W}_0} \mathbb{E}_{\mathcal{P}_{k,\bv}} [\ell(\sqrt{k} (T_k - \Gamma_k(\bv)))] \geq \mathbb{E} \left[\ell(Z)\right],
\end{equation*}
where $Z \sim \mathcal{N}(\0,\bm{D}\bm{Q}^{-1}\bm{D}^\top)$ when $\bm{Q}$ is invertible. If $\bm{Q}$ is singular, then the above inequality~holds with $Z \sim \mathcal{N}(\0,\bm{D}(\bm{Q} + \gamma \bm{I})^{-1}\bm{D}^\top)$ for any $\gamma > 0$. Here, $\mathbb{E}_{\mathcal{P}_{k,\bv}}[\cdot]$ is taken over the randomness~of samples $\mT_k$.
\end{lemma}

\begin{lemma}[Lemma~B.3 in \cite{Na2025Statistical}]\label{lemma_tool}

Let $\alpha_k = \iota_1 (k+1)^{-b_1}$ and $\beta_k = \iota_2 (k+1)^{-b_2}$ for some $\iota_1, \iota_2 >0$ and $b_1, b_2 > 0$. Then the following statements hold:
\begin{enumerate}[label=(\alph*),topsep=0pt]
\setlength\itemsep{0.0em}
\item Define $\chi = 0$ if $0 < b_2 < 1$ and $\chi = -b_1/\iota_2$ if $b_2 = 1$. If $\sum_{t=1}^{l} a_t + \chi> 0$, then
\begin{equation*}
\lim_{k \to \infty} \frac{1}{\alpha_k} \sum_{i=0}^{k} \prod_{j=i+1}^{k} \prod_{t = 1}^{l} \left( 1 - a_{t} \beta_{j} \right) \beta_{i} \alpha_{i} =  \frac{1}{\sum_{t=1}^{l} a_t + \chi}.
\end{equation*}		
Moreover, for any $b \in \mathbb{R}$ and any sequence $e_k$ satisfying $\lim\limits_{k\to\infty}e_{k} \to 0$, we have
\begin{equation*}
\lim_{k \to \infty} \left\{\frac{1}{\alpha_k} \sum_{i=0}^{k} \prod_{j=i+1}^{k} \prod_{t = 1}^{l} \left( 1 - a_{t} \beta_{j} \right) \beta_{i} \alpha_{i} e_{i} + b \prod_{j=0}^{k} \prod_{t = 1}^{l} \left( 1 - a_{t} \beta_{j} \right) \right\} = 0.
\end{equation*}
\item Suppose $b_2=1$ and let $\chi = -b_1/\iota_2$. If $\sum_{t=1}^{l} a_t + p\chi> 0$ for some $p\in(0, 1)$, then
\begin{equation*}
\lim_{k \to \infty} \frac{1}{\alpha_k^p} \sum_{i=0}^{k} \prod_{j=i+1}^{k} \prod_{t = 1}^{l} \left( 1 - a_{t} \beta_{j} \right) \beta_{i} \alpha_{i} = 0.
\end{equation*}
\item If $0 < b_2 < b_1 \leq 1$, then		
\begin{equation*}
\lim_{k \to \infty} \frac{1}{\alpha_k}  \sum_{i=0}^{k} \prod_{j=i+1}^{k} \left( 1 - \alpha_j \right) \left( 1 - \beta_j \right) \alpha_i \beta_i = 1.
\end{equation*}		
\end{enumerate}
\end{lemma}

\begin{lemma}[Stability of Quadratic Programs, Theorem 2.1 in \cite{Daniel1973Stability}]\label{lemma13}
Consider two constrained strongly convex quadratic problems 
\begin{equation*}
\bm{y}^{\star} \in \min_{\bm{y} \in \Omega} \bm{g}^{\top} \bm{y} + \frac{1}{2} \bm{y}^{\top} \bm{Q} \bm{y} \quad\quad \text{and}\quad\quad \bm{y}^{\star \star} \in \min_{\bm{y} \in \Omega} \bm{g}^{\prime \top} \bm{y} + \frac{1}{2} \bm{y}^{\top} \bm{Q}^{\prime} \bm{y},
\end{equation*}
where the feasible region $\Omega$ is convex.~Suppose $\bm{Q}, \bm{Q}^{\prime}\succeq \nu\bI$ and $\max \{\left\| \bm{y}^{\star} \right\|, \left\| \bm{y}^{\star \star} \right\| \} \leq M_{y}$~for~some $\nu, M_{y}>0$. Then,
\begin{equation*}
\|\bm{y}^{\star} - \bm{y}^{\star \star} \| \leq \upsilon^{-1} ( 1 + M_{y} )\max\{\| \bm{g} - \bm{g}^{\prime} \|, \|  \bm{Q} - \bm{Q}^{\prime}\|\}. 
\end{equation*}
\end{lemma}

\section{Proofs of Main Results}

\subsection{Proof of Theorem \ref{thm:1}}\label{app:c1}

Suppose EGMFCQ holds at $\bell\leq\bx\leq \bu$, we denote $\mA(\bx) = \mA_\bell(\bx)\cup\mA_\bu(\bx)$, $\mA(\bx)^-=[d]\setminus\mA(\bx)$, and let $\bz$ satisfy \eqref{MFCQ_z}. If $\bz = \0$, then we define $\bar{\theta} = 1$. Otherwise, we define
\begin{equation}\label{pf:B1}
\bar{\theta} = \frac{\epsilon}{\|\bz\|}\quad\quad \text{where }\quad \epsilon = \min\cbr{[\bu-\bell]_i, [\bx-\bell]_j, [\bu-\bx]_j: i\in \mA(\bx), j\in\mA(\bx)^-}.
\end{equation}
Note that $\bar{\theta}>0$ since $\epsilon>0$, although we do not restrict $\bar{\theta}\leq 1$. We claim $\Omega(\bx; \bar{\theta})\neq \emptyset$. In~fact, $\bar{\theta}\bz \in \Omega(\bx; \bar{\theta})$ since $\bar{\theta}\bc(\bx) + \nabla \bc(\bx)(\bar{\theta}\bz)\stackrel{\eqref{MFCQ_z}}{=}\0$ and
\begin{equation*}
\begin{split}
& \bell_{\mA_\bell(\bx)} \stackrel{\eqref{nsequ:5}}{=} \bx_{\mA_\bell(\bx)}\stackrel{\eqref{MFCQ_z}}{\leq} [\bx + \bar{\theta}\bz]_{\mA_\bell(\bx)} \stackrel{\eqref{nsequ:5}}{=} [\bell + \bar{\theta}\bz]_{\mA_\bell(\bx)} \leq [\bell + \epsilon\1]_{\mA_\bell(\bx)}\stackrel{\eqref{pf:B1}}{\leq} \bu_{\mA_\bell(\bx)},\\
& \bell_{\mA_\bu(\bx)} \stackrel{\eqref{pf:B1}}{\leq} [\bu - \epsilon\1]_{\mA_\bu(\bx)}\leq [\bu + \bar{\theta}\bz]_{\mA_\bu(\bx)} \stackrel{\eqref{nsequ:5}}{=} [\bx + \bar{\theta}\bz]_{\mA_\bu(\bx)} \stackrel{\eqref{MFCQ_z}}{\leq}\bx_{\mA_\bu(\bx)} \stackrel{\eqref{nsequ:5}}{=}\bu_{\mA_\bu(\bx)},\\
& \bell_{\mA(\bx)^-} \stackrel{\eqref{pf:B1}}{\leq} [\bx-\epsilon\1]_{\mA(\bx)^-} \leq [\bx + \bar{\theta}\bz]_{\mA(\bx)^-} \leq [\bx + \epsilon\1]_{\mA(\bx)^-}\stackrel{\eqref{pf:B1}}{\leq} \bu_{\mA(\bx)^-},
\end{split}
\end{equation*}
where $\1=(1,\ldots,1)\in\mR^d$ denotes the all-one vector and the above display also uses~$-\1\leq \bz/\|\bz\|\leq \1$. From the above display, we know $\bell\leq \bx+\bar{\theta}\bz\leq \bu$; thus, $\bar{\theta}\bz\in\Omega(\bx;\bar{\theta})$. Now, we show that $\Omega(\bx;\theta)\neq\emptyset$ for any $\theta\in[0, \bar{\theta}]$. In fact, for any $\bv\in \Omega(\bx;\bar{\theta})$, we consider $\theta/\bar{\theta}\cdot\bv$. First, we have $\theta \bc(\bx) + \nabla \bc(\bx)(\theta/\bar{\theta} \bv) = \0$ since $\bar{\theta}\bc(\bx) + \nabla \bc(\bx)\bv = \0$. Second, we bound~$\bx+\theta/\bar{\theta}\bv$ by considering two cases based on the sign of $\bv$. We apply $\bell\leq \bx+\bv\leq \bu$ and have
\begin{equation*}
\begin{split}
& \bell_{\{i: \bv_i\geq 0\}}\leq \bx_{\{i: \bv_i\geq 0\}}\leq [\bx + \theta/\bar{\theta}\bv]_{\{i: \bv_i\geq 0\}} \leq [\bx + \bv]_{\{i: \bv_i\geq 0\}}\leq \bu_{\{i: \bv_i\geq 0\}},\\
& \bell_{\{i: \bv_i< 0\}}\leq [\bx + \bv]_{\{i: \bv_i<0\}} \leq [\bx + \theta/\bar{\theta}\bv]_{\{i: \bv_i< 0\}} \leq \bx_{\{i: \bv_i< 0\}}\leq \bu_{\{i: \bv_i<0\}}.
\end{split}
\end{equation*}
Thus, $\bell\leq \bx+\theta/\bar{\theta}\bv\leq \bu$. This completes the proof of the first part of the theorem.

For the second part of the theorem, without loss of generality, we suppose $\lim\limits_{k\rightarrow\infty}\theta_k = 0$ and $\lim\limits_{k\rightarrow\infty}\bx_k = \tx$. Suppose $\tx$ satisfies EGMFCQ, then there exists $\tz$ such that \eqref{MFCQ_z} holds. On the other hand, by Lemma \ref{aux:lem:1} we know $\bx_k$ satisfies EGMFCQ for $k$ large enough, implying~that $\nabla \bc_k$ has full row rank. Noting that 
\begin{equation*}
\tz = - (\nabla \bc^\star)^\top(\nabla\bc^\star(\nabla\bc^\star)^\top)^{-1}\bc^\star + \rbr{\bI - (\nabla \bc^\star)^\top(\nabla\bc^\star(\nabla\bc^\star)^\top)^{-1}\nabla\bc^\star} \tz,
\end{equation*}
we define
\begin{equation*}
\bz_k = -\nabla \bc_k^\top(\nabla \bc_k\nabla \bc_k^\top)^{-1}\bc_k + \rbr{\bI - \nabla \bc_k^\top(\nabla \bc_k\nabla \bc_k^\top)^{-1}\nabla \bc_k}\tz
\end{equation*}
and have $\bc_k+\nabla \bc_k\bz_k = \0$ and $\bz_k\rightarrow\tz$ as $k\rightarrow\infty$. Since $\mA(\tx)^-\subseteq \mA(\bx_k)^-$, we know $\mA(\bx_k)\subseteq \mA(\tx)$. 
Since each entry of $\tz_{\mA(\tx)}$ is nonzero by \eqref{MFCQ_z}, we know each entry of $[\bz_k]_{\mA(\bx_k)}$ is also nonzero and has the same sign as $\tz_{\mA(\bx_k)}$. This shows that the constructed $\bz_k$ satisfies~\eqref{MFCQ_z} for $\bx_k$. Let $l_k\coloneqq \inf \{\|\bz\|: \bz~ \text{satisfies \eqref{MFCQ_z} at}~ \bm{x}_k\}$. For any $\bz$ satisfies \eqref{MFCQ_z} at $\bx_k$ with~$k$ large enough, we can follow the analysis of the firs part, define 
\begin{equation*}
\epsilon_k = \min\{0.5[\bu-\bell]_i, [\bx_k-\bell]_j, [\bu-\bx_k]_j: i\in\mA(\tx), j\in\mA(\tx)^-\},
\end{equation*}
and show $\Omega(\bx_k;\epsilon_k/\|\bz\|)\neq \emptyset$. Since $\epsilon_k\geq \epsilon$ for some $\epsilon>0$ independent of $k$, we know $\theta_k\rightarrow0$ implies $\limsup\limits_{k\rightarrow\infty} l_k = \infty$. This leads to a contradiction since $\infty = \limsup\limits_{k\rightarrow\infty} l_k \leq \|\bz^\star\|\eqqcolon l^{\star} < \infty$. Thus, EGMFCQ does not hold at $\bm{x}^{\star}$. This completes the proof of the second part of the theorem.

\subsection{Proof of Theorem \ref{thm:2}}\label{app:c2}

First, by Assumption \ref{ass:1}, we know $\tmu_{\mA^\star}>\0$. For any KKT triplet $(\bx,\blambda,\bmu)$ that is close to $(\tx,\tlambda,\tmu)$, the corresponding active and inactive sets coincide with those~at $(\tx,\tlambda,\tmu)$. In particular, for any $i\in\mA^\star = \mA_\bell^\star\cup\mA_\bu^\star$, we have $\bmu_{\mA^\star} > \0$ since $\tmu_{\mA^\star}>\0$, which implies $\mA^\star\subseteq \mA(\bx)$ due to the KKT conditions \eqref{nsequ:KKT} at $(\bx, \blambda, \bmu)$. On the other hand, since $\bell_{(\mA^\star)^-}<\tx_{(\mA^\star)^-}<\bu_{(\mA^\star)^-}$, the above inequalities also hold for $\bx_{(\mA^\star)^-}$, leading to $(\mA^\star)^-\subseteq \mA(\bx)^-$ and hence $\mA(\bx)\subseteq \mA^\star$. This shows $\mA(\bx)=\mA^\star$.

With this observation, we define the Lagrangian gradient with respect to $\bw = (\bx, \blambda, \bmu_{\mA^\star})\in\mR^{d+m+|\mA_\bell^\star|+|\mA_\bu^\star|}$ as
\begin{equation}\label{eq_A}
\nabla_{\bw}\mL(\bw) = \begin{pmatrix}
\nabla f(\bx) + \nabla \bc(\bx)^\top\blambda - \bI_{\mA_\bell^\star}^\top[\bmu_1]_{\mA_\bell^\star} + \bI_{\mA_\bu^\star}^\top[\bmu_2]_{\mA_\bu^\star}\\
\bc(\bx)\\
[\bell-\bx]_{\mA_\bell^\star}\\
[\bx-\bu]_{\mA_\bu^\star}
\end{pmatrix}\in \mR^{d+m+|\mA_\bell^\star|+|\mA_\bu^\star|},
\end{equation}
and define the mapping $\sigma(\bm{\delta}) = \left\{\bw: \nabla_{\bw}\mL(\bw) = \bm{\delta} \right\}$. This mapping characterizes the set of primal-dual points satisfying the perturbed optimality conditions up to a residual vector $\bm{\delta}$. In fact, $\sigma(\bm{\delta})$ is a single-valued mapping in a neighborhood of $(\bm{\delta}, \bw) = (\0, \tw)$ with $\nabla \sigma(\bm{\delta})|_{\bm{\delta} = \0} = \left(\bm{H}^{\star}\right)^{-1}$. To see this, let $\hat{\bw}, \bar{\bw}\in \sigma(\bdelta)$ for sufficiently small $\bdelta$. By SOSC and LICQ in Assumption \ref{ass:2} and Definition \ref{def_MFCQ}, we know $\nabla_\bw^2\mL(\tw) = \bH^\star$ is nonsingular. Thus, there exists~a~neighborhood of $\tw$ within which $\nabla_\bw^2\mL(\bw)$ remains nonsingular. We then have
\begin{equation}\label{pf:B2}
\b0 = \nabla_{\bw}\mL(\hat{\bw}) - \nabla_{\bw}\mL(\bar{\bw})  = \left(\int_{0}^{1} \nabla_{\bw}^2\mL\left(\bar{\bw} + t(\hat{\bw} - \bar{\bw})\right) dt\right) \cdot \left(\hat{\bw} - \bar{\bw}\right).
\end{equation}
Since both $\hat{\bw}$ and $\bar{\bw}$ lie sufficiently close to $\tw$, the integral matrix is nonsingular, which implies $\hat{\bw} - \bar{\bw} = \0$. Hence, $\sigma(\cdot)$ is locally single-valued. Furthermore, by \eqref{pf:B2} we have
\begin{equation*}
\bdelta = \nabla_{\bw}\mL(\hat{\bw}) - \nabla_{\bw}\mL(\tw) = \rbr{\int_{0}^{1} \nabla_{\bw}^2\mL\left(\tw + t(\hat{\bw} - \tw)\right) dt}\cdot \left(\hat{\bw} - \tw\right),
\end{equation*}
leading to
\begin{equation*}
\sigma(\bm{\delta}) = \hat{\bw} =  \tw + \left(\int_{0}^{1} \nabla_{\bw}^2 \mL\left(\tw + t(\hat{\bw} - \tw)\right) dt\right)^{-1} \bdelta.
\end{equation*}
Differentiating $\sigma(\bm{\delta})$ with respect to $\bm{\delta}$ at $\bm{\delta} = \0$ yields
\begin{equation*}
\nabla \sigma(\bm{\delta})|_{\bm{\delta} = \0} = \lim_{\bm{\delta} \to \0} \left(\int_{0}^{1} \nabla_{\bw}^2 \mL(\tw + t(\hat{\bw} - \tw)) dt\right)^{-1}  = \left(\nabla_{\bw}^2\mL(\tw) \right)^{-1} = \left(\bm{H}^{\star} \right)^{-1}.
\end{equation*}

Let us use $\mT$ to denote the corresponding sample space of the probability measure $\P$. We define the function class
\begin{equation*}
\mathcal{G} = \cbr{\bg: \mathcal{T} \to \mathbb{R}^{d+m+|\mA_\bell^\star|+|\mA_\bu^\star|}: \mathbb{E}_{\zeta \sim \mathcal{P}}\left[\bg(\zeta)\right] = \0, \mathbb{E}_{\zeta \sim \mathcal{P}}\left\|\bg(\zeta)\right\|^2 < \infty}.
\end{equation*}
Fix an arbitrary function $\bg \in \mathcal{G}$, and let $h: \mathbb{R} \to [-1,1]$ be any $C^3$-smooth function whose~first three derivatives are globally bounded and that satisfies $h(t) = t$ for $t \in [-1/2, 1/2]$. For each $\bv\in \mR^{d+m+|\mA_\bell^\star|+|\mA_\bu^\star|}$, we define a perturbed distribution whose density with respect to~$\mathcal{P}$~is
\begin{equation*}
d \mathcal{P}_{\bv}(\zeta) = \frac{1 + h(\bv^{\top} \bg(\zeta))}{C(\bv)} d \mathcal{P}(\zeta)\quad\quad \text{ where }\quad\quad C(\bv) = 1 + \int h(\bv^{\top} \bg(\zeta)) d \mathcal{P}(\zeta).
\end{equation*}
Each vector $\bv$ specifies a perturbed optimization problem
\begin{equation}\label{eq_perturbed}
\min_{\bm{x} \in \mathbb{R}^{d}} \hspace{1em} f_{\bv}(\bm{x}) = \mathbb{E}_{\zeta \sim \P_{\bv}} \left[ F(\bm{x}; \zeta) \right] \quad\quad \text{s.t.}  \quad \bm{c}(\bm{x}) = \bm{0},\quad \bm{\ell} \leq \bm{x} \leq \bm{u}.
\end{equation}
We further define the associated perturbed optimality mapping $\nabla_{\bw}\mL_{\bv}(\bw)$ analogously to \eqref{eq_A}, except that $\nabla f(\bx)$ is replaced by $\nabla f_{\bv}(\bx)$. We introduce the following two lemmas regarding~the mapping $\nabla_{\bw}\mL_{\bv}(\bw)$, which are proved in Appendices \ref{Appendix:pf:D1} and \ref{Appendix:pf:D2}, respectively.

\begin{lemma}\label{lemma_A_u}
The mapping $\nabla_{\bw}\mL_{\bv}(\bw)$ is continuously differentiable in a neighborhood of $(\bw, \bv) = (\tw, \0)$, with derivatives given by
\begin{equation}\label{pf:B4}
\nabla_{\bw}^2 \mL_{\bv}(\tw)|_{\bv = \0} = \nabla_{\bw}^2\mL(\tw) = \bm{H}^{\star},\quad \nabla_{\bv\bw} \mL_{\bv}(\bw)|_{\bv = \0} = \mathbb{E}_{\zeta \sim \P} \left[\nabla_{\bw}\mL(\bw; \zeta) \bg(\zeta)^{\top}\right],
\end{equation}
where $\nabla_{\bw}\mL(\bw; \zeta)$ has the same form as $\nabla_{\bw}\mL(\bw)$ in \eqref{eq_A} but replaces $\nabla f(\bx)$ by $\nabla F(\bx;\zeta)$.
\end{lemma}

\begin{lemma}\label{lemma_solution_map}
The solution mapping $S(\bv) = \left\{\bw:  \nabla_{\bw}\mL_{\bv}(\bw) = \0 \right\}$ admits a single-valued~localization $s(\bv)$ in a neighborhood of $(\bw, \bv) = (\tw, \0)$. Moreover, $s(\cdot)$ is differentiable at $\bv = \0$ with Jacobian $\nabla s(\0) = - \left(\bm{H}^{\star} \right)^{-1} \mathbb{E}_{\zeta \sim \P} [\nabla_{\bw}\mL(\tw; \zeta) \bg(\zeta)^{\top}]$.
\end{lemma}

With the above two lemmas, we then consider the mapping sequence $\Gamma_k(\bv) = s(\bv/\sqrt{k})$,~where $s(\cdot)$ is the single-valued localization of the solution map from Lemma~\ref{lemma_solution_map}. In fact, $\Gamma_k(\bv)$~is~regular with derivative $\nabla s(\0)$ (cf. Definition \ref{def_regular_mapping}). To see this, we just note that by the differentiability of $s(\bv)$ at $\bv = \0$, we have for all small enough $\bv$,
\begin{equation*}
\lim\limits_{k\rightarrow\infty}\sqrt{k}\left(\Gamma_{k}(\bv) - \Gamma_{k}(\0)\right) = \lim\limits_{k\rightarrow\infty}\sqrt{k}(s(\bv/\sqrt{k}) - s(\0)) = \nabla s(\0) \bv.
\end{equation*}
Now, we are ready to combine all above pieces to finalize the proof of Theorem \ref{thm:2}. For each $\bv\in \mR^{d+m+|\mA_\bell^\star|+|\mA_\bu^\star|}$, we define the product probability space $(\mT_k, \mathcal{F}_k, \mathcal{P}_{k,\bv}) := (\mT, \mathcal{F}, \mathcal{P}_{\bv/\sqrt{k}})^{\otimes k}$, that is, the $k$-fold product of $(\mT, \mathcal{F}, \mathcal{P}_{\bv/\sqrt{k}})$. It has been shown in~\citep[Lemma~8.3]{Duchi2021Asymptotic} that the sequence $(\mT_k, \mathcal{F}_k, \mathcal{P}_{k,\bv})$ is locally asymptotically normal with the precision matrix~$\bm{Q} = \mathbb{E}_{\zeta \sim \mathcal{P}} \left[\bg(\zeta) \bg(\zeta)^{\top}\right]$. Let us specify
\begin{equation*}
\bg(\zeta) = \nabla_{\bw}\mL(\tw;\zeta) - \nabla_{\bw}\mL(\tw) = \nabla_{\bw}\mL(\tw;\zeta),
\end{equation*}
then we have $\bSigma^\star=\mathbb{E}_{\zeta \sim \mathcal{P}} \left[\bg(\zeta) \bg(\zeta)^{\top}\right]$ (see \eqref{def:cov}). Let $\ell$ be any symmetric, quasiconvex, and lower semicontinuous function, and let $\bw_k$ denote any sequence of estimators, and $\tw_{\bv}$ denote the primal and (active) dual solution of the perturbed problem~\eqref{eq_perturbed} for small enough $\bv$, i.e., $\tw_{\bv} = s(\bv)$. Note that the active set is consistent with the unperturbed active set $\mA^\star$. For any finite subset $\mathcal{W}_0 \subseteq \mR^{d+m+|\mA_\bell^\star|+|\mA_\bu^\star|}$, we define $c = \max_{\bv \in \mathcal{W}_0} \|\bv\|$.  Since $\tw_{\bv} = s(\bv)$ and~$\Gamma_{k}(\bv) = s(\bv/\sqrt{k})$, we have 
\begin{equation*}
\liminf_{k \to \infty} \sup_{\|\bv\| \leq c/\sqrt{k}} \mathbb{E}_{\mathcal{P}_{\bv}^{\otimes k}} [\ell(\sqrt{k}(\bw_{k} - \tw_{\bv}))] \geq \liminf_{k \to \infty} \max_{\bv \in \mathcal{W}_0} \mathbb{E}_{\mathcal{P}_{k,\bv}} [\ell( \sqrt{k}( \bw_k - \Gamma_{k}(\bv)))].
\end{equation*}
Taking the supremum over all finite $\mathcal{W}_0 \subseteq \mR^{d+m+|\mA_\bell^\star|+|\mA_\bu^\star|}$ (equivalently, letting $c \to \infty$), and applying the H\'ajek and Le Cam's local minimax theorem in Theorem \ref{theorem_hajek_lecam}, we obtain
\begin{equation*}
\lim_{c \to \infty} \liminf_{k \to \infty} \sup_{\|\bv\| \leq c / \sqrt{k}} \mathbb{E}_{\mathcal{P}_{\bv}^{\otimes k}} [\ell(\sqrt{k}(\bw_{k} - \tw_{\bv}))] \geq \mathbb{E}\left[\ell(Z_{\gamma})\right],
\end{equation*}
where $Z_{\gamma} \sim \mathcal{N}(\0, \left(\bm{H}^{\star}\right)^{-1} \bm{\Sigma}^\star (\bm{\Sigma}^\star + \gamma \bI)^{-1} \bm{\Sigma}^\star \left(\bm{H}^{\star}\right)^{-1})$ for any $\gamma > 0$. Since $\lim_{\gamma \to 0} \bm{\Sigma}^\star (\bm{\Sigma}^\star + \gamma \bI)^{-1} \bm{\Sigma}^\star  = \bm{\Sigma}^\star$, the right-hand side converges to $\mathbb{E}\left[\ell(Z)\right]$ with $Z \sim \mathcal{N}(\0, \bm{\Omega}^{\star})$. For the left-hand side, it has been shown in \cite{Duchi2021Asymptotic} that $\P_{\bv}$ with $\|\bv\|\leq c/\sqrt{k}$ implies~$\P_{\bv}\in \mB(\P;c'/k)$ for some $c'>0$. This establishes the claimed lower bound.

\subsection{Proof of Lemma \ref{lem:1}}\label{app:c3}

We first state a preliminary lemma. The proof is omitted, as it follows directly from the observation that $\bx + \alpha \Delta\bx = (1 - \alpha)\bx + \alpha(\bx + \Delta\bx)$ and that the box constraint set is convex.$\quad\quad$

\begin{lemma}\label{lem:2}
Suppose $\bell\leq \bx\leq \bu$ and $\bell\leq \bx+\Delta\bx\leq \bu$.~Then, for any $\alpha\in[0, 1]$, $\bell\leq \bx+\alpha\Delta\bx\\ \leq \bu$. In particular, if $\bell\leq \bx_0\leq \bu$, then the entire SSQP sequence $\{\bx_k\}$ satisfies~$\bell\leq \bx_k\leq \bu$.
\end{lemma}

The KKT conditions for the SQP subproblem at $\bx_k$ with true gradient $\nabla f_k$ show that there exist some dual multipliers $(\blambda_k^{\text{sub}}, \bmu_{k}^{\text{sub}})$ satisfying
\begin{equation}\label{eq6}
\begin{split}
& \nabla f_{k} + \barB_k \Delta\bx_k + \nabla \bm{c}_{k}^{\top} \bm{\lambda}_k^{\text{sub}} - \bm{\mu}_{1,k}^{\text{sub}} + \bm{\mu}_{2,k}^{\text{sub}} = \bm{0},\\
& \theta_k \bm{c}_{k}+ \nabla \bm{c}_{k}\Delta\bx_k = \bm{0}, \quad\quad\quad\quad \bm{\ell} \leq \bm{x}_k + \Delta\bx_k \leq \bm{u},\\
& \bm{\mu}_{1,k}^{\text{sub} \top}(\bell - \bm{x}_k - \Delta\bx_k) = 0, \quad\;  \bm{\mu}_{2,k}^{\text{sub} \top}(\bm{x}_k + \Delta\bx_k - \bm{u}) = 0,\\
& \bm{\mu}_{1,k}^{\text{sub}}, \bm{\mu}_{2,k}^{\text{sub}} \geq \bm{0}.
\end{split}
\end{equation}
Multiplying both sides of the first equality by $\Delta\bx_k$, we obtain
\begin{align}\label{pf:C7}
\nabla f_{k}^{\top}\Delta\bx_k + \Delta\bx_k^{\top} \barB_k \Delta\bx_k & = -\Delta\bx_k^{\top}\nabla \bm{c}_{k}^{\top} \bm{\lambda}_k^{\text{sub}} + \Delta\bx_k^{\top}\bm{\mu}_{1,k}^{\text{sub}} - \Delta\bx_k^{\top}\bm{\mu}_{2,k}^{\text{sub}} \nonumber\\
& \stackrel{\mathclap{\eqref{eq6}}}{=}\; \theta_k \bm{\lambda}_k^{\text{sub} \top} \bm{c}_{k} - \bm{\mu}_{1,k}^{\text{sub} \top}(\bm{x}_k  - \bm{\ell}) + \bm{\mu}_{2,k}^{\text{sub} \top}(\bm{x}_k  - \bm{u}) \nonumber\\
& \leq \theta_k \bm{\lambda}_k^{\text{sub} \top} \bm{c}_{k} \leq \|\bm{\lambda}_k^{\text{sub}}\| \|\bm{c}_{k}\| \leq M_{\text{dual}}\|\bm{c}_{k}\|,
\end{align}
where the third inequality comes from $\bm{\mu}_{1,k}^{\text{sub}}, \bm{\mu}_{2,k}^{\text{sub}} \geq \bm{0}$ and $\bm{\ell} \leq \bm{x}_k \leq \bm{u}$ (by Lemma \ref{lem:2}).~By~the definitions \eqref{nsequ:vap:loc} and \eqref{nsequ:vap:loc:diff}, we have
\begin{equation}\label{eq_30}
\Delta \varphi_\rho^{\text{loc}}(\bx_k,\Delta\bx_k,\barB_k) = -\nabla f_k^\top \Delta\bx_k - \frac{1}{2} \Delta\bx_k^{\top}\barB_k\Delta\bx_k +  \rho \theta_k \left\|\bm{c}_k \right\|.
\end{equation}
Thus, to satisfy $\Delta \varphi_\rho^{\text{loc}}(\bx_k,\Delta\bx_k,\barB_k) \geq \frac{1}{2} \Delta\bx_k^{\top} \barB_k \Delta\bx_k + \nu \rho\theta_k\| \bm{c}_{k}\|$, it suffices to satisfy
\begin{equation*}
\nabla f_{k}^{\top}\Delta\bx_k + \Delta\bx_k^{\top} \barB_k \Delta\bx_k \leq (1-\nu)\rho \theta_k\| \bm{c}_{k}\|.
\end{equation*}	
Combining with \eqref{pf:C7}, we know the above display holds if 
\begin{equation*}
M_{\text{dual}}\|\bm{c}_{k}\| \leq (1-\nu)	\rho \theta_k\left\| \bm{c}_{k}\right\|,
\end{equation*}
which is further implied by the stated condition $\rho \geq \bar{\rho} \coloneqq \frac{M_{\text{dual}}}{(1-\nu) \tau \bar{\theta}}$ since $\theta_k \geq \tau \bar{\theta}$ by Assumption \ref{assump1}. This completes the proof.

\subsection{Proof of Theorem \ref{thm:3}}\label{app:c4}

To simplify notation, we slightly abuse the symbol $\mE_k[\cdot]$ throughout the proof to denote the~conditional expectation given the randomness of $\bx_0$ through $\bx_k$. Since $\baralpha_k\leq \alpha_k+\psi\alpha_k^p$ and $\theta_k\leq 1$, we may, without loss of generality, assume that $\baralpha_k\theta_k\leq 1$ (otherwise, we can simply restrict~our analysis for $k\geq k_0$ for some deterministic $k_0$). For any $\rho > 0$, it follows from Assumptions \ref{assump1}, \ref{assump2}, and \ref{assump3} that
\begin{align*}
& \varphi_{\rho}(\bm{x}_{k+1}) - \varphi_{\rho}(\bm{x}_{k})\\
& =  f(\bm{x}_k+\baralpha_k \barDelta\bx_k) - f(\bx_k) + \rho(\| \bm{c}(\bm{x}_k + \baralpha_k \barDelta\bx_k)\| - \|\bm{c}_{k}\| ) \\
& \leq \baralpha_k \nabla f_{k}^{\top} \barDelta\bx_k + \frac{\kappa_{\nabla f}}{2} \baralpha_k^2 \|\barDelta\bx_k\|^2 + \rho\rbr{\| \bm{c}_{k} + \baralpha_k \nabla \bm{c}_{k} \barDelta\bx_k\| - \|\bm{c}_{k} \| + \frac{\kappa_{\nabla c}}{2} \baralpha_k^2 \|\barDelta\bx_k\|^2 } \\
& = \baralpha_k \nabla f_{k}^{\top} \barDelta\bx_k - \baralpha_k \rho\theta_k \left\| \bm{c}_{k} \right\| + \frac{\kappa_{\nabla f} + \rho \kappa_{\nabla c}}{2} \baralpha_k^2 \|\barDelta\bx_k\|^2 \\
& = \baralpha_k \nabla f_{k}^{\top} \Delta\bx_k + \baralpha_k \nabla f_{k}^{\top} \left(\barDelta\bx_k - \Delta\bx_k \right) - \baralpha_k \rho\theta_k \left\| \bm{c}_{k} \right\| + \frac{\kappa_{\nabla f} + \rho \kappa_{\nabla c}}{2} \baralpha_k^2 \|\barDelta\bx_k\|^2 \\
& \stackrel{\mathclap{\eqref{eq_30}}}{=} -\baralpha_k \Delta \varphi_{\rho}^{\text{loc}}(\bm{x}_k,\Delta\bx_k,\barB_k) -\frac{\baralpha_k}{2}\Delta\bx_k^{\top} \barB_k \Delta\bx_k + \baralpha_k \nabla f_{k}^{\top} \left(\barDelta\bx_k - \Delta\bx_k \right) + \frac{\kappa_{\nabla f} + \rho \kappa_{\nabla c}}{2} \baralpha_k^2 \|\barDelta\bx_k\|^2 \\
& \leq -\baralpha_k \Delta \varphi_{\rho}^{\text{loc}}(\bm{x}_k,\Delta\bx_k,\barB_k) + \baralpha_k \nabla f_{k}^{\top} \left(\barDelta\bx_k - \Delta\bx_k\right) + \frac{\kappa_{\nabla f} + \rho\kappa_{\nabla c}}{2} \baralpha_k^2 \|\barDelta\bx_k\|^2  \\
& \leq - \baralpha_k  \Delta \varphi_{\rho}^{\text{loc}}(\bm{x}_k,\Delta\bx_k,\barB_k) + \baralpha_k M_{\nabla f} \|\barDelta\bx_k - \Delta\bx_k \| + \frac{\kappa_{\nabla f} +  \Bar{\rho} \kappa_{\nabla c}}{2}\baralpha_k^2 M_{\bm{\ell},\bm{u}}^2,
\end{align*}
where the last inequality holds for some constants $M_{\nabla f}, M_{\bm{\ell},\bm{u}}^2>0$ since box constraints form~a compact set. Setting $\rho = \bar{\rho}$, we apply Lemma \ref{lem:1}, Assumptions \ref{assump1} and \ref{assump2}, and obtain
\begin{equation*}
\Delta \varphi_{\bar{\rho}}^{\text{loc}}(\bm{x}_k,\Delta\bx_k,\barB_k) \geq \frac{1}{2} \Delta\bx_k^{\top} \barB_k \Delta\bx_k + \nu \bar{\rho} \tau\Bar{\theta}\| \bm{c}_{k}\| \geq \frac{\kappa_1}{2} \| \Delta\bx_k\|^2 + \nu \bar{\rho} \tau \Bar{\theta}\| \bm{c}_{k}\|.
\end{equation*}
Combining the above two displays and taking conditional expectation over $\bx_0$ to $\bx_k$, we obtain 
\begin{align}\label{eq24}
& \mathbb{E}_{k} \left[ \varphi_{\Bar{\rho}}(\bm{x}_{k+1}) - \min_{\bm{\ell} \leq \bm{x} \leq  \bm{u}}\varphi_{\Bar{\rho}}(\bm{x}) \right] \leq  \cbr{\varphi_{\Bar{\rho}}(\bm{x}_{k}) - \min_{\bm{\ell} \leq \bm{x} \leq \bm{u}}\varphi_{\Bar{\rho}}(\bm{x})} -\alpha_k \mathbb{E}_k \sbr{\frac{\kappa_1}{2} \| \Delta\bx_k\|^2 + \nu \bar{\rho} \tau \Bar{\theta}\| \bm{c}_{k}\|} \nonumber \\
& + M_{\nabla f} \left(\alpha_k + \psi \alpha_k^p \right) \mathbb{E}_k [ \|\barDelta\bx_k - \Delta\bx_k \|] + \frac{\kappa_{\nabla f} + \Bar{\rho} \kappa_{\nabla c}}{2} M_{\bm{\ell},\bm{u}}^2 \left(\alpha_k + \psi \alpha_k^p \right)^2, \quad \forall k\geq 0.
\end{align} 
Now, we introduce the following lemma, which is proved in Appendix \ref{app:d3}.

\begin{lemma}\label{lem:3}
Under the conditions of Theorem \ref{thm:3}, we have $\sum_{k=0}^{\infty} \alpha_k \mathbb{E} [\|\barDelta\bx_k  - \Delta\bx_k \| ] < \infty$.
\end{lemma}

With the above lemma, we directly have $\mathbb{E} [\sum_{k=0}^{\infty} \alpha_k \mathbb{E}_k [\|\barDelta\bx_k - \Delta\bx_k \|]] = \sum_{k=0}^{\infty} \alpha_k \mathbb{E} [\|\barDelta\bx_k - \Delta\bx_k \|] < \infty$, which implies that $\sum_{k=0}^{\infty} \left(\alpha_k + \psi \alpha_k^p \right) \mathbb{E}_k [\|\barDelta\bx_k - \Delta\bx_k\| ] < \infty$ almost surely. From \eqref{eq24}, together with the condition $\sum_{k=0}^{\infty} \alpha_k^2 < \infty$, we apply the Robbins-Siegmund theorem \citep{Robbins1971convergence} and obtain $\sum_{k=0}^{\infty} \alpha_k(\|\Delta\bx_k\|^2+\|\bc_k\|) <\infty$. This completes~the proof by noting that $\sum_{k=0}^{\infty}\alpha_k=\infty$.

\subsection{Proof of Theorem \ref{thm:4}}\label{app:c5}

Let us denote the objective function of Problem \eqref{est_opt_lagrangian} by
\begin{equation*}
G(\blambda, \bmu; \bm{x}_k) = \|\nabla_{\bx}\mL(\bx_k,\blambda,\bmu)\|^2 + \| \bmu_{1} \odot (\bell- \bx_k)\|^2 + \| \bmu_{2} \odot (\bx_k - \bu) \|^2.
\end{equation*}
By the definition of the KKT residual $\bm{R}(\bx_k,\blambda_k^{\star},\bmu_k^{\star})$ in \eqref{equ:KKT}, it suffices to show $G(\bm{\lambda}_{k}^{\star}, \bm{\mu}_{k}^{\star}; \bm{x}_k)\rightarrow 0$ and $\|\bc_k\|\rightarrow 0$ as $k\rightarrow\infty$ almost surely. 

\noindent$\bullet$ \textbf{Convergence of $G(\bm{\lambda}_{k}^{\star}, \bm{\mu}_{k}^{\star}; \bm{x}_k)$.} 
Recall from Assumption \ref{assump2} that $(\bm{\lambda}_{k}^{\text{sub}}, \bm{\mu}_{k}^{\text{sub}})$ denotes the dual multipliers of the SQP subproblem \eqref{equ:SQP:new} with the true gradient $\nabla f_{k}$. It follows from the~KKT conditions in \eqref{eq6} that
\begin{align}\label{equ:C11}
G(\bm{\lambda}_{k}^{\star}, \bm{\mu}_{k}^{\star}; \bm{x}_k) & \leq G(\bm{\lambda}_{k}^{\text{sub}}, \bm{\mu}_{k}^{\text{sub}}; \bm{x}_k) \leq \left\| \bm{B}_k \Delta\bx_k \right\|^2 + \left\| \bm{\mu}_{1,k}^{\text{sub}} \odot \Delta\bx_k \right\|^2 + \left\| \bm{\mu}_{2,k}^{\text{sub}} \odot \Delta\bx_k \right\|^2 \nonumber \\
& \leq (\kappa_2^2 + 2 M_{\text{dual}}^2) \left\| \Delta\bx_k \right\|^2,
\end{align}
which implies from Theorem \ref{thm:3} that $\liminf\limits_{k \to \infty}  G(\bm{\lambda}_{k}^{\star}, \bm{\mu}_{k}^{\star}; \bm{x}_k)  = 0$. Suppose $\limsup\limits_{k \to \infty}  G(\bm{\lambda}_{k}^{\star}, \bm{\mu}_{k}^{\star}; \bm{x}_k)  > 0$, we can find a sufficiently small number $\varepsilon >0$ and two infinite sequences $\{m_i, n_i\}$ with~$m_i < n_i$, such that
\begin{equation*}
G(\bm{\lambda}_{m_i}^{\star}, \bm{\mu}_{m_i}^{\star}; \bm{x}_{m_i}) > 2 \varepsilon, \quad \left\| \Delta\bx_k \right\| \geq \sqrt{\frac{\varepsilon}{\kappa_2^2 + 2M_{\text{dual}} ^2}}~\text{ for } k\in[m_i,n_i), \quad \left\| \Delta\bx_{n_i} \right\| \leq \sqrt{\frac{\varepsilon}{\kappa_2^2 + 2M_{\text{dual}}^2}}.
\end{equation*}
We can always find such sequences since $G(\bm{\lambda}_{m_i}^{\star}, \bm{\mu}_{m_i}^{\star}; \bm{x}_{m_i}) > 2 \varepsilon$ implies $\|\Delta\bx_{m_i}\| \geq \sqrt{\frac{2\varepsilon}{\kappa_2^2 + 2M_{\text{dual}}^2 }}$; and due to $\liminf\limits_{k \to \infty} \| \Delta\bx_k \| = 0$, there must exist $n_i > m_i$ such that $\left\| \Delta\bx_{n_i} \right\| \leq \sqrt{\frac{\varepsilon}{\kappa_2^2 + 2M_{\text{dual}}^2}}$.~Let
\begin{equation*}
\widetilde{G}(\bm{\lambda}, \bm{\mu}; \bm{x}) = G(\bm{\lambda}, \bm{\mu}; \bx) + \frac{\varepsilon}{6 M_{\text{dual}}^2 } \left\|(\bm{\lambda}, \bm{\mu}) \right\|^2
\end{equation*}
and denote $\bm{w}_k = (\tilde{\bm{\lambda}}_{k}, \tilde{\bm{\mu}}_{k}) \in \arg\min_{\bm{\lambda}, \bm{\mu} \geq \bm{0}} \widetilde{G}(\bm{\lambda}, \bm{\mu}; \bm{x}_{k})$. 
By the above definition of $\tilde{G}$ and the construction of the sequences $\{m_i, n_i\}$, we have 
\begin{equation}\label{pf:equ:C11}
\begin{split}
\widetilde{G}(\Tilde{\bm{\lambda}}_{m_i}, \Tilde{\bm{\mu}}_{m_i}; \bm{x}_{m_i}) & \geq G(\Tilde{\bm{\lambda}}_{m_i}, \Tilde{\bm{\mu}}_{m_i}; \bm{x}_{m_i})\geq G(\bm{\lambda}_{m_i}^{\star}, \bm{\mu}_{m_i}^{\star}; \bm{x}_{m_i}) \geq 2 \varepsilon,\\
\widetilde{G}(\Tilde{\bm{\lambda}}_{n_i}, \Tilde{\bm{\mu}}_{n_i}; \bm{x}_{n_i}) & \leq \widetilde{G}(\bm{\lambda}_{n_i}^{\text{sub}}, \bm{\mu}_{n_i}^{\text{sub}}; \bm{x}_{n_i}) \stackrel{\eqref{equ:C11}}{\leq} (\kappa_2^2 + 2 M_{\text{dual}}^2) \left\| \Delta\bx_{n_i} \right\|^2 + \frac{\varepsilon}{3} \leq \frac{4}{3} \varepsilon.
\end{split}
\end{equation}
Next, we analyze the consecutive difference of $\tilde{G}$. In particular, let us write $\widetilde{G}(\bm{\lambda}, \bm{\mu}; \bm{x}_k)$ into~a quadratic form of $\bw = (\blambda, \bmu) = (\blambda, \bmu_1, \bmu_2)$:
\begin{equation*}
\widetilde{G}(\bm{\lambda}, \bm{\mu}; \bm{x}_{k}) = \left\| \nabla f_{k} \right\|^2 +  \bm{q}_{k}^{\top} \bm{w} + \frac{1}{2} \bm{w}^{\top} \nabla^2\tilde{G}_k \bm{w},
\end{equation*}
where
\begin{equation*}
\bm{q}_{k} = \left( \begin{array}{c}
2 \nabla \bm{c}_{k} \nabla f_{k}   \\
-2  \nabla f_{k} \\
2 \nabla f_{k} 
\end{array} \right)
\end{equation*}
and
\begin{equation*}
\nabla^2\tilde{G}_k = \left( \begin{array}{ccc}
2 \nabla \bm{c}_{k} \nabla \bm{c}_{k}^{\top}  & -2 \nabla \bm{c}_{k} & 2 \nabla \bm{c}_{k} \\
-2 \nabla \bm{c}_{k}^{\top} & 2 \bm{I} + 2 \text{diag}^2\left( \bm{\ell} -  \bm{x}_{k}\right) & -2 \bm{I} \\
2 \nabla \bm{c}_{k}^{\top} & -2 \bm{I} & 2 \bm{I} + 2 \text{diag}^2\left( \bm{x}_{k} - \bm{u} \right) 
\end{array} \right) + \frac{\varepsilon}{3M_{\text{dual}}^2 } \bm{I}.
\end{equation*}
Then, we have for any $k\geq 0$,
\begin{align}\label{equ:diff}
&\hskip-0.7cm |\widetilde{G}(\Tilde{\bm{\lambda}}_{k+1}, \Tilde{\bm{\mu}}_{k+1}; \bm{x}_{k+1}) -  \widetilde{G}(\Tilde{\bm{\lambda}}_{k}, \Tilde{\bm{\mu}}_{k}; \bm{x}_{k})| \nonumber\\
&\hskip-0.7cm \leq \left| \bm{q}_{k+1}^{\top} \bm{w}_{k+1} + \frac{1}{2} \bm{w}^{\top}_{k+1} \nabla^2\tilde{G}_{k+1} \bm{w}_{k+1} - \bm{q}_{k}^{\top} \bm{w}_{k} - \frac{1}{2} \bm{w}^{\top}_{k} \nabla^2\tilde{G}_{k} \bm{w}_{k} \right| + \left| \| \nabla f_{k+1} \|^2 - \| \nabla f_{k} \|^2 \right| \nonumber\\
&\hskip-0.7cm \leq \left| \bm{q}_{k+1}^{\top} \bm{w}_{k+1} + \frac{1}{2} \bm{w}^{\top}_{k+1} \nabla^2\tilde{G}_{k+1} \bm{w}_{k+1} - \bm{q}_{k}^{\top} \bm{w}_{k+1} - \frac{1}{2} \bm{w}^{\top}_{k+1} \nabla^2\tilde{G}_{k} \bm{w}_{k+1} \right| \nonumber\\
&\hskip-0.7cm\quad + \left|\bm{q}_{k}^{\top} \bm{w}_{k+1} + \frac{1}{2} \bm{w}^{\top}_{k+1} \nabla^2\tilde{G}_{k} \bm{w}_{k+1} - \bm{q}_{k}^{\top} \bm{w}_{k} - \frac{1}{2} \bm{w}^{\top}_{k} \nabla^2\tilde{G}_{k} \bm{w}_{k} \right| + \left| \| \nabla f_{k+1} \|^2 - \| \nabla f_{k}\|^2 \right| \nonumber\\
&\hskip-0.7cm \leq \|\bm{w}_{k+1}\| \| \bm{q}_{k+1} - \bm{q}_{k}\|  + \frac{1}{2} \| \bm{w}_{k+1}\|^2 \| \nabla^2\tilde{G}_{k+1} - \nabla^2\tilde{G}_{k}\| + \| \bm{q}_{k}\| \| \bm{w}_{k+1} - \bm{w}_{k}\| \nonumber\\
&\hskip-0.7cm\quad + \frac{1}{2} (\| \bm{w}_{k+1}\|+\| \bm{w}_k\|)  \| \nabla^2\tilde{G}_{k}\|  \| \bm{w}_{k+1} - \bm{w}_{k}\|  +  \left\| \nabla f_{k+1} - \nabla f_{k} \right\|    \left( \left\| \nabla f_{k+1} \right\|  + \left\| \nabla f_{k} \right\|\right).
\end{align}
We provide the bound for each of term in the above display. We note that for any $k\geq 0$,$\quad\quad\quad$
\begin{multline*}
\frac{\varepsilon}{6 M_{\text{dual}}^2 } \|\bm{w}_k \|^2 \leq  \widetilde{G}(\Tilde{\bm{\lambda}}_k, \Tilde{\bm{\mu}}_k; \bm{x}_k) \\
\leq G(\bm{\lambda}_{k}^{\text{sub}}, \bm{\mu}_{k}^{\text{sub}}; \bx_k) + \frac{\varepsilon}{6 M_{\text{dual}}^2 } \|(\bm{\lambda}_{k}^{\text{sub}}, \bm{\mu}_{k}^{\text{sub}}) \|^2
\stackrel{\eqref{equ:C11}}{\leq} (\kappa_2^2 + 2 M_{\text{dual}}^2) \| \Delta\bx_k \|^2 + \frac{\varepsilon}{3}.
\end{multline*}
Thus, we have
\begin{equation}\label{eq22}
\|\bm{w}_k\| \leq \sqrt{\frac{6 M_{\text{dual}}^2 \left( \kappa_2^2 + 2 M_{\text{dual}}^2 \right) M_{\bm{\ell},\bm{u}}^{2} }{\varepsilon}} + \sqrt{2}M_{\text{dual}}.
\end{equation}
Furthermore, by the smoothness of the objective $f(\bm{x})$ and the constraints $\bm{c}(\bm{x})$, as well as~the boundedness of $\nabla\bc, \nabla f, \barDelta\bx$, we know there exists a constant $M_{\bq, \nabla^2\tilde{G}}>0$ such that for $k\geq 0$,
\begin{equation}\label{eq20}
\max\{\| \bm{q}_{k+1} - \bm{q}_{k}\|, \| \nabla^2\tilde{G}_{k+1} - \nabla^2\tilde{G}_k\| \} \leq M_{\bq, \nabla^2\tilde{G}}\cdot\alpha_k.
\end{equation}
Finally, we note that for any $k\geq 0$ and any vector $\bm{w} = (\blambda, \bmu_1, \bmu_2) \in \mathbb{R}^{m} \times \mathbb{R}^{d} \times \mathbb{R}^{d}$, 
\begin{equation*}
\frac{1}{2}\bm{w}^{\top} \rbr{\nabla^2\tilde{G}_k  - \frac{\varepsilon}{3M_{\text{dual}}^2 } \bm{I}} \bm{w} = \| \nabla \bm{c}_k^\top \blambda - \bmu_1 + \bmu_2\|^2 + \| \left(\bm{x}_k - \bm{\ell} \right) \odot \bmu_1\|^2 + \| \left( \bm{x}_k - \bm{u} \right) \odot \bmu_2 \|^2 \geq 0.
\end{equation*}
We combine the above display with \eqref{eq22} and \eqref{eq20}, apply Lemma \ref{lemma13}, and have $\|\bm{w}_{k+1} - \bm{w}_{k}\| = \mathcal{O}\left( \frac{\alpha_k}{\varepsilon^{3/2}} \right)$, where we have omitted universal deterministic constants that are independent of $\varepsilon$.
Combining this result with \eqref{eq22} and \eqref{eq20}, and plugging into \eqref{equ:diff}, we obtain$\quad\quad\quad$
\begin{equation*}
|  \widetilde{G}(\Tilde{\bm{\lambda}}_{k+1}, \Tilde{\bm{\mu}}_{k+1}; \bm{x}_{k+1}) -  \widetilde{G}(\Tilde{\bm{\lambda}}_{k}, \Tilde{\bm{\mu}}_{k}; \bm{x}_{k})| \leq M_{G} \frac{\alpha_k}{\varepsilon^2},\quad\quad\quad \forall k\geq 0,
\end{equation*}
for some universal constant $M_{G}>0$ that is independent of $\alpha_k$ and $\varepsilon$. Combining the above~display with \eqref{pf:equ:C11}, we obtain
\begin{multline*}
\frac{2}{3} \varepsilon \leq \widetilde{G}(\Tilde{\bm{\lambda}}_{m_i}, \Tilde{\bm{\mu}}_{m_i}; \bm{x}_{m_i}) - \widetilde{G}(\Tilde{\bm{\lambda}}_{n_i}, \Tilde{\bm{\mu}}_{n_i}; \bm{x}_{n_i}) \\
\leq \sum_{k=m_i}^{n_i -1} \left |\widetilde{G}(\Tilde{\bm{\lambda}}_{k}, \Tilde{\bm{\mu}}_{k}; \bm{x}_{k}) - \widetilde{G}(\Tilde{\bm{\lambda}}_{k+1}, \Tilde{\bm{\mu}}_{k+1}; \bm{x}_{k+1}) \right|  \leq \frac{M_G}{\varepsilon^2} \sum_{k=m_i}^{n_i -1} \alpha_k.
\end{multline*}
Summing up both sides from $i=1$ to $\infty$, we have
\begin{equation*}
\infty = \sum_{i=1}^{\infty} \frac{2\varepsilon^3}{3M_{G}} \leq  \sum_{i=1}^{\infty} \sum_{k=m_i}^{n_i -1} \alpha_k.
\end{equation*}
On the other hand, since $\| \Delta\bx_{k}\| \geq \sqrt{\frac{\varepsilon}{\kappa_2^2 + 2M_{\text{dual}}^2 }}$ for $m_i \leq k \leq n_i-1$, we know that
\begin{equation*}
\sum_{i=1}^{\infty} \sum_{k=m_i}^{n_i -1}  \alpha_k  \leq \frac{ \kappa_2^2 + 2M_{\text{dual}}^2 }{\varepsilon } \sum_{i=1}^{\infty} \sum_{k=m_i}^{n_i -1} \alpha_k  \left\| \Delta\bx_k \right\|^2 \leq \frac{ \kappa_2^2 + 2M_{\text{dual}}^2 }{\varepsilon } \sum_{k=0}^{\infty} \alpha_k  \left\| \Delta\bx_k \right\|^2 < \infty,
\end{equation*}
where the last inequality is due to the proof of Theorem \ref{thm:3}. This derives the contradiction~and we complete the proof of $\lim\limits_{k \to \infty} G(\bm{\lambda}_{k}^{\star}, \bm{\mu}_{k}^{\star}; \bm{x}_k) = 0$.

\noindent$\bullet$ \textbf{Convergence of $\|\bc_k\|$.}
We follow the same proof strategy. By Theorem \ref{thm:3} we know~that $\liminf\limits_{k \to \infty} \left\| \bm{c}_{k} \right\| = 0$. Suppose $\limsup\limits_{k \to \infty} \left\| \bm{c}_{k} \right\| > 0$, then we can find a sufficiently small number $\varepsilon > 0$ and two infinite sequences $\{m_i, n_i\}$ with $m_i < n_i$, such that 
\begin{equation*}
\|\bm{c}_{m_i}\| > 2 \varepsilon, \quad\quad \|\bm{c}_{k}\| \geq \varepsilon~\text{ for }~k\in[m_i, n_i), \quad\quad \|\bm{c}_{n_i}\| \leq \varepsilon.
\end{equation*}
It follows from the construction of the sequences that 
\begin{align*}
\varepsilon & \leq \| \bm{c}_{m_i}\| - \| \bm{c}_{n_i} \| = \sum_{k=m_i}^{n_i-1}(\| \bm{c}_{k} \| - \| \bm{c}_{k+1} \|) \leq \sum_{k=m_i}^{n_i-1}   \left\| \bm{c}_{k} - \bm{c}_{k+1} \right\|  \leq M_c \sum_{k=m_i}^{n_i-1} \alpha_{k}
\end{align*}
for some universal constant $M_c>0$ that is independent of $\alpha_k$ and $\varepsilon$. Multiplying both sides~by $\varepsilon$ and noting the fact that $\| \bm{c}_{k}\| \geq \varepsilon$ for $m_i \leq k < n_i$, we have
\begin{equation*}
\varepsilon^2 \leq M_c \sum_{k=m_i}^{n_i-1} \alpha_{k} \| \bm{c}_{k} \|,
\end{equation*}
which implies that $\infty \leq \sum_{i=1}^{\infty} \sum_{k=m_i}^{n_i-1} \alpha_{k} \left\| \bm{c}_{k} \right\| \leq \sum_{k=0}^{\infty} \alpha_{k} \left\| \bm{c}_{k} \right\| < \infty$. This leads to the contradiction and we complete the proof.

\subsection{Proof of Lemma \ref{lemma_almost_convg}}\label{app:C6}

\noindent $\bullet$ \textbf{Proof of (a).} 
By Assumption \ref{assump5}, we know LICQ holds at $\bx^{\star}$. Thus, when $\bx_k$ is close to $\tx$, the rows of $\tilde{\bJ}_{k} = (\nabla\bc_{k};-\bI_{\mA_{\bell}^{\star}};\bI_{\mA_{\bu}^{\star}})$ are linearly independent. To show $\theta_k=1$ for $k$ large enough, it suffices to show that there exists $\bz_k\in\mR^d$ such that $\bc_k+\nabla \bc_k \bz_k = \0$ and $\bell\leq \bx_k+\bz_k\leq \bu$. To this end, let us define the linear system at $\bx_k$ as
\begin{equation*}
\begin{pmatrix}
\nabla\bc_{k} \\
-\bI_{\mA_{\bell}^{\star}}\\
\bI_{\mA_{\bu}^{\star}}
\end{pmatrix}\bz_k = -\begin{pmatrix}
\bc_k\\
[\bell-\bx_{k}]_{\mA_{\bell}^{\star}}\\
[\bx_k-\bu]_{\mA_{\bu}^{\star}}
\end{pmatrix}.
\end{equation*}
By the full row-rankness of $\tilde{\bJ}_{k}$, the above linear system has a solution given by
\begin{equation*}
\bz_k = -\Tilde{\bJ}_{k}^{\top}(\Tilde{\bJ}_{k}\Tilde{\bJ}_{k}^{\top})^{-1} \begin{pmatrix}
\bc_k\\
[\bell-\bx_{k}]_{\mA_{\bell}^{\star}}\\
[\bx_k-\bu]_{\mA_{\bu}^{\star}}
\end{pmatrix}.
\end{equation*}
Clearly, the above $\bz_k$ satisfies $\bc_k + \nabla \bc_k\bz_k = \0$ and $[\bl]_i\leq [\bx_k+\bz_k]_i\leq [\bu]_i$ for all $i\in \mA^\star = \mA_\bell^\star\cup\mA_\bu^\star$. Furthermore, since $\bz_k\rightarrow \0$ as $\bx_k\rightarrow\tx$, we also have $[\bl]_i\leq [\bx_k+\bz_k]_i\leq [\bu]_i$ for~all $i\in(\mA^{\star})^-$. This implies that $\bz_k \in \Omega(\bx_{k};\theta_{k})$ with $\theta_{k}=1$ and we complete the proof.

\noindent$\bullet$ \textbf{Proof of (b).} Let us consider $\bx_k$ such that $\displaystyle\|\bx_k - \tx\|_{\max} = \max_{i\in[d]}|[\bx_k - \tx]_i|\leq 0.5\epsilon$. For~any $i\in \mA_\bell^\star$, we have $[\bx_{k} - \bell]_{i} \leq \|\bx_k-\tx\|_{\max}\leq 0.5\epsilon$, implying that $\mA_{\bell}^{\star} \subseteq \mA_{\bell,k}(\epsilon)$ by the definition of \eqref{equ:Akeps}. On the other hand, for any $i \in \mA_{\bell,k}(\epsilon)$, we have $[\tx-\bell]_i \leq |[\bx_{k} - \bx^{\star}]_{i}| + [\bx_{k} - \bell]_{i}\leq \|\bx_k-\tx\|_{\max}+[\bx_{k} - \bell]_{i}\leq 1.5\epsilon<[\tx-\bell]_j$, for any $j\in (\mA^{\star})^-$, where the last inequality is by the definition \eqref{equ:eps}. Thus, we know $i\in\mA^\star$. By the same derivations, we also have $[\tx-\bell]_i<[\bu-\bell]_i$, which implies $i\in\mA_\bell^\star$. Therefore, we have $\mA_{\bell,k}(\epsilon) = \mA^{\star}_{\bell}$ as long as $\|\bx_k - \tx\|_{\max} \leq 0.5\epsilon$. Similar arguments hold for showing that $\mA_{\bu,k}(\epsilon) = \mA^{\star}_{\bu}$, and we complete the proof.

\noindent$\bullet$ \textbf{Proof of (c) and (d).} We prove \textbf{(c)} and \textbf{(d)} jointly. We need the following supporting lemma, which is proved in Appendix \ref{app:D6}.

\begin{lemma}\label{lem:4}
Under Assumptions \ref{assump3} and \ref{assump5}, let $\beta_k = \iota_2(k+1)^{-b_2}$ satisfy $\iota_2>0, b_2\in(0.5,1]$. We have $\barg_k\rightarrow\nabla f^\star$ as $k\rightarrow\infty$ almost surely.
\end{lemma}

We will first show that the subproblem \eqref{equ:SQP:new} admits a solution $\barDelta\bx_k$ near the origin for $k$ large enough, with $\barDelta\bx_k\rightarrow 0$ as $k\rightarrow\infty$ and satisfying $\mA_{\bell}(\bx_k+\barDelta\bx_k) = \mA_\bell^\star$ and $\mA_{\bu}(\bx_k+\barDelta\bx_k) = \mA_\bu^\star$. We then show that the subproblem solution is also unique in a neighborhood of the origin. Let us consider the following linear system:
\begin{equation}\label{equ:kkt}
\begin{pmatrix}
\barB_k & \nabla \bc_k^\top & - \bI_{\mA_{\bell,k}(\epsilon)}^\top & \bI_{\mA_{\bu,k}(\epsilon)}^\top\\
\nabla \bc_k & \0 & \0 & \0 \\
- \bI_{\mA_{\bell,k}(\epsilon)} & \0 & \0 & \0 \\
\bI_{\mA_{\bu,k}(\epsilon)} & \0 & \0 & \0
\end{pmatrix}\begin{pmatrix}
\barDelta\bx_k\\
\barblambda_k^{\text{sub}}\\
[\barbmu_{1,k}^\text{sub}]_{\mA_{\bell,k}(\epsilon)}\\
[\barbmu_{2,k}^\text{sub}]_{\mA_{\bu,k}(\epsilon)}
\end{pmatrix} = -\begin{pmatrix}
\barg_k\\
\bc_k\\
[\bell-\bx_k]_{\mA_{\bell,k}(\epsilon)}\\
[\bx_k-\bu]_{\mA_{\bu,k}(\epsilon)}
\end{pmatrix}.
\end{equation}
By the construction of $\barB_k$, the local LICQ in Assumption \ref{assump5}, and the properties $\mA_{\bell,k}(\epsilon)=\mA_\bell^\star$ and $\mA_{\bu,k}(\epsilon)=\mA_\bu^\star$ proved in \textbf{(b)}, we know the above system has a (unique) solution. Let us rewrite the system as
\begin{multline}\label{eq:local_linear}
\begin{pmatrix}
\barB_k & \nabla \bc_k^\top & - \bI_{\mA_{\bell,k}(\epsilon)}^\top & \bI_{\mA_{\bu,k}(\epsilon)}^\top\\
\nabla \bc_k & \0 & \0 & \0 \\
- \bI_{\mA_{\bell,k}(\epsilon)} & \0 & \0 & \0 \\
\bI_{\mA_{\bu,k}(\epsilon)} & \0 & \0 & \0
\end{pmatrix}\begin{pmatrix}
\barDelta\bx_k\\
\barblambda_k^{\text{sub}}-\tlambda\\
[\barbmu_{1,k}^\text{sub}-\tmu_1]_{\mA_{\bell,k}(\epsilon)}\\
[\barbmu_{2,k}^\text{sub}-\tmu_2]_{\mA_{\bu,k}(\epsilon)}
\end{pmatrix} \\ = -\begin{pmatrix}
\barg_k+\nabla\bc_k^\top\tlambda - \bI_{\mA_{\bell,k}(\epsilon)}^\top[\tmu_1]_{\mA_{\bell,k}(\epsilon)} + \bI_{\mA_{\bu,k}(\epsilon)}^\top[\tmu_2]_{\mA_{\bu,k}(\epsilon)}\\
\bc_k\\
[\bell-\bx_k]_{\mA_{\bell,k}(\epsilon)}\\
[\bx_k-\bu]_{\mA_{\bu,k}(\epsilon)}
\end{pmatrix}.	
\end{multline}
By the convergence of $\bx_k$, $\mA_{\bell,k}(\epsilon)$, $\mA_{\bell,k}(\epsilon)$, and Lemma \ref{lem:4}, we know the right hand side of~\eqref{eq:local_linear} converges to 0 because it is the KKT conditions at $(\tx,\tlambda,\tmu)$, implying that $(\barDelta\bx_k; \barblambda_k^{\text{sub}}-\tlambda; 	[\barbmu_{1,k}^\text{sub}-\tmu_1]_{\mA_{\bell,k}(\epsilon)}; [\barbmu_{2,k}^\text{sub}-\tmu_2]_{\mA_{\bu,k}(\epsilon)})\rightarrow 0$. Now, let us show that $\mA_{\bell}(\bx_k + \barDelta\bx_k) = \mA_{\bell,k}(\epsilon)=\mA_\bell^\star$ and $\mA_{\bu}(\bx_k + \barDelta\bx_k) = \mA_{\bu,k}(\epsilon)=\mA_\bu^\star$ when $\|\barDelta\bx_k\|_{\max}\leq \epsilon$.~By the system \eqref{eq:local_linear}, we have $\mA_{\bell,k}(\epsilon)\subseteq \mA_\bell(\bx_k+\barDelta\bx_k)$. On the other hand, for any $i\in\mA_\bell(\bx_k+\barDelta\bx_k)$, we have $|[\bx_k]_i-[\bell]_i|\leq |[\barDelta\bx_k]_i|\leq \epsilon$, suggesting that $i\in\mA_{\bell,k}(\epsilon)$ and hence $\mA_{\bell}(\bx_k+\barDelta\bx_k)\subseteq\mA_{\bell,k}(\epsilon)$.~This~leads~to $\mA_{\bell,k}(\epsilon)= \mA_\bell(\bx_k+\barDelta\bx_k)$. Similarly, we can show $\mA_{\bu,k}(\epsilon)= \mA_\bu(\bx_k+\barDelta\bx_k)$. Next, we complement $[\barbmu_k^\text{sub}]_{\mA_k(\epsilon)} = ([\barbmu_{1,k}^\text{sub}]_{\mA_{\bell,k}(\epsilon)}, [\barbmu_{2,k}^\text{sub}]_{\mA_{\bu,k}(\epsilon)})$ by defining $[\barbmu_k^\text{sub}]_{\mA_k^-(\epsilon)}=\0$, and aim to show~that $(\barDelta\bx_k, \barblambda_k^{\text{sub}}, \barbmu_k^{\text{sub}})$ is the local solution of the SSQP subproblem \eqref{equ:SQP:new}. By \eqref{equ:kkt} and the complement $[\barbmu_k^\text{sub}]_{\mA_k^-(\epsilon)}=\0$, we have 
\begin{equation}\label{equ:sub1}
\begin{split}
& \barg_k + \barB_k \barDelta\bx_k + \nabla \bc_k^{\top} \barblambda_{k}^{\text{sub}} - \barbmu_{1,k}^{\text{sub}} + \barbmu_{2,k}^{\text{sub}} = \0,\\
& \barbmu_{1,k}^{\text{sub} \top}\left(\bell - \bx_k - \barDelta\bx_k\right) = \0,\\
& \barbmu_{2,k}^{\text{sub} \top}\left(\bx_k + \barDelta\bx_k-\bu\right) = \0.
\end{split}
\end{equation}
Furthermore, since $[\bmu^{\star}]_{\mA_{k}(\epsilon)} > 0$ by strict complementarity condition and $[\barbmu_{k}^{\text{sub}} - \bmu^{\star}]_{\mA_{k}(\epsilon)} \to 0$ as $k\rightarrow\infty$, we have $[\barbmu_{k}^{\text{sub}}]_{\mA_{k}(\epsilon)} > 0$ for $k$ large enough. This verifies the KKT conditions of the subproblem \eqref{equ:SQP:new}. The conditions of LICQ, SOSC, and strict complementarity of the subproblem are trivial due to the facts that $\mA_{\bell}(\bx_k + \barDelta\bx_k) = \mA_{\bell,k}(\epsilon)$ and $\mA_{\bell}(\bx_k + \barDelta\bx_k) = \mA_{\bell,k}(\epsilon)$.

Finally, we show the uniqueness of the solution $\barDelta\bx_k$ in the neighborhood of $\0$. Suppose there exists another solution $\barDelta\bx_{k}^{\prime}$ with $\|\barDelta\bx_k'\|_{\max}\leq \epsilon$. We know for any $i\in \mA_{\bell}(\bx_k + \barDelta\bx_k^{\prime})$, $[\bx_k-\bell]_i\leq \|\barDelta\bx_k'\|_{\max}\leq \epsilon$, suggesting that $i\in\mA_{\bell,k}(\epsilon)$ and hence $\mA_{\bell}(\bx_k + \barDelta\bx_k^{\prime})\subseteq \mA_{\bell,k}(\epsilon)$. Similarly, we can show $\mA_{\bu}(\bx_k + \barDelta\bx_k^{\prime})\subseteq \mA_{\bu,k}(\epsilon)$. This suggests that the LICQ of the subproblem holds at $\barDelta\bx_k'$ when $\bx_k$ is close to $\tx$, as $\bJ_k = (\nabla \bc_k; - \bI_{\mA_{\bell,k}(\epsilon)}; \bI_{\mA_{\bu,k}(\epsilon)})$ has full row-rank when $\bx_k$ is close to $\tx$. Then, the KKT condition of the subproblem at $\barDelta\bx_k'$ implies the existence of~unique dual multipliers $(\barblambda_{k}^{\text{sub} \prime}, \barbmu_{k}^{\text{sub} \prime})$ such that
\begin{equation*}
\barg_k + \barB_k \barDelta\bx_k^{\prime} + \nabla \bc_k^{\top} \barblambda_{k}^{\text{sub} \prime} - \bI_{\mA_{\bell}(\bx_k + \barDelta\bx_k^{\prime})}^{\top}[\barbmu_{1,k}^{\text{sub} \prime}]_{\mA_{\bell}(\bx_k + \barDelta\bx_k^{\prime})} + \bI_{\mA_{\bu}(\bx_k + \barDelta\bx_k^{\prime})}^{\top}[\barbmu_{2,k}^{\text{sub}\prime}]_{\mA_{\bu}(\bx_k + \barDelta\bx_k^{\prime})} = \0.
\end{equation*}
Comparing the above display with \eqref{equ:sub1}, we obtain
\begin{equation*}
\bJ_k^\top \left[\barblambda_{k}^{\text{sub}} - \barblambda_{k}^{\text{sub} \prime};[\barbmu_{1,k}^{\text{sub}} - \barbmu_{1,k}^{\text{sub} \prime}]_{\mA_{\bell,k}(\epsilon)}; [\barbmu_{2,k}^{\text{sub}} - \barbmu_{2,k}^{\text{sub} \prime}]_{\mA_{\bu,k}(\epsilon)} \right] = \barB_{k} (\barDelta\bx_k^{\prime} - \barDelta\bx_k).
\end{equation*}
By the full row-rankness of $\bJ_k$ as implied by LICQ at $\tx$, we know $\|[\barbmu_k^{\text{sub}} - \barbmu_k^{\text{sub}\prime}]_{\mA_{k}(\epsilon)}\|\leq \texttt{const}\cdot(\|\barDelta\bx_k'\|+\|\barDelta\bx_k\|)$ for some $\texttt{const}>0$. Thus, using the fact that $[\barbmu_{k}^{\text{sub}}]_{\mA_{k}(\epsilon)} > 0$, we can choose the neighborhood of the origin small enough such that if $\barDelta\bx_k', \barDelta\bx_k$ are in the~neighborhood, we have $[\barbmu_{k}^{\text{sub}\prime}]_{\mA_{k}(\epsilon)}>0$ as well, further implying that $\mA_{\bell,k}(\epsilon)\subseteq\mA_{\bell}(\bx_k+\barDelta\bx_k')$ and $\mA_{\bu,k}(\epsilon)\subseteq\mA_{\bu}(\bx_k+\barDelta\bx_k')$ due to the complementarity conditions of the subproblem. This leads to $\mA_{\bell}(\bx_k + \barDelta\bx_k^{\prime}) =\mA_{\bell,k}(\epsilon)$ and $\mA_{\bell}(\bx_k + \barDelta\bx_k^{\prime}) =\mA_{\bu,k}(\epsilon)$. However, by the KKT condition of the SSQP subproblem, the solution $\barDelta\bx_k^{\prime}$ must also follow the linear system  \eqref{equ:kkt}, which admits a unique solution. Thus, the solution of SSQP subproblem is unique near $\0$. The argument about~$\Delta\bx_{k}$ follows the same analysis by noting that $\nabla f_{k} \to \nabla f^{\star}$ almost surely.
This completes the proof of \textbf{(c)} and \textbf{(d)}.

\subsection{Proof of Lemma \ref{lemma_almost_eq}}\label{app:c7}

We first prove the convergence of $(\bx_k,\blambda_k,\bmu_k)$. By the proof of Lemma \ref{lemma_almost_convg}(c) in Appendix~\ref{app:C6}, we know $\barblambda_k^{\text{sub}}-\tlambda\rightarrow\infty$ and $[\barbmu_{k}^{\text{sub}} - \tmu]_{\mA_{k}(\epsilon)} \to \0$ as $k\rightarrow\infty$ almost surely. Let $\epsilon$ satisfy \eqref{equ:eps} and we have $\mA_{\bell,k}(\epsilon) = \mA^\star$ for $k$ large enough by Lemma \ref{lemma_almost_convg}(b). Since $[\barbmu_{k}^{\text{sub}}]_{\mA_k^-(\epsilon)} = \0 = \tmu_{(\mA^\star)^{-}}$, we have $\barbmu_{k}^{\text{sub}} - \tmu\rightarrow \0$ as $k\rightarrow\infty$ almost surely. With this result, we now prove $(\blambda_k, \bmu_k)\rightarrow(\tlambda, \tmu)$. We take $\blambda_k$ as an example. By the update \eqref{equ:update}, we have
\begin{multline*}
\blambda_{k+1}-\tlambda = \blambda_{k}-\tlambda + \baralpha_{k} \left(\barblambda_{k}^{\text{sub}} - \blambda_{k}\right) = (1-\baralpha_{k}) (\blambda_{k}-\tlambda) + \baralpha_{k}(\barblambda_{k}^{\text{sub}}-\tlambda) \\ = \prod_{i=0}^{k} (1-\baralpha_i) (\blambda_{0}-\tlambda) + \sum_{i=0}^{k} \prod_{j=i+1}^{k} (1-\baralpha_j) \baralpha_{i} (\barblambda_{i}^{\text{sub}}-\tlambda).
\end{multline*}

To proceed, we present the following lemma, which is proved in Appendix \ref{app:d7}.

\begin{lemma}\label{lem:5}
Suppose $\beta_k\geq 0$ satisfies $\beta_k\rightarrow0$ and $\sum_{k=0}^{\infty}\beta_k = \infty$, then for any scalar $a$ and sequence $e_k\rightarrow 0$, we have $\sum_{i=0}^{k}\prod_{j=i+1}^{k}(1-\beta_j)\beta_ie_i + a\prod_{j=0}^{k}(1-\beta_j) \rightarrow 0$ as $k\rightarrow\infty$.
\end{lemma}

By Lemma \ref{lem:5} and the fact that $\sum_{k=0}^{\infty}\baralpha_k=\infty$, we know $\blambda_k\rightarrow\tlambda$ almost surely. Following the same derivations, we obtain $\bmu_k\rightarrow\tmu$. Now, we consider the convergence of $\barB_k$. By the construction of $\bDelta_k$, the facts that $\mA_{\bell,k}(\epsilon)=\mA_\bell^\star$ and $\mA_{\bu,k}(\epsilon)=\mA_\bu^\star$, and the conditions of~LICQ and SOSC at $(\tx,\tlambda,\tmu)$, it suffices to show $\barQ_k\rightarrow \nabla^2 f^\star$ almost surely. This is an analogy of Lemma \ref{lem:4} and we summarize it to the following lemma. See Appendix \ref{app:D8} for the proof.$\quad\quad$

\begin{lemma}\label{lem:6}
Under Assumptions \ref{assump5} and \ref{assump7}, let $\gamma_k = \iota_3(k+1)^{-b_3}$ satisfy $\iota_3>0, b_3\in(0.5,1]$. We have $\barQ_k\rightarrow\nabla^2 f^\star$ as $k\rightarrow\infty$ almost surely.
\end{lemma}

The above lemma directly implies $\barB_k\rightarrow\nabla_{\bx}^2\mL^\star$ and we complete the proof.

\subsection{Proof of Theorem \ref{thm:5}}\label{app:c8}

We introduce some notation first. We let
\begin{align*}
\scriptsize \bH_k = \begin{pmatrix}
\barB_k & \tilde{\bJ}_{k}^\top\\
\tilde{\bJ}_{k} & \0
\end{pmatrix} = \begin{pmatrix}
\barB_k & \nabla \bc_k^\top & - \bI_{\mA_\bell^\star}^\top & \bI_{\mA_\bu^\star}^\top\\
\nabla \bc_k & \0 & \0 & \0 \\
- \bI_{\mA_\bell^\star} & \0 & \0 & \0 \\
\bI_{\mA_\bu^\star} & \0 & \0 & \0
\end{pmatrix},\quad 
 \bnabla_{\bw}\mL_k = \begin{pmatrix}
\barg_k +\nabla \bc_k^\top\blambda_k - \bI_{\mA_\bell^\star}^\top[\bmu_{1,k}]_{\mA_\bell^\star} + \bI_{\mA_\bu^\star}^\top[\bmu_{2,k}]_{\mA_\bu^\star}\\
\bc_k\\
[\bell-\bx_k]_{\mA_\bell^\star}\\
[\bx_k-\bu]_{\mA_{\bu}^\star}
\end{pmatrix},
\end{align*}
and $\nabla_{\bw}\mL_k$ is defined in the same way as $\bnabla_{\bw}\mL_k$ except that $\barg_k$ is replaced by the true gradient $\nabla f_k$. We further define the step direction
\begin{equation}\label{equ:local:step}
\barDelta\bw_k = \begin{pmatrix}
\barDelta\bx_{k}\\
\barDelta\blambda_{k}\\
[\barDelta\bmu_{1,k}]_{\mA_\bell^\star}\\
[\barDelta\bmu_{2,k}]_{\mA_\bu^\star}\\
\end{pmatrix} = - \bH_k^{-1}\bnabla_{\bw}\mL_k\quad\quad \text{ and }\quad\quad [\barDelta\bmu_{k}]_{(\mA^\star)^-} =  - [\bmu_{k}]_{(\mA^\star)^-}.
\end{equation}
By the KKT conditions of the SSQP subproblem \eqref{equ:kkt}, the facts from Lemma \ref{lemma_almost_convg} that $\mA_\bell(\bx_k+\barDelta\bx_k) = \mA_{\bell,k}(\epsilon)=\mA_\bell^\star$ and $\mA_\bu(\bx_k+\barDelta\bx_k) = \mA_{\bu,k}(\epsilon)=\mA_\bu^\star$, and using the relation of $(\barblambda_{k}^\text{sub}, \barbmu_{k}^\text{sub})\\ = (\blambda_k + \barDelta\blambda_k, \bmu_k+\barDelta\bmu_k)$, we know the update of the iterates when $k$ is large enough can~be~written by utilizing \eqref{equ:local:step} as
\begin{equation}\label{equ:local:update}
\bw_{k+1} = \bw_k + \baralpha_k \barDelta\bw_k\quad\quad\text{ and }\quad\quad [\bmu_{k+1}]_{(\mathcal{A}^{\star})^-} = (1-\baralpha_k)\cdot[\bmu_{k}]_{(\mathcal{A}^{\star})^-}.
\end{equation}
Let $\Upsilon>0$ be a constant and we propose the following conditions:
\begin{enumerate}[label=(\alph*),topsep=0pt]
\setlength\itemsep{0.0em}
\item $\max\{\beta_k, \baralpha_k\}\leq 1$; $\|\bm{H}_k^{-1} - \left(\bm{H}^{\star}\right)^{-1}\| \leq \Upsilon \| \bm{H}_k - \bm{H}^{\star}\|$; $\|\bm{H}_k^{-1}\| \leq \Upsilon$.
\item $\max\{\|\nabla f_k-\nabla f^\star\|, \|\nabla^2 f_k - \nabla^2 f^\star\|\}\leq \Upsilon\|\bx_k-\tx\|$; $\|\nabla f_k-\nabla f_{k-1}\|\leq \Upsilon\|\bx_k-\bx_{k-1}\|$.
\item $\max\{\|\bw_k - \tw\|, \|\nabla f_k-\nabla f^\star\|\}\leq \frac{1}{\Upsilon}$; $\max\{\|\nabla_\bw\mL_k\|, \|\nabla_{\bw}^2\mL_k - \nabla_{\bw}^2\mL^\star\|\}\leq \Upsilon\|\bw_k-\tw\|$.
\item $\|\nabla_\bw\mL_k - \tbH(\bw_k-\tw)\|\leq \Upsilon\|\bw_k-\tw\|^2$; $\|\bw_k-\tw - \bH_k^{-1}\nabla_\bw\mL_k\|\leq \|\bw_k-\tw\|/\Upsilon$.
\item $\mA_\bell(\bx_k+\barDelta\bx_k) = \mA_{\bell,k}(\epsilon)=\mA_\bell^\star$; $\mA_\bu(\bx_k+\barDelta\bx_k) = \mA_{\bu,k}(\epsilon)=\mA_\bu^\star$.
\item The subproblem solution $(\barDelta\bx_{k}, \barblambda_{k}^\text{sub}, \barbmu_{k}^\text{sub}) = (\barDelta\bx_k, \blambda_k + \barDelta\blambda_k, \bmu_k+\barDelta\bmu_k)$ is from \eqref{equ:local:step}.
\end{enumerate}
For any $k_0\geq0$, we also define the stopping time
\begin{equation}\label{equ:tau}
\tau_{k_0}:=\inf_{j}\{j \geq k_0: \text{any of the above conditions does not hold at}~j\text{-th iteration}\}.
\end{equation}
By Lemmas \ref{lemma_almost_convg}, \ref{lemma_almost_eq}, Assumption \ref{assump5}, and choosing $\Upsilon$ to be large enough, we know for any~run of the method, there exists a (potentially random) $\tilde{k}_0<\infty$ such that $\tau_{k_0}=\infty$, $\forall k_0\geq\tilde{k}_0$.

With the above preparation definitions, we first focus on the inactive dual components. For any run of the method, we know from \eqref{equ:local:update} that there exists a (potentially random) $\tilde{k}_0<\infty$ such that for all $k\geq \tilde{k}_0$,
\begin{align*}
\|[\bmu_{k}]_{(\mathcal{A}^{\star})^-}\| & = \prod_{j=\tilde{k_0}}^{k-1}(1-\baralpha_j)\cdot\|[\bmu_{\tilde{k}_0}]_{(\mathcal{A}^{\star})^-}\| \leq \prod_{j=0}^{\tk_0-1}(1-\alpha_j)^{-1}\|[\bmu_{\tilde{k}_0}]_{(\mathcal{A}^{\star})^-}\|\cdot\prod_{j=0}^{k-1}(1-\alpha_j)\\
& \leq \prod_{j=0}^{\tk_0-1}(1-\alpha_j)^{-1}\|[\bmu_{\tilde{k}_0}]_{(\mathcal{A}^{\star})^-}\|\exp\rbr{-\sum_{j=0}^{k-1}\alpha_j}.
\end{align*}
Note that
\begin{equation}\label{pf:int}
\sum_{i=0}^{k-1} \alpha_{k} \geq \iota_1\int_{1}^{k+1} x^{-b_1} dx = \begin{cases*}
\frac{\iota_1}{1-b_1}\rbr{(k+1)^{1-b_1} - 1} & \text{ if } $b_1<1$,\\
\iota_1\log(k+1) & \text{ if } $b_1=1$.
\end{cases*}
\end{equation}
Combining the above two displays, we know
\begin{equation*}
\|[\bmu_{k}]_{(\mathcal{A}^{\star})^-}\| = \begin{cases*}
o(k^{-b}) \text{ for any $b>0$} & \text{ if } $b_1<1$,\\
\mathcal{O}(k^{-\iota_1}) & \text{ if } $b_1=1$.
\end{cases*}
\end{equation*}

We now focus on the primal and active dual components. For notational simplicity, we~denote $\barDelta\bw_k = (\barDelta\bx_k, \barblambda_k^{\text{sub}}-\blambda_k, [\barbmu_{k}^\text{sub} - \bmu_k]_{\mA^\star})$ for any $k\geq 0$, while noting from \eqref{equ:local:step} that we only have $\barDelta\bw_k=- \bH_k^{-1}\bnabla_{\bw}\mL_k$ for $k_0\leq k< \tau_{k_0}$ with any given $k_0 \geq 0$. For any $k\geq 0$,~we~have $\quad$
\begin{align*}
& \bw_{k+1}-\tw \\
& = \bw_k-\tw +\baralpha_k \barDelta\bw_k = \bw_k-\tw +\alpha_k \barDelta\bw_k + (\baralpha_k-\alpha_k)\barDelta\bw_k\\
& = \bw_k-\tw -\alpha_k(\tbH)^{-1}\nabla_\bw\mL_k + \alpha_k(\tbH)^{-1}(\nabla_\bw\mL_k - \bnabla_{\bw}\mL_k) + \alpha_k((\tbH)^{-1}-\bH_k^\dagger)\bnabla_{\bw}\mL_k \\
&\quad + \alpha_k(\bH_k^\dagger\bnabla_{\bw}\mL_k + \barDelta\bw_k) + (\baralpha_k-\alpha_k)\barDelta\bw_k\\
& = \bw_k-\tw - \alpha_k\bH_k^\dagger\nabla_\bw\mL_k + \alpha_k(\tbH)^{-1}(\nabla_\bw\mL_k - \bnabla_{\bw}\mL_k) + \alpha_k((\tbH)^{-1}-\bH_k^\dagger)(\bnabla_{\bw}\mL_k-\nabla_\bw\mL_k) \\
&\quad + \alpha_k(\bH_k^\dagger\bnabla_{\bw}\mL_k + \barDelta\bw_k) + (\baralpha_k-\alpha_k)\barDelta\bw_k\\
& = (1-\alpha_k)(\bw_k-\tw) - \alpha_k\bH_k^\dagger(\nabla_\bw\mL_k - \tbH(\bw_k-\tw)) - \alpha_k(\bH_k^\dagger\tbH - \bI)(\bw_k-\tw)\\
& \quad - \alpha_k(\bH_k^\dagger-(\tbH)^{-1})(\bnabla_{\bw}\mL_k-\nabla_\bw\mL_k) + \alpha_k(\bH_k^\dagger\bnabla_{\bw}\mL_k + \barDelta\bw_k) + (\baralpha_k-\alpha_k)\barDelta\bw_k\\
& \quad - \alpha_k(\tbH)^{-1}(\bnabla_{\bw}\mL_k-\nabla_\bw\mL_k).
\end{align*}
Applying the above recursion from $0$ to $k$, we have
\begin{align}\label{pf:equ:J}
& \bw_{k+1}-\tw - \prod_{i=0}^{k}(1-\alpha_i)(\bw_0-\tw) \nonumber\\
& =  - \sum_{i=0}^{k}\prod_{j=i+1}^{k}(1-\alpha_j)\alpha_i\bigg\{\bH_i^\dagger(\nabla_\bw\mL_i - \tbH(\bw_i-\tw)) + (\bH_i^\dagger\tbH - \bI)(\bw_i-\tw)\bigg\} \nonumber\\
& \quad - \sum_{i=0}^{k}\prod_{j=i+1}^{k}(1-\alpha_j)\alpha_i\cbr{(\bH_i^\dagger-(\tbH)^{-1})(\bnabla_{\bw}\mL_i-\nabla_\bw\mL_i) - (\bH_i^\dagger\bnabla_{\bw}\mL_i + \barDelta\bw_i) - \frac{\baralpha_i-\alpha_i}{\alpha_i}\barDelta\bw_i} \nonumber\\
& \quad - \sum_{i=0}^{k}\prod_{j=i+1}^{k}(1-\alpha_j)\alpha_i(\tbH)^{-1}(\bnabla_{\bw}\mL_i-\nabla_\bw\mL_i) \eqqcolon -\mJ_{1,k}-\mJ_{2,k}-\mJ_{3,k}.
\end{align}
First, for the term $\prod_{i=0}^{k}(1-\alpha_i)(\bw_0-\tw)$ on the left, we note from \eqref{pf:int} that $\prod_{i=0}^{k}(1-\alpha_i)(\bw_0-\tw) = o(\sqrt{\alpha_k})$ as long as $\iota_1>0.5$ when $b_1=1$. Next, we aim to show $\mJ_{1,k} = o_p(\sqrt{\alpha_k})$, $\mJ_{2,k} = o_p(\sqrt{\alpha_k})$, and asymptotic normality of $\mJ_{3,k}$. Throughout the proof, we use $o_p(\cdot)$ and $\mO_p(\cdot)$ to denote the standard small- and big-$O$ notation in the probability sense. We state~the~following preparation lemma.

\begin{lemma}[Lemma~C.3 in \cite{Na2025Derivative}]\label{lem:c8}
Consider a sequence of random variables $\{X_k\}_{k=0}^\infty$~and a sequence of events $\{\mA_k\}_{k=0}^\infty$. Let $\tau_{k_0} = \inf\{k\geq k_0: \mA_k \text{ happens}\}$ be the first index $k$ after $k_0$ such that $\mA_k$ happens. Suppose that for each realization of the sequence, there exists a (potentially random) $\tilde{k}_0<\infty$ such that $\tau_{\tilde{k}_0} = \infty$. Also, for the sequence $\alpha_k = \iota_1(k+1)^{-b_1}$ with $b_1\in(0,1]$ and $\iota_1>0.5\1_{b_1=1}$, suppose there exists a deterministic $\bar{k}_0>0$ such that for any fixed $k_0\geq \bar{k}_0$, $X_k\1_{\tau_{k_0}>k} = o_p(\sqrt{\alpha_k})$.~Then, we have 
\begin{equation*}
\sum_{i=0}^{k}\prod_{j=i+1}^{k}(1-\alpha_j)\alpha_iX_i = o_p(\sqrt{\alpha_k}).
\end{equation*}	
\end{lemma}

By the above lemma and \eqref{pf:equ:J}, we know $\mJ_{1,k} = o_p(\sqrt{\alpha_k})$ and $\mJ_{2,k} = o_p(\sqrt{\alpha_k})$ as long as
\begin{subequations}\label{pf:equ:todo}
\begin{align}
\|\bH_k^\dagger(\nabla_\bw\mL_k - \tbH(\bw_k-\tw))\| \1_{\tau_{k_0}>k} & = o_p(\sqrt{\alpha_k}), \label{pf:equ:todo1}\\
\|(\bH_k^\dagger\tbH - \bI)(\bw_k-\tw)\|\1_{\tau_{k_0}>k} & = o_p(\sqrt{\alpha_k}), \label{pf:equ:todo2}\\
\|(\bH_k^\dagger-(\tbH)^{-1})(\bnabla_{\bw}\mL_k-\nabla_\bw\mL_k)\|\1_{\tau_{k_0}>k} & = o_p(\sqrt{\alpha_k}), \label{pf:equ:todo3}\\
\|(\bH_k^\dagger\bnabla_{\bw}\mL_k + \barDelta\bw_k)\| \1_{\tau_{k_0}>k} & = o_p(\sqrt{\alpha_k}), \label{pf:equ:todo4}\\
\frac{\baralpha_k-\alpha_k}{\alpha_k}\|\barDelta\bw_k\|\1_{\tau_{k_0}>k} & = o_p(\sqrt{\alpha_k}). \label{pf:equ:todo5}
\end{align}
\end{subequations}
For \eqref{pf:equ:todo1}, we apply the definition of $\tau_{k_0}$ in \eqref{equ:tau} and have
\begin{equation*}
\|\bH_k^\dagger(\nabla_\bw\mL_k - \tbH(\bw_k-\tw))\| \1_{\tau_{k_0}>k} \leq \Upsilon^2\|\bw_k-\tw\|^2\1_{\tau_{k_0}>k}.
\end{equation*}
For \eqref{pf:equ:todo2}, we have
\begin{align*}
\|(\bH_k^\dagger\tbH - \bI)(\bw_k-\tw)\|\1_{\tau_{k_0}>k} & \; \stackrel{\mathclap{\eqref{equ:tau}}}{\leq}\; \Upsilon \|\bH_k-\tbH\|\|\bw_k-\tw\|\1_{\tau_{k_0}>k} \\
\; &\;  \leq 0.5\Upsilon \|\bH_k-\tbH\|^2\1_{\tau_{k_0}>k} +  0.5\Upsilon\|\bw_k-\tw\|^2\1_{\tau_{k_0}>k}.
\end{align*}
For \eqref{pf:equ:todo3}, we have
\begin{multline*}
\|(\bH_k^\dagger-(\tbH)^{-1})(\bnabla_{\bw}\mL_k-\nabla_\bw\mL_k)\|\1_{\tau_{k_0}>k} \;\stackrel{\mathclap{\eqref{equ:tau}}}{\leq}\; \Upsilon\|\bH_k-\tbH\| \|\barg_k - \nabla f_k\|\1_{\tau_{k_0}>k} \\
\leq 0.5\Upsilon\|\bH_k - \tbH\|^2\1_{\tau_{k_0}>k}  + 0.5\Upsilon\|\barg_k - \nabla f_k\|^2\1_{\tau_{k_0}>k}.
\end{multline*}
For \eqref{pf:equ:todo4}, it is trivial due to \eqref{equ:local:step}.  For \eqref{pf:equ:todo5}, we have
\begin{align*}
\frac{\baralpha_k-\alpha_k}{\alpha_k}\|\barDelta\bw_k\|\1_{\tau_{k_0}>k} & \; \leq\; \psi\alpha_k^{p-1}\|\barDelta\bw_k\|\1_{\tau_{k_0}>k} \stackrel{\mathclap{\eqref{equ:local:step}}}{\leq} \Upsilon\psi \alpha_k^{p-1}\|\bnabla_{\bw}\mL_k\|\1_{\tau_{k_0}>k} \\
& \; \stackrel{\mathclap{\eqref{equ:tau}}}{\leq}\; \Upsilon\psi \alpha_k^{p-1}\|\barg_k-\nabla f_k\|\1_{\tau_{k_0}>k} + \Upsilon^2\psi \alpha_k^{p-1}\|\bw_k-\tw\|\1_{\tau_{k_0}>k}.
\end{align*}
Combining the above four displays, we know \eqref{pf:equ:todo} is implied by showing that
\begin{equation}\label{equ:rate}
\begin{gathered}
\rbr{\|\bw_k-\tw\|^2 +  \|\barg_k-\nabla f_k\|^2 + \|\bH_k - \tbH\|^2 }\1_{\tau_{k_0}>k}  = o_p(\sqrt{\alpha_k}),\\
\alpha_k^{p-1}\rbr{\|\bw_k-\tw\| + \|\barg_k-\nabla f_k\|}\1_{\tau_{k_0}>k} = o_p(\sqrt{\alpha_k}).
\end{gathered}
\end{equation}
We present the following two lemmas to corroborate the above result \eqref{equ:rate}. See Appendices \ref{app:d9} and \ref{app:d11} for the proofs, respectively.

\begin{lemma}\label{lem:c9}
Under Assumption \ref{assump3} and suppose $\alpha_k = \iota_1 (k+1)^{-b_1}$, $\beta_k = \iota_2 (k+1)^{-b_2}$ satisfy $\iota_1,\iota_2>0$, $0<b_2<b_1\leq 1$, and $\iota_1>0.5$ if $b_1=1$. There exists a deterministic integer $\bar{k}_0>0$ such that for any $k_0\geq \bar{k}_0$, the following result holds for some deterministic constant $\Upsilon(k_0)>0$:
\begin{equation*}
\mE\big[(\|\bw_k-\tw\|^2 + \|\barg_k-\nabla f_k\|^2)\1_{\tau_{k_0}>k}\big] \leq \Upsilon(k_0)\beta_k,\quad \quad \forall k\geq k_0.
\end{equation*}
\end{lemma}

\begin{lemma}\label{lem:c10}
Under Assumptions \ref{assump3}, \ref{assump7} and suppose $\alpha_k = \iota_1 (k+1)^{-b_1}$, $\beta_k = \iota_2 (k+1)^{-b_2}$, $\gamma_k=\iota_3(k+1)^{-b_3}$ satisfy $\iota_1,\iota_2, \iota_3>0$, $b_1\in(0, 1]$, $b_2\in(0.5b_1, b_1)$, $b_3\in(0.5b_1, 1]$, and $\iota_1>0.5$ if $b_1=1$ and $\iota_3>0.25b_1$ if $b_3=1$. There exists a deterministic integer $\bar{k}_0>0$ such that~for~any $k_0\geq \bar{k}_0$, the following result holds for some deterministic constant $\Upsilon(k_0)>0$:
\begin{equation*}
\mE[\|\bH_k-\tbH\|^2 \1_{\tau_{k_0}>k}] / (\Upsilon(k_0)\sqrt{\alpha_k}) \rightarrow 0 \quad \text{ as }\quad k\rightarrow\infty.
\end{equation*}
\end{lemma}

With Lemmas \ref{lem:c9} and \ref{lem:c10}, and applying the conditions that $b_2>0.5\geq 0.5b_1$ and $p>1.5-0.5b_2/b_1$, we know \eqref{equ:rate} holds, leading to $\mJ_{1,k} = o_p(\sqrt{\alpha_k})$ and $\mJ_{2,k} = o_p(\sqrt{\alpha_k})$. Finally, we present the following lemma demonstrating the asymptotic normality property~of~$\mJ_{3,k}$. See Appendix \ref{app:d12} for the proof.

\begin{lemma}\label{lem:c11}
\hskip-0.15cm Under Assumptions \ref{assump5}, \ref{assump9} and suppose $\alpha_k = \iota_1 (k+1)^{-b_1}$, $\beta_k = \iota_2 (k+1)^{-b_2}$~satisfy $\iota_1,\iota_2>0$, $b_1\in((2-2\delta)/(2+\delta), 1]$, $b_2\in(0, b_1)$, and $\iota_1>2/3$ if $b_1=1$.~Then,~we~have $1/\sqrt{\alpha_k}\cdot\mJ_{3,k}\stackrel{d}{\longrightarrow} \mathcal{N} \left( \bm{0}, \eta \cdot \bm{\Omega}^{\star} \right)$, where $\eta$ and $\tOmega$ are defined in Theorem \ref{thm:5}.
\end{lemma}

Noting that $\alpha_k/\baralpha_k\rightarrow 1$ almost surely since $p>1.5-0.5b_2/b_1$ and $b_1>b_2>0$ imply~$p>1$, we combine Lemma \ref{lem:c11} with \eqref{pf:equ:J} and complete the proof.

\subsection{Proof of Theorem \ref{thm:6}}\label{app:c9}

By Assumption \ref{assump5} and Lemma \ref{lemma_almost_eq}, it suffices to show the convergence of $\text{Cov}(\{\nabla F(\bx_i;\zeta_i)\}_{i=0}^k)$. In particular, we have
\begin{align}\label{equ:c9:1}
\big\|\text{Cov}& (\{\nabla F(\bx_i; \zeta_i)\}_{i=0}^k)  - \text{Cov}(\nabla F(\tx;\zeta))\big\| \nonumber\\
& \leq \nbr{\frac{1}{k+1}\sum_{i=0}^{k} \nabla F(\bx_{i};\zeta_{i}) \nabla F(\bx_{i};\zeta_{i})^{\top} - \mathbb{E} \left[ \nabla F(\bm{x}^{\star};\zeta)   \nabla F(\bm{x}^{\star};\zeta)^{\top} \right]} \nonumber\\
& \quad + \nbr{\left( \frac{1}{k+1}\sum_{i=0}^{k} \nabla F(\bx_{i};\zeta_{i}) \right)\left( \frac{1}{k+1}\sum_{i=0}^{k} \nabla F(\bx_{i};\zeta_{i}) \right)^{\top} - \nabla f^{\star}\nabla f^{\star \top}}.
\end{align}
We analyze the two terms on the right hand side separately. For the first term in \eqref{equ:c9:1}, we~have
\begin{align*}
\bigg\|\frac{1}{k+1}\sum_{i=0}^{k} & \nabla F(\bx_{i};\zeta_{i})   \nabla F(\bx_{i};\zeta_{i})^{\top} - \mathbb{E} \left[ \nabla F(\bm{x}^{\star};\zeta)   \nabla F(\bm{x}^{\star};\zeta)^{\top} \right]\bigg\|\\
& \leq \left\|  \frac{1}{k+1}\sum_{i=0}^{k} \nabla F(\bx_{i};\zeta_{i})  \nabla F(\bx_{i};\zeta_{i})^{\top} - \mathbb{E}_i\left[ \nabla F(\bm{x}_{i};\zeta_i)   \nabla F(\bm{x}_{i};\zeta_i)^{\top}\right]\right\| \\
& \quad + \left\|  \frac{1}{k+1}\sum_{i=0}^{k}   \mathbb{E}_i\left[ \nabla F(\bm{x}_{i};\zeta_i)   \nabla F(\bm{x}_{i};\zeta_i)^{\top}\right] -\mE\left[ \nabla F(\bm{x}^{\star};\zeta)   \nabla F(\bm{x}^{\star};\zeta)^{\top} \right]\right\|.
\end{align*}
The first term converges to zero almost surely by bounded 4-th moment condition~of~$\nabla F(\bx_k;\zeta_k)$ and the strong law of large numbers for square integrable martingale \citep[Theorem 1.3.15]{Duflo1997Random}; the second term converges to zero almost surely by Assumptions \ref{assump5}, \ref{assump9}, and the fact that $x_k\rightarrow 0\Rightarrow\frac{1}{k+1}\sum_{i=0}^{k}x_i\rightarrow 0$. Similarly, for the second term in \eqref{equ:c9:1}, we just note that
\begin{equation*}
\left\|\frac{1}{k+1}\sum_{i=0}^{k} \nabla F(\bm{x}_{i};\zeta_{i})  - \nabla f^{\star} \right\| \leq   \left\|\frac{1}{k+1}\sum_{i=0}^{k} \left( \nabla F(\bm{x}_{i};\zeta_{i})  - \nabla f_{i} \right) \right\| +  \left\|\frac{1}{k+1}\sum_{i=0}^{k}  \left(\nabla f_{i}  - \nabla f^{\star}  \right) \right\|,
\end{equation*}
and it converges to zero almost surely due to the same reasons as above. This completes the~proof.

\section{Proofs of Supporting Lemmas}

\subsection{Proof of Lemma \ref{lemma_A_u}}\label{Appendix:pf:D1}

We adapt the proof from~\citep[Lemma B.1]{Davis2024Asymptotic}. Consider the normalizing constant $C(\bv) = 1 + \int h(\bv^{\top} \bg(\zeta)) d \mathcal{P}(\zeta)$. Since $h$ is $C^3$-smooth with bounded derivatives and $\bg \in \mathcal{G}$ satisfies $\mathbb{E}\|\bg(\zeta)\|^2 < \infty$, the dominated convergence theorem~\mbox{ensures}~$C(\bv)$ is twice~continuously differentiable with $\nabla C(\bv) = \int h^{\prime}(\bv^{\top}\bg(\zeta))\bg(\zeta) d \mathcal{P}(\zeta)$, $\nabla^2 C(\bv) = \int h^{\prime \prime}(\bv^{\top} \bg(\zeta))  \bg(\zeta) \bg(\zeta)^{\top} d \mathcal{P}(\zeta)$. Similarly, $\nabla_{\bw}\mL(\bw)$ is $C^1$-smooth with~$\nabla_{\bw}^2\mL(\bw) = \mathbb{E}_{\zeta \sim \mathcal{P}} \left[\nabla_{\bw}^2\mL(\bw;\zeta)\right]$. Define $\tilde{\nabla}_{\bw}\mL_\bv(\bw) = \int h(\bv^{\top} \bg(\zeta)) \nabla_{\bw}\mL(\bw;\zeta) d \mathcal{P}(\zeta)$. By the dominated convergence theorem, $\tilde{\nabla}_{\bw}\mL_\bv(\bw)$ is continuously differentiable in both $\bv$ and $\bw$, and its partial derivatives are given by
\begin{equation}\label{pf:D1}
\begin{split}
\nabla_{\bv}(\tilde{\nabla}_{\bw}\mL_\bv(\bw)) & = 
\int h^{\prime}(\bv^{\top} \bg(\zeta)) \nabla_{\bw}\mL(\bw;\zeta) \bg(\zeta)^{\top} d \mathcal{P}(\zeta), \\
\nabla_{\bw}(\tilde{\nabla}_{\bw}\mL_\bv(\bw)) & = \int h(\bv^{\top} \bg(\zeta)) \nabla_{\bw}^2\mL(\bw;\zeta) d \mathcal{P}(\zeta),
\end{split}
\end{equation}
both of which are continuous in $(\bw, \bv)$. Thus, in a small neighborhood of $(\bw, \bv) = (\tw, \0)$,~we apply $h(t)=t$ and $\mE_{\zeta\sim\P}[\bg(\zeta)]=\0$, and know $\nabla_{\bw}\mL_{\bv}(\bw) = \tilde{\nabla}_{\bw}\mL_\bv(\bw)/C(\bv)$ is $C^1$-smooth.~Finally, the first identity in \eqref{pf:B4} directly follows from the fact that $\nabla_{\bw}\mL_{\bv}(\bw)|_{\bv = \0} = \nabla_{\bw}\mL(\bw)$ for any $\bw$. For the second identity, we note that $\nabla C(\0) = \0$ and $\nabla^2 C(\0) = \0$, implying $C(\bv) = 1 + o(\|\bv\|^2)$. Thus, $\nabla_{\bv\bw}\mL_{\bv}(\bw)|_{\bv = \0} = \nabla_{\bv}(\tilde{\nabla}_{\bw}\mL_\bv(\bw))|_{\bv = \0} \stackrel{\eqref{pf:D1}}{=}\mE_{\zeta\sim \P}[\nabla_{\bw}\mL(\bw; \zeta) \bg(\zeta)^{\top}]$. This completes the proof.

\subsection{Proof of Lemma \ref{lemma_solution_map}}\label{Appendix:pf:D2}

We adapt the proof from \citep[Lemma B.2]{Davis2024Asymptotic}. Define the linearized mapping $\Psi(\bw) = \nabla_{\bw}\mL(\tw) + \nabla_{\bw}^2\mL(\tw) \left(\bw - \tw\right) = \bm{H}^{\star}\left(\bw - \tw\right)$, which is invertible and is $C^{1}$-smooth. By Lemma \ref{lemma_A_u}, we can equivalently express $\Psi(\bw) = \nabla_{\bw}\mL_{\bv}(\tw)|_{\bv = \0} + \nabla_{\bw}^2\mL_{\bv}(\tw)|_{\bv = \0} \left(\bw - \tw\right)$. 
By the implicit function theorem \citep[Theorem~2D.6]{Dontchev2009Implicit},
the mapping $S(\bv)$ admits a single-valued localization $s(\bv)$ around $\tw$ and $\bv = \0$ with $\nabla s(\0) = - \nabla \sigma(\0) \cdot \nabla_{\bv\bw}\mL_{\bv}(\tw)|_{\bv = \0} = - \left(\bm{H}^{\star} \right)^{-1}  \cdot \mathbb{E}_{\zeta \sim \P} \left[\nabla_{\bw}\mL(\tw; \zeta) \bg(\zeta)^{\top}\right]$. This completes the proof.

\subsection{Proof of Lemma \ref{lem:3}}\label{app:d3}

We present two preparation lemmas that are proved in Appendices \ref{app:pf:D4} and \ref{app:d5}, respectively.

\begin{lemma}\label{lemma20}
Let $\alpha_k = \iota_1 (k+1)^{-b_1}$, $\beta_k = \iota_2 (k+1)^{-b_2}$ for $\iota_1, \iota_2, b_1, b_2 >0$ with $b_2<\min\{b_1,1\}$. Under Assumption \ref{assump3} and the Lipschitz continuity of $\nabla f$ (cf. Assumption \ref{assump2}), we have
\begin{equation*}
\mathbb{E} [ \| \bar{\bm{g}}_k - \nabla f_{k}\|^2 ] = \mathcal{O}\left(\beta_{k} + \alpha_k^2 / \beta_k^2\right). 
\end{equation*}
\end{lemma}

\begin{lemma}\label{lemma19}
Suppose $\bB_k\succeq \kappa_1\bI$, then  
\begin{equation*}
\left\| \barDelta\bx_k - \Delta\bx_k \right\| \leq \kappa_1^{-1} \left\| \Bar{\bm{g}}_{k} - \nabla f_{k}  \right\|,
\end{equation*}
where $\barDelta\bx_k$ is the solution of \eqref{equ:SQP:new} and $\Delta\bx_k$ is the solution with $\barg_k$ replaced by $\nabla f_k$.
\end{lemma}

Now, we apply the results in Lemmas \ref{lemma20} and \ref{lemma19} and obtain
\begin{equation*}
\mathbb{E} [\| \barDelta\bx_k - \Delta\bx_k \| ] \leq \kappa_1^{-1} \mathbb{E}[\|\Bar{\bm{g}}_{k} - \nabla f_{k} \|] \leq \kappa_1^{-1} \cbr{\mathbb{E}[\|\Bar{\bm{g}}_{k} - \nabla f_{k} \|^2]}^{1/2} = \mathcal{O}\left(\sqrt{\beta_k} + \alpha_k / \beta_k \right).
\end{equation*}
By the setup of $\alpha_k, \beta_k$, it follows that $\sum_{k=0}^{\infty} \alpha_k \mathbb{E} [\|\barDelta\bx_k  - \Delta\bx_k \|] < \infty$ if $b_1 + 0.5b_2 >1$ and $2b_1 - b_2 >1$. Equivalently, this requires $b_1 \in (0.75,1]$ and $b_2 \in \left( 2-2b_1,2b_1-1\right)$. We complete~the proof.

\subsection{Proof of Lemma \ref{lemma20}}\label{app:pf:D4}

From the update rule of $\Bar{\bm{g}}_{k}$ in \eqref{equ:gradHess:average}, we have
\begin{align}\label{equ:gk}
& \bar{\bm{g}}_{k} - \nabla f_{k} = \beta_k (\nabla F(\bx_k;\zeta_k)- \nabla f_{k}) + (1-\beta_k) \left( \Bar{\bm{g}}_{k-1} - \nabla f_{k-1} \right) + (1-\beta_k) \left( \nabla f_{k-1} - \nabla f_{k} \right) \nonumber\\
& = \beta_k (\nabla F(\bx_k;\zeta_k) - \nabla f_{k}) + (1-\beta_k)  \left\{ \beta_{k-1} ( \nabla F(\bx_{k-1};\zeta_{k-1}) - \nabla f_{k-1})\right. \nonumber\\
& \quad \left. + (1-\beta_{k-1}) \left( \Bar{\bm{g}}_{k-2} - \nabla f_{k-2} \right) + (1-\beta_{k-1}) \left( \nabla f_{k-2} - \nabla f_{k-1} \right) \right\} + (1-\beta_k) \left( \nabla f_{k-1} - \nabla f_{k} \right) \nonumber\\
& = \sum_{i=0}^{k} \left( \prod_{j=i+1}^{k} (1-\beta_j) \right) \beta_i \left( \nabla F(\bx_i;\zeta_i) - \nabla f_i \right) + \sum_{i=0}^{k} \left( \prod_{j=i}^{k} (1-\beta_j) \right) \left( \nabla f_{i-1} - \nabla f_{i} \right) \nonumber\\
& \coloneqq \mathcal{W}_{1,k} + \mathcal{W}_{2,k},
\end{align}
where we denote $\nabla f_{-1}=\barg_{-1}$ in the second last equality. By Lemma \ref{lemma_tool} and noting that $\alpha_k \leq \Bar{\alpha}_{k} \leq \alpha_k + \psi \alpha_k^p $, we have
\begin{equation*}
\|\mathcal{W}_{2,k}\| \leq\prod_{j=0}^{k}(1-\beta_j)(\|\barg_{-1}\|+\|\nabla f_0\|) +  \sum_{i=1}^{k} \left( \prod_{j=i}^{k} (1-\beta_j) \right) \Bar{\alpha}_{i-1} M_{\bm{\ell},\bm{u}} = \mathcal{O}(\alpha_k/\beta_k).
\end{equation*}	
Moreover, by Assumption \ref{assump3} and Lemma \ref{lemma_tool}, we have	
\begin{align*}
\mathbb{E}[ \|\mathcal{W}_{1,k}\|^2 ] 
& =  \sum_{i=0}^{k} \left( \prod_{j=i+1}^{k} (1-\beta_j) \right)^2 \beta_i^2 \mathbb{E}[\| \nabla F(\bx_i;\zeta_i) - \nabla f_{i} \|^2] \\
& \leq \sigma_{g}^2 \sum_{i=0}^{k} \left( \prod_{j=i+1}^{k} (1-\beta_j ) \right)^2 \beta_i^2 = \mathcal{O}\left(\beta_k\right).
\end{align*}
Combining the above three displays together, we complete the proof.

\subsection{Proof of Lemma \ref{lemma19}}\label{app:d5}
	
The subproblem \eqref{equ:SQP:new} at $\bm{x}_k$ with the averaged gradient $\Bar{\bm{g}}_{k}$ can be written as 
\begin{equation*}
\min_{ \barDelta\bx \in \Omega(\bx_k;\theta_k)} \frac{1}{2}\left\| \barDelta\bx + \bm{B}_k^{-1} \Bar{\bm{g}}_{k} \right\|_{\bm{B}_k}^2,
\end{equation*}
where $\|\bx\|_{\bB_k}^2 = \bx^\top\bB_k\bx$. By the optimality condition using variational inequalities, we have
\begin{equation*}
\left \langle \Delta\bx_k - \barDelta\bx_k, - \bm{B}_k^{-1} \Bar{\bm{g}}_{k} - \barDelta\bx_k \right \rangle_{\bm{B}_k} \leq 0.
\end{equation*}	
Similarly, since $\Delta\bx_k$ is the solution with true gradient $\nabla f_{k}$, it satisfies
\begin{equation*}
\left \langle  \barDelta\bx_k - \Delta\bx_k, - \bm{B}_k^{-1} \nabla f_{k} - \Delta\bx_k \right \rangle_{\bm{B}_k} \leq 0.
\end{equation*}	
Adding the two inequalities yields	 
\begin{align*}
0 & \geq \left \langle \Delta\bx_k - \barDelta\bx_k, - \bm{B}_k^{-1} \Bar{\bm{g}}_{k} - \barDelta\bx_k + \bm{B}_k^{-1} \nabla f_{k} + \Delta\bx_k \right \rangle_{\bm{B}_k} \\
& = \left \|  \Delta\bx_k - \barDelta\bx_k \right \|_{\bm{B}_k}^2 + \left \langle  \barDelta\bx_k - \Delta\bx_k, \Bar{\bm{g}}_{k} - \nabla f_{k}  \right \rangle\\
& \geq  \|  \barDelta\bx_k - \Delta\bx_k  \|_{\bm{B}_k}^2 - \|  \barDelta\bx_k - \Delta\bx_k \| \cdot \| \Bar{\bm{g}}_{k} - \nabla f_{k} \|.
\end{align*}	
Since $\bB_k\succeq \kappa_1\bI$, we know $\|\barDelta\bx_k - \Delta\bx_k \|_{\bm{B}_k}^2 \geq \kappa_1 \|  \barDelta\bx_k - \Delta\bx_k \|^2$. This completes the proof.

\subsection{Proof of Lemma \ref{lem:4}}\label{app:D6}

We follow the proof of Lemma \ref{lemma20} in Appendix \ref{app:pf:D4} and slightly abuse the notation of $\mathcal{W}_{1,k}$~and $\mathcal{W}_{2,k}$ there. In particular, we revisit the definition of $\Bar{\bm{g}}_k$ and have
\begin{align*}
\Bar{\bm{g}}_{k} - \nabla f^{\star} & = (1-\beta_k)( \Bar{\bm{g}}_{k-1} - \nabla f^{\star}) + \beta_k ( \nabla F(\bx_k;\zeta_k)- \nabla f_{k} ) + \beta_{k} (\nabla f_{k} - \nabla f^{\star})\\
& = \sum_{i=0}^{k} \left( \prod_{j=i+1}^{k} (1-\beta_j) \right) \beta_i (\nabla F(\bx_i;\zeta_i) - \nabla f_i ) \\
& \quad + \sum_{i=0}^{k} \left( \prod_{j=i+1}^{k} (1-\beta_j) \right) \beta_{i} ( \nabla f_{i} - \nabla f^{\star} ) + \prod_{j=0}^{k} (1-\beta_j) (\barg_{-1} - \nabla f^{\star})\\
& \coloneqq \mathcal{W}_{1,k} + \mathcal{W}_{2,k}.
\end{align*}
Applying Lemma \ref{lem:5} and the fact that $\nabla f_i\rightarrow\nabla f^\star$, we immediately have $\mathcal{W}_{2,k} \to \0$ as $k \to \infty$. For the term $\mathcal{W}_{1,k}$, we rewrite it as 
\begin{equation*}
\mathcal{W}_{1,k} = (1-\beta_{k}) \mathcal{W}_{1,k-1} + \beta_{k} \left(\nabla F(\bx_{k};\zeta_{k}) - \nabla f_{k} \right).
\end{equation*}
Thus, for large enough $k$, we have from Assumption \ref{assump3} that
\begin{align*}
\mathbb{E}_{k}[\|\mathcal{W}_{1,k}\|^2] & = (1 - \beta_{k})^2 \|\mW_{1,k-1}\|^2 + \beta_{k}^2 \mathbb{E}_{k}[\| \nabla F(\bx_{k};\zeta_{k}) - \nabla f_{k} \|^2]  \\
& \leq (1 - \beta_{k}) \|\mW_{1,k-1}\|^2 + \beta_{k}^2 \sigma_{g}^2.
\end{align*}
Since $\sum_{k=0}^{\infty} \beta_{k} = \infty$ and $\sum_{k=0}^{\infty} \beta_{k}^2 < \infty$, by Robbins-Siegmund theorem \citep{Robbins1971convergence}, we know $\|\mW_{1,k}\|^2$ is convergent and $\sum_{k=0}^{\infty} \beta_{k} \|\mW_{1,k-1}\|^2 < \infty$ almost surely. This further implies $\|\mW_{1,k}\| \to 0$ almost surely and we complete the proof.

\subsection{Proof of Lemma \ref{lem:5}}\label{app:d7}

For any $\epsilon>0$, there exists $i'>0$ such that $|e_i|\leq \epsilon$ and $\beta_i\in[0, 1]$ for any $i\geq i'$. Thus,~for~$k\geq i'$,
\begin{equation*}
\abr{a\prod_{j=0}^{k}(1-\beta_j)} \leq |a|\prod_{j=0}^{i'-1}|1-\beta_j|\cdot \prod_{j=i'}^{k}(1-\beta_j)\leq |a|\prod_{j=0}^{i'-1}|1-\beta_j|\cdot\exp\rbr{-\sum_{j=i'}^{k}\beta_j},
\end{equation*}
and
\begin{align*}
\abr{\sum_{i=0}^{k}\prod_{j=i+1}^{k}(1-\beta_j)\beta_ie_i} & \leq \sum_{i=0}^{i'-1}\prod_{j=i+1}^{k}|1-\beta_j|\beta_i|e_i| + \sum_{i=i'}^{k}\prod_{j=i+1}^{k}(1-\beta_j)\beta_i|e_i|\\
& \leq \prod_{j=i'}^{k}(1-\beta_j)\cdot\sum_{i=0}^{i'-1}\prod_{j=i+1}^{i'-1}|1-\beta_j|\beta_i|e_i| + \epsilon \sum_{i=i'}^{k}\prod_{j=i+1}^{k}(1-\beta_j)\beta_i\\
& = \prod_{j=i'}^{k}(1-\beta_j)\cdot\sum_{i=0}^{i'-1}\prod_{j=i+1}^{i'-1}|1-\beta_j|\beta_i|e_i| + \epsilon\cbr{1-\prod_{j=i'}^{k}(1-\beta_j)}\\
& \leq \exp\rbr{-\sum_{j=i'}^{k}\beta_j}\cdot\sum_{i=0}^{i'-1}\prod_{j=i+1}^{i'-1}|1-\beta_j|\beta_i|e_i| + \epsilon.
\end{align*}
Since $\sum_{i=0}^{\infty}\beta_i=\infty$, we can find $k'\geq i'$ large enough such that $\exp(-\sum_{j=i'}^{k}\beta_j)\cdot\{|a|\prod_{j=0}^{i'-1}|1-\beta_j| + \sum_{i=0}^{i'-1}\prod_{j=i+1}^{i'-1}|1-\beta_j|\beta_i|e_i| \}\leq \epsilon$ for any $k\geq k'$. Then, we obtain for $k\geq k'$ that
\begin{equation*}
\abr{a\prod_{j=0}^{k}(1-\beta_j)} + \abr{\sum_{i=0}^{k}\prod_{j=i+1}^{k}(1-\beta_j)\beta_ie_i}\leq 2\epsilon.
\end{equation*}
This shows $a\prod_{j=0}^{k}(1-\beta_j) + \sum_{i=0}^{k}\prod_{j=i+1}^{k}(1-\beta_j)\beta_ie_i\rightarrow0$ as $k\rightarrow\infty$. We complete the~proof.

\subsection{Proof of Lemma \ref{lem:6}}\label{app:D8}

We follow the proof of Lemma \ref{lem:4} in Appendix \ref{app:D6}. By the update rule \eqref{equ:gradHess:average}, we have
\begin{align*}
\barQ_{k} - \nabla^2 f^{\star} & = (1-\gamma_k) (\barQ_{k-1} - \nabla^2 f^{\star} ) + \gamma_{k} ( \nabla^2 F(\bx_k;\zeta_k)- \nabla^2 f_{k} )+ \gamma_{k} ( \nabla^2 f_{k} - \nabla^2 f^{\star} )\\
& = \sum_{i=0}^{k} \left( \prod_{j=i+1}^{k} (1-\gamma_j) \right) \gamma_i \left( \nabla^2 F(\bx_i;\zeta_i) - \nabla^2 f_i \right) \\
& \quad + \sum_{i=0}^{k} \left( \prod_{j=i+1}^{k} (1-\gamma_j) \right) \gamma_{i}\left( \nabla^2 f_{i} - \nabla^2 f^{\star} \right) + \prod_{j=0}^{k} (1-\gamma_j) (\barQ_{-1} - \nabla^2 f^{\star}) \\
& := \mathcal{V}_{1,k} + \mathcal{V}_{2,k}.
\end{align*}
Applying Lemma \ref{lem:5} and the fact that $\nabla^2 f_i\rightarrow\nabla^2 f^\star$, we immediately have $\mathcal{V}_{2,k}\rightarrow\0$ as~$k\rightarrow\infty$. For the term $\mathcal{V}_{1,k}$, we rewrite it as
\begin{equation*}
\mathcal{V}_{1,k} = (1-\gamma_{k}) \mathcal{V}_{1,k-1} + \gamma_{k} (\nabla^2 F(\bx_{k};\zeta_{k}) - \nabla^2 f_{k} ).
\end{equation*}
Thus, for large enough $k$, we have from Assumption \ref{assump7} that
\begin{align*}
\mathbb{E}_{k}[\|\mV_{1,k}\|_F^2] & = (1 - \gamma_{k})^2 \|\mV_{1,k-1}\|_F^2 + \gamma_{k}^2 \mathbb{E}_{k}[\| \nabla^2 F(\bx_{k};\zeta_{k}) - \nabla^2 f_{k} \|_F^2]  \\
& \leq (1 - \gamma_{k}) \|\mV_{1,k-1}\|_F^2 + \gamma_{k}^2 d\sigma_{H}^2.
\end{align*}
Since $\sum_{k=0}^{\infty} \gamma_{k} = \infty$ and $\sum_{k=0}^{\infty} \gamma_{k}^2 < \infty$, by Robbins-Siegmund theorem \citep{Robbins1971convergence}, we know $\|\mV_{1,k}\|_F^2$ is convergent and $\sum_{k=0}^{\infty} \gamma_{k} \|\mV_{1,k-1}\|_F^2 < \infty$ almost surely. This further implies $\|\mV_{1,k}\|_F \to 0$ almost surely and we complete the proof.

\subsection{Proof of Lemma \ref{lem:c9}}\label{app:d9}

We introduce a lemma that shows the recursive relation between $\|\bw_k-\tw\|^2$ and $\|\barg_k-\nabla f_k\|^2$. See Appendix \ref{app:d10} for the proof.

\begin{lemma}\label{lem:d3}
Under Assumption \ref{assump3} and suppose $\alpha_k = \iota_1(k+1)^{-b_1}$, $\beta_k=\iota_2(k+1)^{-b_2}$ satisfy $\iota_1, \iota_2, b_1>0$, $b_2\in(0, 1)$. There exists a deterministic integer $\bar{k}_0>0$ such that for any $k_0\geq \bar{k}_0$, the following result holds for some deterministic constant $\Upsilon_1(k_0)>0$: for any $k\geq k_0$, $\quad\quad\quad$
\begin{align*}
\mE[& \|\bw_{k+1} - \tw\|^2\1_{\tau_{k_0}>k+1}]\\
& \leq \cbr{1 - 2(1-2/\Upsilon)\alpha_k}\mE[\|\bw_k-\tw\|^2\1_{\tau_{k_0}>k}] + (2+\psi)\Upsilon^3\alpha_k\mE[\|\barg_k-\nabla f_k\|^2\1_{\tau_{k_0}>k}],\\
\mE[& \|\barg_{k+1} - \nabla f_{k+1}\|^2 \bm{1}_{\tau_{k_0} > k+1}] \\
& \leq \Upsilon_1(k_0)\cbr{\beta_k+\rbr{\sum_{i=k_0}^{k}\prod_{j=i+1}^{k}(1-\beta_j) \alpha_{i} \cbr{\mE[(\|\barg_i - \nabla f_i \|^2 + \|\bw_i - \bw^{\star}\|^2)\1_{\tau_{k_0}>i}]}^{1/2} }^2},
\end{align*}
where $\Upsilon\geq 2$ is defined in the stopping time $\tau_{k_0}$ in \eqref{equ:tau}.
\end{lemma}

With the above recursion, we claim that for any $q\geq 0$, there exists a deterministic integer $\bar{k}_0>0$ such that for any $k_0\geq \bar{k}_0$, we have for some constant $\Upsilon_2(k_0)>0$ that
\begin{equation}\label{pf:ind}
\mE\big[(\|\bw_k-\tw\|^2 + \|\barg_k-\nabla f_k\|^2)\1_{\tau_{k_0}>k}\big] \leq \Upsilon_2(k_0)\rbr{\beta_k + (\alpha_k/\beta_k)^{2q}},\quad \quad \forall k\geq k_0.
\end{equation}
We prove the above statement by induction. When $q=0$, we apply the definition of $\tau_{k_0}$ in \eqref{equ:tau} and have $\mE[\|\bw_k-\tw\|^2\1_{\tau_{k_0}>k}]\leq 1/\Upsilon$ for any $k\geq k_0$. Furthermore, by the recursive~form of $\barg_k$ in~\eqref{equ:gradHess:average} and Assumption \ref{assump3}, for any $k\geq k_0$,
\begin{align*}
\mE[\|\barg_k\|^2\1_{\tau_{k_0}>k}] & \leq (1-\beta_k)\mE[\|\barg_{k-1}\|^2\1_{\tau_{k_0}>k-1}] + \beta_k\mE[\|\nabla F(\bx_k;\zeta_k)\|^2\1_{\tau_{k_0}>k}]\\
& \leq (1-\beta_k)\mE[\|\barg_{k-1}\|^2\1_{\tau_{k_0}>k-1}] + 3\beta_k(\sigma_g^2+\|\nabla f_k-\nabla f^\star\|^2\1_{\tau_{k_0}>k} + \|\nabla f^\star\|^2)\\
& \stackrel{\mathclap{\eqref{equ:tau}}}{\leq} (1-\beta_k)\mE[\|\barg_{k-1}\|^2\1_{\tau_{k_0}>k-1}] + 3\beta_k\rbr{\sigma_g^2+1/\Upsilon + \|\nabla f^\star\|^2}.
\end{align*}
The above display, together with the fact that $\mE[\|\barg_{k_0}\|^2]\leq \Upsilon_3(k_0)$ for some constant $\Upsilon_3(k_0)$ due to Assumption \ref{assump3}, implies that $\mE[\|\barg_k-\nabla f_k\|^2\1_{\tau_{k_0}>k}]\leq \Upsilon_4(k_0)$, $\forall k\geq k_0$ for some constant $\Upsilon_4(k_0)$. This verifies \eqref{pf:ind} for $q=0$. Suppose that \eqref{pf:ind} holds for $q \geq 0$, then we establish the results for $q+1$. In particular, we apply Lemmas \ref{lem:d3} and \ref{lemma_tool} and have ($\lesssim$ hides deterministic constants that may depend on $k_0$)
\begin{align*}
\mE[\|\barg_{k+1} & - \nabla f_{k+1}\|^2 \bm{1}_{\tau_{k_0} > k+1}] \\
& \lesssim \beta_k+\rbr{\sum_{i=k_0}^{k}\prod_{j=i+1}^{k}(1-\beta_j) \alpha_{i} \cbr{\mE[(\|\barg_i - \nabla f_i \|^2 + \|\bw_i - \bw^{\star}\|^2)\1_{\tau_{k_0}>i}]}^{1/2} }^2\\
& \lesssim \beta_k+\rbr{\sum_{i=k_0}^{k}\prod_{j=i+1}^{k}(1-\beta_j) \alpha_{i} \cbr{\beta_{i} + \left(\frac{\alpha_i}{\beta_i}\right)^{2q}}^{1/2} }^2  \lesssim \beta_k + \left(\frac{\alpha_k}{\beta_k}\right)^{2(q+1)}.
\end{align*}
Moreover, we still apply Lemma \ref{lem:d3} and have for any $b>0$ and $k\geq k_0$,
\begin{align*}
\mE[& \|\bw_{k+1} - \tw\|^2\1_{\tau_{k_0}>k+1}]\\
& \leq \cbr{1 - 2(1-2/\Upsilon)\alpha_k}\mE[\|\bw_k-\tw\|^2\1_{\tau_{k_0}>k}] + (2+\psi)\Upsilon^3\alpha_k\mE[\|\barg_k-\nabla f_k\|^2\1_{\tau_{k_0}>k}]\\
& \leq (2+\psi)\Upsilon^3\sum_{i=k_0}^{k} \prod_{j=i+1}^{k} \cbr{1 - 2(1-2/\Upsilon)\alpha_j} \alpha_{i} \mE[\|\barg_i-\nabla f_i\|^2\1_{\tau_{k_0}>i}] \\
& \quad + \prod_{j=k_0}^{k} \cbr{1 - 2(1-2/\Upsilon)\alpha_j} \mE[\|\bw_{k_0}-\tw\|^2\1_{\tau_{k_0}>k_0}] \\
& \stackrel{\mathclap{\eqref{pf:int}}}{\lesssim}\; \sum_{i=k_0}^{k} \prod_{j=i+1}^{k} \cbr{1 - 2(1-2/\Upsilon)\alpha_j} \alpha_{i} \mE[\|\barg_i-\nabla f_i\|^2\1_{\tau_{k_0}>i}] + \left(k^{-b} \bm{1}_{b_1<1} + k^{-2(1-2/\Upsilon)\iota_1} \bm{1}_{b_1=1}\right)\\
& \lesssim \sum_{i=0}^{k} \prod_{j=i+1}^{k} \cbr{1 - 2(1-2/\Upsilon)\alpha_j} \alpha_{i} \cbr{\beta_i + \left(\frac{\alpha_i}{\beta_i}\right)^{2(q+1)}} + \left(k^{-b} \bm{1}_{b_1<1} + k^{-2(1-2/\Upsilon)\iota_1} \bm{1}_{b_1=1}\right)\\
&  \lesssim \beta_k + \left(\alpha_k/\beta_k\right)^{2(q+1)},
\end{align*}
where the last inequality is because when $b_1=1$, we have $2(1-2/\Upsilon)\iota_1>1>b_2$ by choosing $\Upsilon$ is large enough. This completes the induction step of \eqref{pf:ind} and further completes the proof.

\subsection{Proof of Lemma \ref{lem:d3}}\label{app:d10}

By the local update \eqref{equ:local:update}, we know for any $k\geq k_0$,
\begin{align*}
& \|\bw_{k+1}-\tw\|^2 = \|\bw_k - \tw + \baralpha_k\barDelta\bw_k\|^2 = \|\bw_k-\tw - \baralpha_k \bH_k^{-1}\bnabla_{\bw}\mL_k\|^2 \nonumber\\
& = \|\bw_k-\tw - \baralpha_k \bH_k^{-1} \nabla_\bw\mL_k - \baralpha_k \bH_k^{-1}(\bnabla_{\bw}\mL_k-\nabla_\bw\mL_k)\|^2 \nonumber\\
& = \|\bw_k-\tw - \baralpha_k \bH_k^{-1} \nabla_\bw\mL_k\|^2 + \baralpha_k^2\|\bH_k^{-1}(\bnabla_{\bw}\mL_k-\nabla_\bw\mL_k)\|^2 \nonumber \\
& \quad - 2\baralpha_k\langle\bw_k-\tw - \baralpha_k \bH_k^{-1} \nabla_\bw\mL_k,  \bH_k^{-1}(\bnabla_{\bw}\mL_k-\nabla_\bw\mL_k)\rangle \nonumber\\
& \leq \rbr{1+\frac{1.5\baralpha_k}{\Upsilon}}\|\bw_k-\tw - \baralpha_k \bH_k^{-1} \nabla_\bw\mL_k\|^2 + \rbr{\baralpha_k^2 + \frac{\Upsilon\baralpha_k}{1.5}}\|\bH_k^{-1}(\bnabla_{\bw}\mL_k-\nabla_\bw\mL_k)\|^2.
\end{align*}
For the first term on the right hand side, we have
\begin{align*}
& \|\bw_k-\tw - \baralpha_k \bH_k^{-1} \nabla_\bw\mL_k\|^2\1_{\tau_{k_0}>k}\\
& = \rbr{\|\bw_k-\tw\|^2 - 2\baralpha_k \langle \bw_k-\tw, \bH_k^{-1}\nabla_\bw\mL_k\rangle + \baralpha_k^2\|\bH_k^{-1}\nabla_\bw\mL_k\|^2} \1_{\tau_{k_0}>k}\\
& = \rbr{(1-2\baralpha_k)\|\bw_k-\tw\|^2 + 2\baralpha_k\langle \bw_k-\tw, \bw_k-\tw - \bH_k^{-1}\nabla_\bw\mL_k\rangle + \baralpha_k^2\|\bH_k^{-1}\nabla_\bw\mL_k\|^2 }\1_{\tau_{k_0}>k}\\
& \stackrel{\mathclap{\eqref{equ:tau}}}{\leq}\; \rbr{1-2\baralpha_k+\frac{2\baralpha_k}{\Upsilon} + \Upsilon^4\baralpha_k^2}\|\bw_k-\tw\|^2\1_{\tau_{k_0}>k}.
\end{align*}
For the second term on the right hand side, we have
\begin{equation*}
\|\bH_k^{-1}(\bnabla_{\bw}\mL_k-\nabla_\bw\mL_k)\|^2\1_{\tau_{k_0}>k}\stackrel{\mathclap{\eqref{equ:tau}}}{\leq} \Upsilon^2\|\barg_k-\nabla f_k\|^2\1_{\tau_{k_0}>k}.
\end{equation*}
Combining the above three displays and noting that $\1_{\tau_{k_0}>k+1}\leq \1_{\tau_{k_0}>k}$, we obtain
\begin{align*}
\|\bw_{k+1} - \tw\|^2\1_{\tau_{k_0}>k+1}  & \leq \rbr{1+\frac{1.5\baralpha_k}{\Upsilon}}\rbr{1-2\baralpha_k+\frac{2\baralpha_k}{\Upsilon} + \Upsilon^4\baralpha_k^2}\|\bw_k-\tw\|^2\1_{\tau_{k_0}>k} \\
&\quad + \Upsilon^2\rbr{\baralpha_k^2 + \frac{\Upsilon\baralpha_k}{1.5}}\|\barg_k-\nabla f_k\|^2\1_{\tau_{k_0}>k}\\
& \leq \rbr{1-2\baralpha_k + \frac{3.5\baralpha_k}{\Upsilon} + \Upsilon^4\baralpha_k^2 + \frac{3\baralpha_k^2}{\Upsilon^2}+1.5\Upsilon^3\baralpha_k^3}
\|\bw_k-\tw\|^2\1_{\tau_{k_0}>k}\\
& \quad + \Upsilon^2\rbr{\baralpha_k^2 + \frac{\Upsilon\baralpha_k}{1.5}}\|\barg_k-\nabla f_k\|^2\1_{\tau_{k_0}>k}.
\end{align*}
To simplify the above display, we let $\bar{k}_0$ be large enough (with a deterministic threshold) such~that
\begin{equation}\label{con:k0}
\Upsilon^4\baralpha_k + \frac{3\baralpha_k}{\Upsilon^2}+1.5\Upsilon^3\baralpha_k^2\leq \frac{0.5}{\Upsilon}\quad \text{ and }\quad \rbr{\baralpha_k+\frac{\Upsilon}{1.5}}\baralpha_k\leq (2+\psi)\Upsilon\alpha_k, \quad\; \forall k\geq \bar{k}_0.
\end{equation}
This is achievable since $b_1>0$ and $\limsup_{k\rightarrow\infty}\baralpha_k/\alpha_k\leq 1+\psi$. Then, we obtain
\begin{align*}
\|\bw_{k+1} & - \tw\|^2\1_{\tau_{k_0}>k+1} \\
& \leq \cbr{1 - 2(1-2/\Upsilon)\baralpha_k}\|\bw_k-\tw\|^2\1_{\tau_{k_0}>k} + (2+\psi)\Upsilon^3\alpha_k\|\barg_k-\nabla f_k\|^2\1_{\tau_{k_0}>k}\\
& \leq \cbr{1 - 2(1-2/\Upsilon)\alpha_k}\|\bw_k-\tw\|^2\1_{\tau_{k_0}>k} + (2+\psi)\Upsilon^3\alpha_k\|\barg_k-\nabla f_k\|^2\1_{\tau_{k_0}>k},
\end{align*}
where the last inequality uses $\Upsilon\geq 2$. This shows the first argument of the lemma. For the~second argument of the lemma, we recall \eqref{equ:gk} from the proof in Appendix \ref{app:pf:D4} and have 
\begin{align}\label{d3:1}
\bar{\bm{g}}_{k+1} - \nabla f_{k+1} & = \sum_{i=0}^{k+1} \prod_{j=i+1}^{k+1} (1-\beta_j) \beta_i \left( \nabla F(\bx_i;\zeta_i) - \nabla f_i \right) + \sum_{i=0}^{k+1} \prod_{j=i}^{k+1} (1-\beta_j) \left( \nabla f_{i-1} - \nabla f_{i} \right) \nonumber\\
& \coloneqq \mathcal{W}_{1,k+1} + \mathcal{W}_{2,k+1}.
\end{align}
By Assumption \ref{assump3} and Lemma \ref{lemma_tool}, we have
\begin{align}\label{d3:4}
\mathbb{E}[\|\mathcal{W}_{1,k+1}\|^2\1_{\tau_{k_0}>k+1}] & \leq \mathbb{E}[\|\mathcal{W}_{1,k+1}\|^2] \nonumber\\
& =  \sum_{i=0}^{k+1} \prod_{j=i+1}^{k+1} (1-\beta_j)^2 \beta_i^2 \mathbb{E}[\| \nabla F(\bx_i;\zeta_i) - \nabla f_{i} \|^2] = \mathcal{O}\left(\beta_k\right).
\end{align}
For the term $\mW_{2,k+1}$, we use $\beta_k\leq 1$ for $k\geq k_0$ and separate the sum in $\mW_{2,k+1}$ into two~terms:
\begin{align}\label{d3:2}
& \|\mathcal{W}_{2,k+1}\|\1_{\tau_{k_0}>k+1} \nonumber\\
& \leq \sum_{i=0}^{k_0}\prod_{j=i}^{k+1} |1-\beta_j|\|\nabla f_{i-1}-\nabla f_i\| + \sum_{i=k_0+1}^{k+1}\prod_{j=i}^{k+1} (1-\beta_j)\|\nabla f_{i-1}-\nabla f_i\|\1_{\tau_{k_0}>k+1} \nonumber\\
& \stackrel{\mathclap{\eqref{equ:tau}}}{\leq}\;\; \prod_{j=0}^{k+1}|1-\beta_j|\sum_{i=0}^{k_0}\prod_{j=0}^{i-1}|1-\beta_j|^{-1}\|\nabla f_{i-1}-\nabla f_i\| + \Upsilon\sum_{i=k_0+1}^{k+1}\prod_{j=i}^{k+1}(1-\beta_j)\baralpha_{i-1}\|\barDelta\bx_{i-1}\|\1_{\tau_{k_0}>k+1} \nonumber\\
& \leq \Upsilon_1(k_0)\prod_{j=0}^{k+1}|1-\beta_j| + \Upsilon\sum_{i=k_0}^{k}\prod_{j=i+1}^{k+1}(1-\beta_j)\baralpha_i\|\barDelta\bx_i\|\1_{\tau_{k_0}>i} \nonumber\\
& \stackrel{\mathclap{\substack{\eqref{equ:tau} \\ \eqref{con:k0}}}}{\leq}\; \Upsilon_1(k_0)\prod_{j=0}^{k+1}|1-\beta_j| + (3+1.5\psi)\Upsilon\sum_{i=k_0}^{k}\prod_{j=i+1}^{k}(1-\beta_j)\alpha_i\|\barDelta\bx_i\|\1_{\tau_{k_0}>i},
\end{align}
where the second last inequality holds for some deterministic constant $\Upsilon_1(k_0)>0$ due to the boundedness of $\nabla f_i$. For the first term on the right hand side, we apply \eqref{pf:int} and know~that $\prod_{j=0}^{k+1}|1-\beta_j| =o(\beta_k)$. For the second term on the right hand side, we have for $k_0\leq i <\tau_{k_0}$,$\quad\quad$
\begin{align}\label{equ:d4}
\|\barDelta\bx_i\|^2 = \|\bH_i^{-1} \bnabla_{\bw}\mL_i\|^2 & \; \stackrel{\mathclap{\eqref{equ:tau}}}{\leq}\;  2\Upsilon^2 (\|\bnabla_{\bw}\mL_i - \nabla_{\bw}\mL_i\|^2 + \|\nabla_\bw\mL_i\|^2) \nonumber\\
&\; \stackrel{\mathclap{\eqref{equ:tau}}}{\leq}\;2\Upsilon^4(\|\barg_i-\nabla f_i\|^2 + \|\bw_i -\tw\|^2).
\end{align}
Therefore, we have
\begin{align}\label{d3:3}
& \mE\sbr{\rbr{\sum_{i=k_0}^{k}\prod_{j=i+1}^{k}(1-\beta_j)\alpha_i\|\barDelta\bx_i\|\1_{\tau_{k_0}>i}}^2} \leq \rbr{\sum_{i=k_0}^{k}\prod_{j=i+1}^{k}(1-\beta_j)\alpha_i\cbr{\mE[\|\barDelta\bx_i\|^2\1_{\tau_{k_0}>i}]}^{1/2} }^2 \nonumber\\
& \stackrel{\mathclap{\eqref{equ:d4}}}{\leq}2\Upsilon^4\rbr{\sum_{i=k_0}^{k}\prod_{j=i+1}^{k}(1-\beta_j)\alpha_i\cbr{\mE[(\|\barg_i-\nabla f_i\|^2 + \|\bw_i -\tw\|^2)\1_{\tau_{k_0}>i}]}^{1/2} }^2.
\end{align}
Finally, combining \eqref{d3:1}, \eqref{d3:4}, \eqref{d3:2}, \eqref{d3:3}, we complete the proof.

\subsection{Proof of Lemma \ref{lem:c10}}\label{app:d11}

We follow the proof in Appendix \ref{app:D8} and have
\begin{align*}
\barQ_{k} - \nabla^2 f^{\star} & = \sum_{i=0}^{k} \prod_{j=i+1}^{k} (1-\gamma_j) \gamma_i \left( \nabla^2 F(\bx_i;\zeta_i) - \nabla^2 f_i \right) \\
& + \sum_{i=0}^{k} \prod_{j=i+1}^{k} (1-\gamma_j) \gamma_{i}\left( \nabla^2 f_{i} - \nabla^2 f^{\star} \right) + \prod_{j=0}^{k} (1-\gamma_j) (\barQ_{-1} - \nabla^2 f^{\star}) \coloneqq \mathcal{V}_{1,k} + \mathcal{V}_{2,k}.
\end{align*}
We analyze each term separately. For the term $\mV_{1,k}$, we apply Assumption \ref{assump7} and have
\begin{equation*}
\mathbb{E}[\| \mathcal{V}_{1,k}\|_F^2]  = \sum_{i=0}^{k} \prod_{j=i+1}^{k} (1-\gamma_j)^2 \gamma_i^2 \mathbb{E}[\| \nabla^2 F(\bx_i;\zeta_i) - \nabla^2 f_i \|_F^2] = \begin{cases*}
\mO(\gamma_k) & \hskip-8pt \text{if } $b_3<1$ \\
o(\sqrt{\alpha_k}) & \hskip-8pt \text{if } $b_3=1$
\end{cases*}=o(\sqrt{\alpha_k}),
\end{equation*}
where the second last equality uses Lemma \ref{lemma_tool} and the condition that $\iota_3>0.25b_1$ if $b_3=1$;~and the last equality uses the condition $b_3>0.5b_1$. For the term $\mV_{2,k}$, we apply \eqref{pf:int} and~have$\quad\quad$
\begin{equation}\label{equ:c10}
\prod_{j=0}^{k} |1-\gamma_j|^2= \begin{cases*}
o(\gamma_k) & \text{if } $b_3<1$\\
\mO\rbr{\frac{1}{k^{2\iota_3}}} & \text{if } $b_3=1$
\end{cases*} = o(\sqrt{\alpha_k}).
\end{equation}
Next, we analyze the first term in $\mV_{2,k}$. We have for some constants $\Upsilon_1(k_0), \Upsilon_2(k_0)>0$ that
\begin{align*}
& \mE\sbr{\nbr{\sum_{i=0}^{k} \prod_{j=i+1}^{k} (1-\gamma_j) \gamma_{i}(\nabla^2 f_{i} - \nabla^2 f^{\star} )}^2\1_{\tau_{k_0}>k}} \\
& \leq 2 \mE\sbr{\nbr{\sum_{i=0}^{k_0}\prod_{j=i+1}^{k}(1-\gamma_j)\gamma_i(\nabla^2 f_{i} - \nabla^2 f^{\star} ) }^2 + \nbr{\sum_{i=k_0}^{k}\prod_{j=i+1}^{k}(1-\gamma_j)\gamma_i(\nabla^2 f_{i} - \nabla^2 f^{\star})}^2\1_{\tau_{k_0}>k}}\\
& \stackrel{\mathclap{\eqref{equ:c10}}}{\leq} o(\Upsilon_1(k_0)\sqrt{\alpha_k}) + 2\rbr{\sum_{i=k_0}^{k}\prod_{j=i+1}^{k}(1-\gamma_j)\gamma_i\cbr{\mE[\|\nabla f_i-\nabla f^\star\|^2\1_{\tau_{k_0}>i}]}^{1/2}}^2\\
&\stackrel{\mathclap{\eqref{equ:tau}}}{\leq}o(\Upsilon_1(k_0)\sqrt{\alpha_k}) + 2\Upsilon^2\rbr{\sum_{i=k_0}^{k}\prod_{j=i+1}^{k}(1-\gamma_j)\gamma_i\cbr{\mE[\|\bx_i-\tx\|^2\1_{\tau_{k_0}>i}]}^{1/2}}^2 = o(\Upsilon_2(k_0)\sqrt{\alpha_k}),
\end{align*}
where the last equality is due to Lemmas \ref{lemma_tool}, \ref{lem:c9}, and the facts that
\begin{enumerate}[label=(\alph*),topsep=0pt]
\setlength\itemsep{0.0em}
\item $b_3\in(0,1)$: We apply $b_2>0.5b_1$ and have
\begin{equation*}
\sum_{i=k_0}^{k}\prod_{j=i+1}^{k}(1-\gamma_j)\gamma_i\sqrt{\beta_i} = O(\sqrt{\beta_k}) = o(\alpha_k^{0.25}).
\end{equation*}
\item $b_3=1$: We apply $b_2>0.5b_1$ and $\iota_3>0.25b_1\Leftrightarrow 1-\frac{0.5b_1}{b_2}\cdot\frac{0.5b_2}{\iota_3}>0$, and have
\begin{equation*}
\frac{1}{\alpha_k^{0.25}}\sum_{i=k_0}^{k}\prod_{j=i+1}^{k}(1-\gamma_j)\gamma_i\sqrt{\beta_i} = \frac{(\sqrt{\beta_k})^{0.5b_1/b_2}}{\alpha_k^{0.25}}\frac{1}{(\sqrt{\beta_k})^{0.5b_1/b_2}}\sum_{i=k_0}^{k}\prod_{j=i+1}^{k}(1-\gamma_j)\gamma_i\sqrt{\beta_i} = o(1).
\end{equation*}
\end{enumerate}
Thus, we have shown that $\mE[\|\barQ_{k} - \nabla^2 f^{\star}\|^2\1_{\tau_{k_0}>k}] = o(\Upsilon_3(k_0)\sqrt{\alpha_k})$ for some constant $\Upsilon_3(k_0)>0$. Finally, by noting from \eqref{equ:tau} that 
\begin{align*}
\|\bH_k-\tbH\|\1_{\tau_{k_0}>k} & \leq \rbr{\|\barQ_{k} - \nabla^2 f^{\star}\| + \|\nabla_\bw^2\mL_k - \nabla_\bw^2\mL^\star\| + \|\nabla^2 f_k - \nabla^2f^\star\|}\1_{\tau_{k_0}>k} \\
& \leq \rbr{\|\barQ_{k} - \nabla^2 f^{\star}\| + 2\Upsilon\|\bw_k-\tw\|}\1_{\tau_{k_0}>k},
\end{align*}
and applying Lemma \ref{lem:c9}, we complete the proof.

\subsection{Proof of Lemma \ref{lem:c11}}\label{app:d12}

Recalling the definition of $\mJ_{3,k}$ from \eqref{pf:equ:J}, we have
\begin{align*}
\mJ_{3,k} & = \sum_{i=0}^{k}\prod_{j=i+1}^{k}(1-\alpha_j)\alpha_i(\tbH)^{-1}(\bnabla_{\bw}\mL_i-\nabla_\bw\mL_i)\\
& \stackrel{\mathclap{\eqref{def:cov}}}{=}\; \sum_{i=0}^{k}\prod_{j=i+1}^{k}(1-\alpha_j)\alpha_i\begin{pmatrix}
\nabla_{\bx}^2\mL^\star	& (\bJ^\star)^\top\\
\bJ^\star & \0
\end{pmatrix}^{-1}\begin{pmatrix}
\barg_i-\nabla f_i\\
\0
\end{pmatrix}\\
& \stackrel{\mathclap{\eqref{equ:gk}}}{=}\; \sum_{i=0}^{k}\prod_{j=i+1}^{k}(1-\alpha_j)\alpha_i\begin{pmatrix}
\nabla_{\bx}^2\mL^\star	& (\bJ^\star)^\top\\
\bJ^\star & \0
\end{pmatrix}^{-1}\cbr{\begin{pmatrix}
\mW_{1,i}\\
\0
\end{pmatrix}+\begin{pmatrix}
\mW_{2,i}\\
\0
\end{pmatrix}}= \mJ_{3,k}^{(1)} + \mJ_{3,k}^{(2)}.
\end{align*}
We first analyze $\mJ_{3,k}^{(2)}$ and aim to show that $\mJ_{3,k}^{(2)}=o_p(\sqrt{\alpha_k})$. By Lemma \ref{lem:c8}, it suffices to~show $\mW_{2,k}\1_{\tau_{k_0}>k} = o_p(\sqrt{\alpha_k})$. From the definition of $\mW_{2,k}$ in \eqref{equ:gk} and Lemma \ref{lem:c8}, this result~is~further implied by
\begin{equation*}
\|\nabla f_{k}-\nabla f_{k+1}\| \1_{\tau_{k_0}>k+1} \stackrel{\eqref{equ:tau}}{\leq} \Upsilon\baralpha_k\|\barDelta\bx_k\|\1_{\tau_{k_0}>k}\leq  o_p(\beta_k\sqrt{\alpha_k}).
\end{equation*}
By \eqref{equ:d4}, Lemma \ref{lem:c9}, and the fact that $\alpha_k\sqrt{\beta_k} = o(\beta_k\sqrt{\alpha_k})$, we obtain the above desired~result. Next, we analyze $\mJ_{3,k}^{(1)}$. By the definition of $\mW_{1,k}$ in \eqref{equ:gk}, we have
\begin{align*}
\mJ_{3,k}^{(1)} & = \sum_{i=0}^{k}\prod_{j=i+1}^{k}(1-\alpha_j)\alpha_i\begin{pmatrix}
\nabla_{\bx}^2\mL^\star	& (\bJ^\star)^\top\\
\bJ^\star & \0
\end{pmatrix}^{-1}\begin{pmatrix}
\mW_{1,i}\\
\0
\end{pmatrix}\\
& = \sum_{i=0}^{k}\prod_{j=i+1}^{k}(1-\alpha_j)\alpha_i \sum_{h=0}^{i} \prod_{l=h+1}^{i} (1-\beta_l) \beta_h\underbrace{\begin{pmatrix}
\nabla_{\bx}^2\mL^\star	& (\bJ^\star)^\top\\
\bJ^\star & \0
\end{pmatrix}^{-1}\begin{pmatrix}
\nabla F(\bx_h;\zeta_h) - \nabla f_h\\
\0
\end{pmatrix}}_{\bm{\phi}^{\star}_{h}}\\
& = \sum_{h=0}^{k} \underbrace{\sum_{i = h}^{k}\prod_{j=i+1}^{k} \left(1-\alpha_j \right) \alpha_{i} \prod_{l=h+1}^{i} (1-\beta_l) \beta_h}_{a_{h,k}}\tbphi_h = \sum_{h=0}^{k} a_{h,k}\tbphi_h.
\end{align*}
We first claim that $\mathbb{E}_{h}[\bm{\phi}^{\star}_{h}\bm{\phi}^{\star\top}_{h}] \to \bm{\Omega}^{\star}$ almost surely as $h \to \infty$. In fact, by Assumptions \ref{assump5},~\ref{assump9}, and the smoothness of $\nabla f(\bx)$, we have
\begin{align*}
& \left\|\mathbb{E}_{h}[\nabla F(\bx_{h},\zeta_{h}) \nabla F(\bx_{h},\zeta_{h})^{\top} - \nabla f_{h} \nabla f_{h}^{\top}] - \mathbb{E}[ \nabla F(\bm{x}^{\star};\zeta) \nabla F(\bm{x}^{\star};\zeta)^{\top} \right. \left. - \nabla f^{\star} \nabla f^{\star \top}] \right\| \\
& \leq \left\|\mathbb{E}_{h}[\nabla F(\bm{x}_{h};\zeta_{h}) \nabla F(\bm{x}_{h};\zeta_{h})^{\top}] -  \mE[\nabla F(\bm{x}^{\star};\zeta) \nabla F(\tx;\zeta)^{\top}] \right\| \\
& \quad + \rbr{\| \nabla f_{h} \| + \| \nabla f^{\star} \|} \left\|\nabla f_{h} - \nabla f^{\star} \right\| \to 0 \quad ~\text{as}\quad h \to \infty.
\end{align*}
With this result, we then analyze the conditional variance process. We have
\begin{align*}
& \frac{1}{\alpha_k}\sum_{h=0}^{k} a_{h,k}^2\mE_h[\bm{\phi}^{\star}_{h}\bm{\phi}^{\star\top}_{h}]\\
& = \frac{1}{\alpha_k} \sum_{h=0}^{k}\sum_{i=h}^k\sum_{i'=h}^{k}\prod_{j=i+1}^{k}(1-\alpha_j)\alpha_i\prod_{l=h+1}^{i}(1-\beta_l)\beta_h\prod_{j'=i'+1}^{k}(1-\alpha_{j'})\alpha_{i'}\prod_{l'=h+1}^{i'}(1-\beta_{l'})\beta_h\mE_h[\bm{\phi}^{\star}_{h}\bm{\phi}^{\star\top}_{h}]\\
& = \frac{1}{\alpha_k}\sum_{i=0}^{k}\sum_{i'=0}^{k}\prod_{j=i+1}^{k}(1-\alpha_j)\alpha_i\prod_{j'=i'+1}^{k}(1-\alpha_{j'})\alpha_{i'}\sum_{h=0}^{\min\{i,i'\}}\prod_{l=h+1}^{i}(1-\beta_l)\prod_{l'=h+1}^{i'}(1-\beta_{l'})\beta_h^2\mE_h[\bm{\phi}^{\star}_{h}\bm{\phi}^{\star\top}_{h}]\\
& = \frac{2}{\alpha_k}\sum_{i=0}^{k}\sum_{i'=0}^{i}\prod_{j=i+1}^{k}(1-\alpha_j)\alpha_i\prod_{j'=i'+1}^{k}(1-\alpha_{j'})\alpha_{i'}\sum_{h=0}^{i'}\prod_{l=h+1}^{i}(1-\beta_l)\prod_{l'=h+1}^{i'}(1-\beta_{l'})\beta_h^2\mE_h[\bm{\phi}^{\star}_{h}\bm{\phi}^{\star\top}_{h}]\\
& \quad - \frac{1}{\alpha_k}\sum_{i=0}^{k}\prod_{j=i+1}^{k}(1-\alpha_j)^2\alpha_i^2\sum_{h=0}^{i}\prod_{l=h+1}^{i}(1-\beta_l)^2\beta_h^2\mE_h[\bm{\phi}^{\star}_{h}\bm{\phi}^{\star\top}_{h}]\\
& = \frac{2}{\alpha_k}\sum_{i=0}^{k}\prod_{j=i+1}^{k}(1-\alpha_j)^2\alpha_i\sum_{i'=0}^{i}\prod_{j'=i'+1}^{i}(1-\alpha_{j'})(1-\beta_{j'})\alpha_{i'}\sum_{h=0}^{i'}\prod_{l'=h+1}^{i'}(1-\beta_{l'})^2\beta_h^2\mE_h[\bm{\phi}^{\star}_{h}\bm{\phi}^{\star\top}_{h}]\\
& \quad - \frac{1}{\alpha_k}\sum_{i=0}^{k}\prod_{j=i+1}^{k}(1-\alpha_j)^2\alpha_i^2\sum_{h=0}^{i}\prod_{l=h+1}^{i}(1-\beta_l)^2\beta_h^2\mE_h[\bm{\phi}^{\star}_{h}\bm{\phi}^{\star\top}_{h}].
\end{align*}
We apply Lemma \ref{lemma_tool} and note that
\begin{align*}
& \lim\limits_{i\rightarrow\infty}\frac{1}{\beta_i}\sum_{h=0}^{i}\prod_{l=h+1}^{i}(1-\beta_l)^2\beta_h^2\mE_h[\bm{\phi}^{\star}_{h}\bm{\phi}^{\star\top}_{h}] = 0.5\tOmega,\\
& \lim\limits_{i\rightarrow \infty}\frac{1}{\alpha_i}\sum_{i'=0}^{i}\prod_{j'=i'+1}^{i}(1-\alpha_{j'})(1-\beta_{j'})\alpha_{i'}\beta_{i'} = 1,\\
& \lim\limits_{k\rightarrow\infty}\frac{1}{\alpha_k} \sum_{i=0}^{k}\prod_{j=i+1}^{k}(1-\alpha_j)^2\alpha_i^2 = \eta \coloneqq \begin{cases*}
0.5, & \text{if }\; $b_1\in(0,1$,\\
\frac{\iota_{1}}{2\iota_{1}-1}, & \text{if }\; $b_1=1$,
\end{cases*}\\
& \lim\limits_{k\rightarrow\infty}\frac{1}{\alpha_k} \sum_{i=0}^{k}\prod_{j=i+1}^{k}(1-\alpha_j)^2\alpha_i^2\beta_i = 0.
\end{align*}
Combining the above two displays, we obtain almost surely,
\begin{equation}\label{dequ:2}
\lim\limits_{k\rightarrow\infty}\frac{1}{\alpha_k}\sum_{h=0}^{k} a_{h,k}^2\mE_h[\bm{\phi}^{\star}_{h}\bm{\phi}^{\star\top}_{h}] = \eta\cdot\tOmega.
\end{equation}
We next verify the Lindeberg condition. It is equivalent to showing that for any $\epsilon>0$,
\begin{equation*}
\lim_{k \to \infty} \frac{1}{\alpha_k } \sum_{h=0}^{k} a_{h,k}^2 \mathbb{E}_{h}\left[ \left\| \bm{\phi}_{h}^{\star}\right\|^2 \cdot \bm{1}_{\|a_{h,k} \bm{\phi}_{h}^{\star}\| \geq \epsilon  \sqrt{\alpha_k}} \right] \leq \lim_{k \to \infty} \frac{1}{\epsilon^{\delta} \alpha_k^{1+0.5\delta}} \sum_{h=0}^{k} a_{h,k}^{2+\delta} \mathbb{E}_{h}[\| \bm{\phi}_{h}^{\star}\|^{2+\delta} ] = 0.
\end{equation*}
By Assumption \ref{assump9}, it suffices to show
\begin{multline}\label{dequ:3}
\frac{1}{\alpha_k^{1+0.5\delta}}\sum_{h=0}^{k}a_{h,k}^{2+\delta} \leq \frac{k+1}{\alpha_k^{1+0.5\delta}}\rbr{\frac{1}{k+1}\sum_{h=0}^{k}a_{h,k}^3}^{\frac{2+\delta}{3}} = \rbr{\frac{\sum_{h=0}^{k}a_{h,k}^3}{\alpha_k^{\frac{3+1.5\delta}{2+\delta}+\frac{1-\delta}{b_1(2+\delta)}}}}^{\frac{2+\delta}{3}}\rightarrow 0.
\end{multline}
In particular, we have
\begin{align*}
\sum_{h=0}^{k}a_{h,k}^3 & = \sum_{h=0}^k\sum_{i=h}^{k}\sum_{i'=h}^{k}\sum_{i''=h}^{k}\prod_{j=i+1}^{k}(1-\alpha_j)\alpha_i\prod_{l=h+1}^{i}(1-\beta_l)\beta_h\prod_{j'=i'+1}^{k}(1-\alpha_{j'})\alpha_{i'}\prod_{l'=h+1}^{i'}(1-\beta_{l'})\beta_h\cdot\\
&\quad\quad \prod_{j''=i''+1}^{k}(1-\alpha_{j''})\alpha_{i''}\prod_{l''=h+1}^{i''}(1-\beta_{l''})\beta_h\\
& = \sum_{i=0}^{k}\sum_{i'=0}^{k}\sum_{i''=0}^{k}\prod_{j=i+1}^{k}(1-\alpha_j)\alpha_i\prod_{j'=i'+1}^{k}(1-\alpha_{j'})\alpha_{i'}\prod_{j''=i''+1}^{k}(1-\alpha_{j''})\alpha_{i''}\cdot\\
& \quad\quad \sum_{h=0}^{\min\{i,i',i''\}}\prod_{l=h+1}^{i}(1-\beta_l)\prod_{l'=h+1}^{i'}(1-\beta_{l'})\prod_{l''=h+1}^{i''}(1-\beta_{l''}) \beta_h^3\\
& \leq 6\sum_{i=0}^{k}\sum_{i'=0}^{i}\sum_{i''=0}^{i'}\prod_{j=i+1}^{k}(1-\alpha_j)\alpha_i\prod_{j'=i'+1}^{k}(1-\alpha_{j'})\alpha_{i'}\prod_{j''=i''+1}^{k}(1-\alpha_{j''})\alpha_{i''}\cdot\\
& \quad\quad\quad \sum_{h=0}^{i''}\prod_{l=h+1}^{i}(1-\beta_l)\prod_{l'=h+1}^{i'}(1-\beta_{l'})\prod_{l''=h+1}^{i''}(1-\beta_{l''}) \beta_h^3 \quad\quad (i\geq i'\geq i'')\\
& = 6\sum_{i=0}^{k}\prod_{j=i+1}^{k}(1-\alpha_j)^3\alpha_i\sum_{i'=0}^{i}\prod_{j'=i'+1}^{i}(1-\alpha_{j'})^2(1-\beta_{j'})\alpha_{i'}\cdot\\
&\quad\quad\quad \sum_{i''=0}^{i'}\prod_{j''=i''+1}^{i'}(1-\alpha_{j''})(1-\beta_{j''})^2\alpha_{i''}\sum_{h=0}^{i''}\prod_{l''=h+1}^{i''}(1-\beta_{l''})^3 \beta_h^3.
\end{align*}
We apply Lemma \ref{lemma_tool} and note that
\begin{align*}
&\lim\limits_{i''\rightarrow\infty} \frac{1}{\beta_{i''}^2}\sum_{h=0}^{i''}\prod_{l''=h+1}^{i''}(1-\beta_{l''})^3 \beta_h^3 = \frac{1}{3},\\
& \lim\limits_{i'\rightarrow\infty} \frac{1}{\alpha_{i'}\beta_{i'}}\sum_{i''=0}^{i'}\prod_{j''=i''+1}^{i'}(1-\alpha_{j''})(1-\beta_{j''})^2\alpha_{i''}\beta_{i''}^2 = \frac{1}{2},\\
& \lim\limits_{i\rightarrow\infty} \frac{1}{\alpha_i^2}\sum_{i'=0}^{i}\prod_{j'=i'+1}^{i}(1-\alpha_{j'})^2(1-\beta_{j'})\alpha_{i'}^2\beta_{i'} = 1,\\
& \lim\limits_{k\rightarrow\infty}\frac{1}{\alpha_k^{\frac{3+1.5\delta}{2+\delta}+\frac{1-\delta}{b_1(2+\delta)}}}\sum_{i=0}^{k}\prod_{j=i+1}^{k}(1-\alpha_j)^3\alpha_i^3 = 0,
\end{align*}
where the last equality holds due to the following facts: 
\begin{enumerate}[label=(\alph*),topsep=0pt]
\setlength\itemsep{0.0em}
\item $b_1\in(0, 1)$: We apply $b_1>(1-\delta)/(1+0.5\delta)$ and have
\begin{equation*}
\frac{1}{\alpha_k^{\frac{3+1.5\delta}{2+\delta}+\frac{1-\delta}{b_1(2+\delta)}}}\sum_{i=0}^{k}\prod_{j=i+1}^{k}(1-\alpha_j)^3\alpha_i^3 = \mO\rbr{\frac{\alpha_k^2}{\alpha_k^{\frac{3+1.5\delta}{2+\delta}+\frac{1-\delta}{b_1(2+\delta)}} }}\rightarrow 0.
\end{equation*}
\item $b_1=1$: We apply $\iota_1>2/3$ and have
\begin{equation*}
\frac{1}{\alpha_k^{\frac{4+0.5\delta}{2+\delta}}}\sum_{i=0}^{k}\prod_{j=i+1}^{k}(1-\alpha_j)^3\alpha_i^3 = \mO\rbr{\frac{\alpha_k^2}{\alpha_k^{\frac{4+0.5\delta}{2+\delta}}} }\rightarrow 0.
\end{equation*}
\end{enumerate}
Thus, we have verified the Lindeberg condition. By the central limit theorem of martingale arrays \citep[Corollary 3.1]{Hall2014Martingale}, the results \eqref{dequ:2} and \eqref{dequ:3} lead to $1/\sqrt{\alpha_k}\cdot\mJ_{3,k}^{(1)}\stackrel{d}{\longrightarrow} \mathcal{N} \left( \bm{0}, \eta \cdot \bm{\Omega}^{\star} \right)$. This completes the proof.

\section{Boundedness of Dual Multipliers}\label{Appendix:E}

Theorem \ref{lemma_mfcq} shows that if the iteration sequence $\bx_k$ generated by our method converges to~a~feasible point $\barbx$ satisfying EGMFCQ (Definition \ref{def_MFCQ}), then the corresponding Lagrangian dual~multipliers $(\blambda_k^{\text{sub}}, \bmu_k^{\text{sub}})$ associated with the subproblem \eqref{equ:SQP:new} with the exact true gradient $\nabla f_k$ are bounded.
This result validates and provides theoretical justification for the boundedness assumption on the dual variables in Assumption \ref{assump2}.

\begin{theorem}\label{lemma_mfcq}
Suppose $\barbx$ is feasible for Problem \eqref{original_problem0} and satisfies EGMFCQ. If $\lim\limits_{k \to \infty} \bx_k = \barbx$, then the corresponding dual multipliers $(\blambda_{k}^{\text{sub}}, \bmu_{k}^{\text{sub}})$ of the SQP subproblem \eqref{equ:SQP:new} are bounded.
\end{theorem}

\vspace{-0.5cm}

\begin{proof}

We prove it by contradiction. Suppose that there exists  a sequence $\{(\bx_k, \barB_k, \blambda_{k}^{\text{sub}}, \bmu_{k}^{\text{sub}})\}$ such that $\bx_k \to \bar{\bm{x}}$, $\left\| (\blambda_{k}^{\text{sub}}, \bmu_{k}^{\text{sub}})\right\| \to \infty$ and $\kappa_1 \mathbf{I} \preceq \barB_k \preceq \kappa_2 \mathbf{I}$. Recall that $\Delta\bx_k$ and $(\blambda_{k}^{\text{sub}}, \bmu_{k}^{\text{sub}})$ denote the primal and dual solutions of the SQP subproblem \eqref{equ:SQP:new} at $\bx_k$ with the true gradient $\nabla f_k$, satisfying the following KKT conditions:
\begin{equation}\label{KKT_subproblem}
\begin{split}
& \nabla f_k + \barB_k \Delta\bx_k+ \nabla \bm{c}_{k}^{\top} \blambda_{k}^{\text{sub}} - \bmu_{1,k}^{\text{sub}} + \bmu_{2,k}^{\text{sub}} = \bm{0},\\
& \theta_k \bm{c}_{k} + \nabla \bm{c}_{k} \Delta\bx_{k} = \bm{0},\quad\quad \bm{\ell} \leq \bx_k + \Delta\bx_{k} \leq \bm{u},\quad\quad \bmu_{1,k}^{\text{sub}}, \bmu_{2,k}^{\text{sub}} \geq \bm{0},\\
& \bmu_{1,k}^{\text{sub} \top} (\bell - \bx_k - \Delta\bx_{k}) = 0,\quad  \bmu_{2,k}^{\text{sub}\top}(\bx_{k} + \Delta\bx_{k} - \bm{u}) = 0.
\end{split}
\end{equation}
Note that the sequences $(\blambda_{k}^{\text{sub}}, \bmu_{k}^{\text{sub}}) / \|(\blambda_{k}^{\text{sub}}, \bmu_{k}^{\text{sub}})\|$ and $\Delta\bx_{k}$ are bounded. Without loss of generality, we assume $(\blambda_{k}^{\text{sub}}, \bmu_{k}^{\text{sub}}) / \| (\blambda_{k}^{\text{sub}}, \bmu_{k}^{\text{sub}})\| \to (\Bar{\blambda}, \Bar{\bmu})$ and $\Delta\bx_{k} \to \Delta\Bar{\bx}$.~Dividing the~two~sides of \eqref{KKT_subproblem} by $\| (\blambda_{k}^{\text{sub}}, \bmu_{k}^{\text{sub}})\|$, taking the limit $k \to \infty$, and applying the feasibility of $\barbx$, we have 
\begin{equation}\label{eq17}
\begin{split}
& \nabla \bc(\bar{\bx})^{\top} \bar{\bm{\lambda}} - \bar{\bm{\mu}}_{1} + \bar{\bm{\mu}}_{2} = \bm{0}, \quad\quad \nabla \bm{c}(\Bar{\bm{x}})\Delta\Bar{\bx} = \0,\\
& \bar{\bm{\mu}}_{1}^{\top} (\bell - \Bar{\bm{x}}) = \Bar{\bm{\mu}}_{1}^{\top} \Delta\Bar{\bx}, \quad \quad \Bar{\bm{\mu}}_{2}^{\top} (\Bar{\bm{x}} - \bm{u}) = -\Bar{\bm{\mu}}_{2}^{\top} \Delta\Bar{\bx}.
\end{split}
\end{equation}
The above four equalities imply
\begin{equation}\label{eq18}
\Bar{\bm{\mu}}_{1}^{\top} (\bm{\ell} - \Bar{\bm{x}}) + \Bar{\bm{\mu}}_{2}^{\top} (\Bar{\bm{x}} - \bm{u}) = 0.
\end{equation}
Since $\bar{\bm{\mu}}_{1}, \bar{\bm{\mu}}_{2} \geq \bm{0}$, we deduce from \eqref{eq18} that $[\Bar{\bm{\mu}}_{1}]_{i} > 0$ only if $[\Bar{\bm{x}}]_{i} = [\bm{\ell}]_i$ and $[\Bar{\bm{\mu}}_{2}]_{i} > 0$ only if $[\Bar{\bm{x}}]_{i} = [\bm{u}]_i$. The EGMFCQ condition at $\Bar{\bm{x}}$ (cf.~Definition \ref{def_MFCQ}) implies that there exists~$\bz \in \mathbb{R}^{d}$ such that $\bm{c}(\Bar{\bm{x}}) + \nabla \bm{c}(\Bar{\bm{x}}) \bz = \bm{0}$, $[\bz]_{i} > 0$ if $[\Bar{\bm{x}}]_{i} = [\bm{\ell}]_i$, and $[\bz]_{i} < 0$ if $[\Bar{\bm{x}}]_{i} = [\bm{u}]_i$. Then, we~have $- \bz^{\top} \Bar{\bm{\mu}}_{1} + \bz^{\top} \Bar{\bm{\mu}}_{2} < 0$ if $\Bar{\bm{x}}$ is on the boundary of the box constraints. 
Multiplying $-\bz$ on both sides of the first equality in \eqref{eq17}, we have 
\begin{equation*}
0 = -\bz^{\top}(\nabla \bm{c}(\Bar{\bm{x}})^{\top} \Bar{\bm{\lambda}} - \Bar{\bm{\mu}}_{1} + \Bar{\bm{\mu}}_{2}) = \bm{c}(\Bar{\bm{x}})^{\top} \Bar{\bm{\lambda}} + \bz^{\top} \Bar{\bm{\mu}}_{1} - \bz^{\top} \Bar{\bm{\mu}}_{2} = \bz^{\top} \Bar{\bm{\mu}}_{1} - \bz^{\top} \Bar{\bm{\mu}}_{2},
\end{equation*}
which contradicts $- \bz^{\top} \Bar{\bm{\mu}}_{1} + \bz^{\top} \Bar{\bm{\mu}}_{2} < 0$. On the other hand, if $\Bar{\bm{x}}$ is in the interior of the box constraints, then $\Bar{\bm{\mu}}_{1} = \Bar{\bm{\mu}}_{2} = \bm{0}$.
Together with the first equality of \eqref{eq17}, the linear independence of the rows of $\nabla \bm{c}(\Bar{\bm{x}})$ shows $\Bar{\bm{\lambda}} = \bm{0}$, which contradicts to the fact that $\|(\Bar{\bm{\lambda}}, \Bar{\bm{\mu}})\| = 1$.~This completes the proof.
\end{proof}

\endgroup

%\printbibliography[title={Appendix References}]
%\end{refsection}

\end{document}